\vspace{-8pt}

\documentclass[10pt,twocolumn,letterpaper]{article}

\usepackage{iccv}
\usepackage{times}
\usepackage{epsfig}
\usepackage{graphicx}
\usepackage{amsmath}
\usepackage{amssymb}

\usepackage{cite}
\usepackage[dvipsnames,table]{xcolor}
\usepackage{multirow}
\usepackage{rotating}
\usepackage{booktabs}

\usepackage{xparse}
\usepackage{cite}

\newcommand{\PAR}[1]{\vskip4pt \noindent{\bf #1~}}
\newcommand{\GT}{pGT\xspace}

\definecolor{plum}{RGB}{112,48,160}
\definecolor{mygreen}{RGB}{146,208,80}

\usepackage[pagebackref=true,breaklinks=true,letterpaper=true,colorlinks,bookmarks=false]{hyperref}

\iccvfinalcopy 
 
\def\iccvPaperID{xxxx} 

\begin{document}

\title{On the Limits of Pseudo Ground Truth in Visual Camera Re-localisation}

\author{Eric Brachmann$^1$
\quad
Martin Humenberger$^2$
\quad 
Carsten Rother$^3$
\quad
Torsten Sattler$^4$\\
$^1$Niantic \quad
$^2$NAVER LABS Europe \quad
$^3$Visual Learning Lab, HCI/IWR, Heidelberg University\\ 
$^4$Czech Institute of Informatics, Robotics and Cybernetics, Czech Technical University in Prague
}

\maketitle
\thispagestyle{empty}

\vspace{-8pt}
\begin{abstract}
Benchmark datasets that measure camera pose accuracy have driven progress in visual re-localisation research. 
To obtain poses for thousands of images, it is common to use a reference algorithm to generate \emph{pseudo ground truth}.
Popular choices
include Structure-from-Motion (SfM) and Simultaneous-Localisation-and-Mapping (SLAM) using additional sensors like depth cameras if available. 
Re-localisation benchmarks thus measure how well each  
method replicates the results of the reference algorithm. 
This begs the question whether the choice of the reference algorithm favours a certain family of re-localisation methods. 
This paper analyzes two widely used re-localisation datasets and shows 
that evaluation outcomes indeed vary with the choice of the reference algorithm. 
We thus question common beliefs in the re-localisation literature, namely that learning-based scene coordinate regression outperforms classical feature-based methods, and that RGB-D-based methods outperform RGB-based methods. 
We argue that any claims on ranking re-localisation methods should take the type of the reference algorithm, and the similarity of the 
methods to the reference algorithm, into account. 
\end{abstract}


\vspace{-12pt}
\section{Introduction}
\label{sec:intro}

\begin{figure}[!t]
    \centering
    \includegraphics[width=0.9\linewidth]{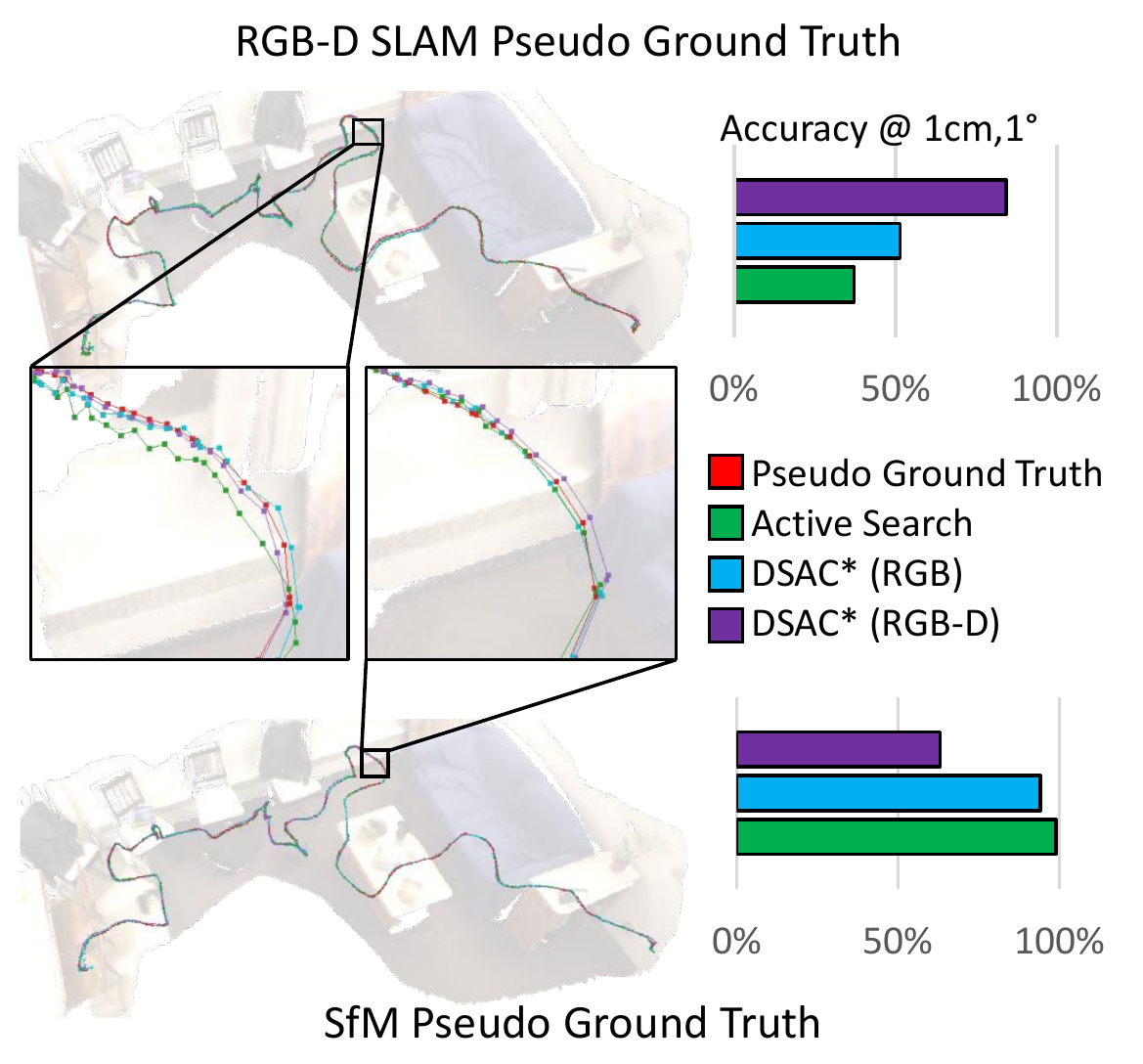}
    \caption{\textbf{Visualisation of the same scene reconstructed by two different reference algorithms.} We show the trajectories of test images estimated by state-of-the-art visual localisation algorithms (left) and the percentage of images localised within 1cm and 1$^\circ$ of error \wrt the pseudo ground truth (right, higher is better). While the underlying 3D scene models are very similar, the relative ranking of methods is inverted. In this paper, we show that the reference algorithm used to create the pseudo ground truth has a significant impact on which method achieves the best results.}
    \label{fig:teaser}
\end{figure}

The availability of benchmark datasets~\cite{Shotton2013CVPR,Kendall2015ICCV,Sattler2018CVPR,Li2010ECCV,Wald2020ECCV,Zhang2020IJCV,Taira2018CVPR,Li2012ECCV,Chen2011CVPR,Maddern2017IJRR,Valentin20163DV,Torii2019TPAMI} has been a driving factor for research on visual re-localisation, a core technology to make autonomous robots~\cite{Lim12CVPR}, self-driving cars~\cite{Heng2019ICRA}, and augmented / mixed reality (AR / MR)  systems~\cite{Castle08ISWC,Arth2011ISMAR,Lynen2015RSS} a reality.
These benchmarks provide camera poses for a set of training and test images.
The training images can be used to create a scene representation, and the test images serve as queries to determine the 3D position and 3D orientation (6DoF pose) of the camera with respect to the scene.
Due to the challenge of jointly estimating the poses of thousands or more images, benchmark datasets are typically generated by a reference algorithm such as SfM or (RGB-)D SLAM~\cite{Kendall2015ICCV,Shotton2013CVPR,Valentin20163DV,Li2010ECCV,Li2012ECCV}. 
As such, benchmarks measure how well visual re-localisation methods are able to replicate the results of the reference algorithm. 

Ideally, the choice of reference algorithm should not matter  as long as it faithfully estimates the camera poses of the training and test images. 
In particular, the choice of reference algorithm should not affect the ranking of methods on a benchmark. 
In practice however, different reference algorithms optimise different cost functions, \eg, reprojection errors of sparse point clouds for SfM~\cite{Schoenberger2016CVPR,Wu133DV} or alignment errors in 3D space for depth-based SLAM methods~\cite{Newcombe2011ISMAR,Izadi2011UIST,Schoeps2019CVPR,Dai2017TOG}, leading to different local minima. 
We ask to what degree the choice of reference algorithm impacts the ranking of methods on a benchmark.
This is an important question as it pertains to whether or not we can draw absolute conclusions, \eg, algorithm A is better than algorithm B or using component C improves accuracy. 
Interestingly, to the best of our knowledge, this question has not received much attention in the re-localisation literature. 

The main focus of this paper is to investigate how the choice of reference algorithms impacts the measured performance of visual re-localisation algorithms. 
To this end, we compare two types of reference algorithms (depth-based SLAM and SfM) on two popular benchmark datasets~\cite{Shotton2013CVPR,Valentin20163DV}. 
Detailed experiments with state-of-the-art re-localisation algorithms show that the choice of reference algorithm can have a profound impact on the ranking of methods. 
In particular, as illustrated in Fig.~\ref{fig:teaser}, we show that depending on the reference algorithm, a modern end-to-end-trainable approach~\cite{brachmann2020ARXIV} either outperforms or is outperformed by a classical, nearly 10 year-old baseline~\cite{Sattler2012ECCV,Sattler2017PAMI}. 
Similarly, the choice of whether to use depth maps or SfM point clouds to represent a scene can improve or decrease performance depending on the reference algorithm.
Our results show that we as a community should be careful when drawing conclusions from existing benchmarks. 
Instead, it is necessary to take into account that certain approaches more closely resemble the reference algorithm than others. 
The former are better able to replicate imperfections in a reference algorithm's \emph{pseudo ground truth} (\GT). 
This natural advantage should be discussed when evaluating localisation results and designing new benchmarks. 

In detail, this paper makes the following contributions: 

\noindent \textbf{1)} 
we show that the choice of a reference algorithm for obtaining \GT 
poses can have a significant impact on the relative ranking of methods,
to the extend that the 
rankings of methods can be (nearly) completely reversed. 
This implies that published results for visual re-localisation should always be considered under the aspect of which algorithm was used to create the \GT. 

\noindent \textbf{2)} we provide a comparison of \GT generated by RGB-only SfM and \mbox{(RGB-)D} SLAM on the 7Scenes~\cite{Shotton2013CVPR} and 12Scenes~\cite{Valentin20163DV} datasets, which are widely used~\cite{Shotton2013CVPR,Guzman2014CVPR,Brachmann2016CVPR,Brachmann2017CVPR,Brachmann2018CVPR,Brachmann2019ICCVa,Brachmann2019ICCVb,Kendall2015ICCV,Kendall2017CVPR,Walch2017ICCV,Brahmbhatt2018CVPR,Valentin2015CVPR,Valentin20163DV,Cavallari2017CVPR,Cavallari20193DV}. 
We show that none is clearly superior than the other. 
We show that commonly accepted results from the literature (RGB-D variants of re-localisation methods outperform their RGB-only counterparts; scene coordinate regression is more accurate than feature-based methods) are not absolute but depending on the \GT. 

\noindent \textbf{3)} we are not aware of prior work aimed at evaluating the extent to which conclusions about localisation performance can be drawn from existing benchmarks. 
As such, this paper is the first to raise awareness that the limitations of the \GT for re-localisation need to be discussed in order to make valid comparisons across methods.

Our new \GT 
and our evaluation pipeline are available at \href{https://github.com/tsattler/visloc_pseudo_gt_limitations/}{github.com/tsattler/visloc\_pseudo\_gt\_limitations/}.

\vspace{-6pt}
\section{Related Work}
\label{sec:rel_work}
\vspace{-3pt}

\begin{figure*}[!t]
    \centering
    \includegraphics[width=1\linewidth]{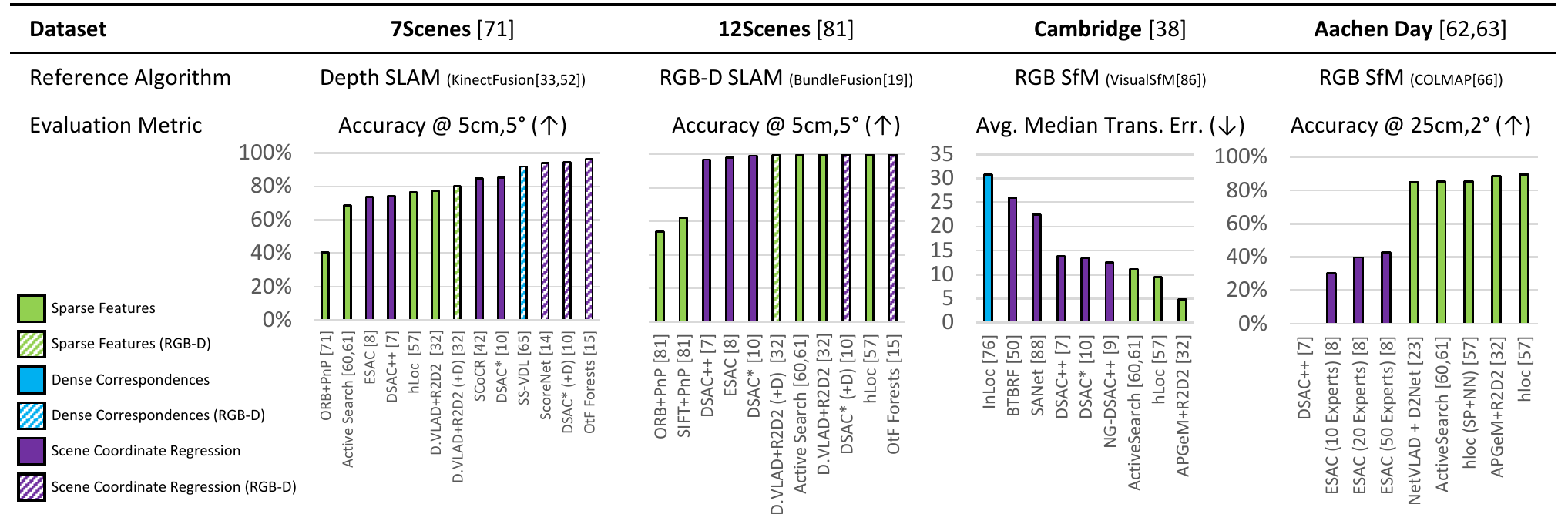}
    \caption{\textbf{State-of-the-art results for four benchmarks.} For indoor scenes (7Scenes) learned scene coordinate regression methods (\textcolor{plum}{purple}) outperform feature-based approaches (\textcolor{mygreen}{green}), and depth-based methods (dashed) outperform RGB-only methods.
    For outdoor scenes (Cambridge and Aachen) feature-based methods, in turn, outperform scene coordinate regression.
    We show that this behavior is coupled to the type of reference algorithm used to create the pGT. Results of 12Scenes are saturated under the common 5cm,5$^\circ$ threshold.\nocite{Shotton2013CVPR,Valentin20163DV,brachmann2020ARXIV,Brachmann2019ICCVa,Brachmann2018CVPR,Brachmann2019ICCVb,Cavallari2017CVPR,Cavallari2019TPAMI,Cavallari20193DV,Schmidt2017RAL,Sattler2017PAMI,Taira2018CVPR,Meng2017IROS,Yang2019ICCV,Sarlin2019CVPR,Sarlin2020CVPR,DeTone2018CVPRWorkshops,Shotton2013CVPR,Valentin20163DV,Kendall2015ICCV,Sattler2012BMVC,Sattler2018CVPR,Newcombe2011ISMAR,Dai2017TOG,Wu-2011-Visualsfm,Schoenberger2016CVPR,Dusmanu2019CVPR}}
    \label{fig:relwork}
\end{figure*}

The difficulty of obtaining ground truth varies for different tasks in computer visions. 
For tasks with a low dimensional structure, \eg image classification and object detection, human annotations are effective 
\cite{pascal-voc-2012} 
and can be scaled via crowd sourcing \cite{Deng2009CVPR}.
For tasks with a more complex output, \eg image segmentation or optical flow, annotation time quickly rises 
to a level which severely affects the scale and cost of associated datasets \cite{Donath2013ICVS, Cordts2016CVPR,geiger2013vision,Menze2015CVPR}.

The 6DoF camera pose estimation task comes with the added difficulty that humans are not skilled at directly annotating the
camera poses. 
Instead they annotate image correspondences as input to an optimization problem to recover the pose indirectly \cite{Sattler2018CVPR,Taira2018CVPR, Torii2019TPAMI}.
Since many correspondences are required for a stable pose estimate, such an annotation approach does not scale beyond a few hundred images.
Further, 
the annotations are usually only precise up to a few pixels, which, depending on the distance to the scene, can result in significant pose uncertainty \cite{Sattler2018CVPR,Zhang2020IJCV,Torii2019TPAMI}.

As an alternative, 
the recording camera can be 
tracked by an external tracking system \cite{sturm12iros, Burri2016IJRR, Schoeps2019CVPR}.
While providing high precision poses, capturing a diverse set of scenes is challenging due to the complicated setup, \ie, 
installation and calibration, 
ensuring good visibility of the sensors, \etc 
Similarly, industrial level LiDAR scanners have been used to produce high quality scans of landmarks, but the corresponding datasets provide only few scenes with limited spatial extent \cite{Strecha2008CVPR, Knapitsch2017TOG, Schoeps2017CVPR}.
GPS-INS systems that combine GPS with inertial navigation systems (INS) have also been used to track camera poses on a large scale \cite{geiger2013vision, Maddern2017IJRR}. 
Yet, post-processing is still required to obtain a higher accuracy~\cite{Sattler2018CVPR}. 

Synthetic datasets come with \emph{true} ground truth, but  most current datasets, \eg, Habitat\cite{habitat19iccv}, are limited in diversity of low-level noise, illumination conditions or specular reflections.
Therefore, data association, which is at the core of re-localisation, can become too easy.
An example are the very low errors reported in~\cite{Sattler2019CVPR} for Active Search~\cite{Sattler2012ECCV, Sattler2017PAMI}  on a synthetic version of the Cambridge Landmarks dataset. 

The vast majority of re-localisation benchmarks follow an automatic approach to ground truth recovery using a reference algorithm \cite{Li2010ECCV, Chen2011CVPR,Li2012ECCV, Shotton2013CVPR,Kendall2015ICCV,Valentin20163DV,Sattler2018CVPR, Taira2018CVPR, Torii2019TPAMI,Zhang2020IJCV, Wald2020ECCV, Jin2020}.
Popular choices are 
SfM~\cite{Schoenberger2016CVPR,Wu133DV}, often for large-scale outdoor environments \cite{Li2010ECCV,Chen2011CVPR, Li2012ECCV, Torii2019TPAMI, Kendall2015ICCV, Sattler2018CVPR}, and depth-based 
SLAM~\cite{Newcombe2011ISMAR,Izadi2011UIST,Schoeps2019CVPR,Dai2017TOG}, often for small-scale indoor environments \cite{Shotton2013CVPR,Valentin20163DV, Wald2020ECCV}. 
Hybrid solutions also exist, such as ICP-based registration of LiDAR scans followed by an SfM-based registration of RGB images \cite{Sattler2018CVPR, Taira2018CVPR}.
Some benchmarks use human visual inspection 
as a final quality control and verification stage \cite{Sattler2018CVPR, Taira2018CVPR, Jin2020, Zhang2020IJCV}, and state-of-the-art reference algorithms are found to provide high-quality reconstructions and pose tracks.
However, as
shown in Fig.~\ref{fig:teaser}, subtle differences in the output of reference algorithms, unlikely to be recognized by visual checks, can have significant influence on the evaluation outcome of a benchmark.

Such evaluation artifacts have the potential to challenge some conclusions that have previously been drawn in the literature.
Fig.~\ref{fig:relwork} shows published results in re-localisation research on the popular indoor datasets 7Scenes \cite{Shotton2013CVPR} and 12Scenes \cite{Valentin20163DV}, and the  popular outdoor datasets Cambridge Landmarks \cite{Kendall2015ICCV} and Aachen Day \cite{Sattler2012BMVC, Sattler2018CVPR}.
We compare the dominant families of re-localisation methods, scene coordinate regression and sparse feature-based matching. 
Scene coordinate regression methods use a learned model, a neural network or a random forest, to predict dense image-to-scene correspondences \cite{Shotton2013CVPR, Valentin2015CVPR, Brachmann2016CVPR, Brachmann2017CVPR, Brachmann2018CVPR, Brachmann2019ICCVa, Brachmann2019ICCVb, brachmann2020ARXIV, Cavallari2017CVPR, Cavallari2019TPAMI, Yang2019ICCV, li2020hierarchical}.
RGB-D variants of scene coordinate regression methods dominate rankings for indoor re-localisation, which has been attributed to the inherent difficulty of the indoor scenario regarding texture-less surfaces and ambiguous structures that make it difficult to find and match sparse features \cite{Shotton2013CVPR, Kendall2015ICCV, Walch2017ICCV, Brachmann2017CVPR}. 
For outdoor re-localisation, classical approaches, which match hand-crafted \cite{Sattler2017PAMI, Shotton2013CVPR, Valentin20163DV} or learned descriptors \cite{Sarlin2019CVPR, DeTone2018CVPRWorkshops, Sarlin2020CVPR, HumenbergerX20Kapture} at sparse feature locations to a 3D SfM reconstruction, achieve vastly superior results compared to scene coordinate regression.
This has been attributed to an inability of scene coordinate regression to scale to spatially large scenes \cite{Sattler2018CVPR,Taira2019TPAMI}. 
We offer a different explanation regarding the performance of re-localisers in different settings by taking the reference algorithms into account that were used to create the associated benchmarks.

\vspace{-6pt}
\section{Datasets and Reference Algorithms}
\vspace{-3pt}
In order to measure the impact different reference algorithms have on localisation performance, we consider \GT generated using (RGB-)D and sparse RGB-only data. 
We use the popular 7Scenes~\cite{Shotton2013CVPR} and 12Scenes~\cite{Valentin20163DV} datasets as they provide depth maps and \GT poses for both test and training images. 
This is in contrast to other common benchmarks~\cite{Sattler2018CVPR,Taira2018CVPR,Kendall2015ICCV,Li2010ECCV,Wald2020ECCV,Torii2019TPAMI}, which do not provide depth information for test and train images~\cite{Sattler2018CVPR,Taira2018CVPR,Kendall2015ICCV,Li2010ECCV,Torii2019TPAMI}\footnote{
Estimated depth using motion stereo requires and is influenced by the \GT.  
Single-view depth prediction offers limited quality and stability.
}
or do not make the poses of the test images publicly available~\cite{Sattler2018CVPR,Taira2018CVPR,Wald2020ECCV}. 
In the following, we describe the datasets, 
their original \GT, 
and how we create an additional \GT for each dataset via RGB-only SfM. 
The purpose of this section is to familiarize the reader with the datasets and reference algorithms before evaluating the resulting \GT variants (Sec.~\ref{sec:comparison_pgt}) and measuring their impact on re-localisation performance (Sec.~\ref{sec:reloc_eval}).

\vspace{-2pt}
\subsection{Incremental Depth SLAM}
\label{sec:kinectfusion}
\vspace{-2pt}
Camera poses can be tracked by incrementally registering dense depth measurements 
to a 3D scene representation. 
KinectFusion~\cite{Newcombe11ICCV,Izadi2011UIST}, an early incarnation of such a system, uses a truncated signed distance function (TSDF) to represent the scene. 
The TSDF is 
updated by merging depth maps $\mathbf{D}_i$ of frames $i$ into a weighted average 
\vspace{-6pt}
\begin{equation} 
    \min_{\mathbf{F}} \sum_i  ||\mathbf{W}_{\mathbf{D}_i}\mathbf{F}_{\mathbf{D}_i}-\mathbf{F}|| \enspace , \label{eq:kfv}\\
    \vspace{-8pt} \nonumber 
\end{equation}
where $\mathbf{F}$ and $\mathbf{F}_{\mathbf{D}_i}$ denote the TSDF representations of the scene and of depth map $\mathbf{D}_i$, respectively.
Weights $\mathbf{W}_{\mathbf{D}_i}$ capture the measurement uncertainty of the depth recording.
For tracking the 6DoF camera pose of a new frame with rotation $\mathtt{R}_i$ and translation $\mathbf{t}_i$, KinectFusion minimises the point-plane distance between the measured depth and a depth rendering of the scene's TSDF volume 
\begin{equation}
    \min_{\mathtt{R}_i, \mathbf{t}_i} \sum_\mathbf{x}  ||\left(\mathbf{V}_i(\mathbf{x}) - [\mathtt{R}_i|\mathbf{t}_i]~\hat{\mathbf{V}}^g_{i-1}(\hat{\mathbf{x}})\right)^\top \hat{\mathbf{N}}^g_{i-1}(\hat{\mathbf{u}})|| \enspace . \label{eq:kft}\\
    \vspace{-8pt} 
\end{equation}
The objective is minimised over 2D pixel positions $\mathbf{x}$. 
Measured depth and rendered depth are back-projected to 3D vertex maps $\mathbf{V}_i$ and $\hat{\mathbf{V}}^g_{i-1}$, respectively. 
Particularly, $\hat{\mathbf{V}}^g$ denotes the rendered vertex map of the scene in world (or \emph{global}) coordinates and 
$\hat{\mathbf{N}}^g$ denotes rendered normals.

\noindent \textbf{KinectFusion \GT for 7Scenes.}
Shotton \etal \cite{Shotton2013CVPR} created the 7Scenes dataset for re-localisation by scanning seven small-scale indoor environments with Kinect v1 and KinectFusion.
Every scene was scanned multiple times by different users, and the resulting 3D scene models were registered using ICP~\cite{Rusinkiewicz2001ICP}.
No global optimization within a single scan or across multiple scans was performed, and any camera drift remains unaccounted for in the \GT of 7Scenes.
In terms of RGB-D images, the 7Scenes dataset only provides uncalibrated output of the Kinect, \ie, RGB images and depth maps are not registered, and the camera poses align with the depth sensor, 
not the RGB camera.

\subsection{Globally Optimised RGB-D SLAM}
\label{sec:bundlefusion}

To reduce 
camera drift during incremental scanning, more recent RGB-D SLAM systems like BundleFusion~\cite{Dai2017TOG} jointly optimise all 6DoF camera poses. 
The parameter vector $\mathcal{X}= (\mathtt{R}_0, \mathbf{t}_0, ..., \mathtt{R}_N, \mathbf{t}_N)$ stacks rotations and translations of all frames recorded and 
BundleFusion optimises 
\vspace{-6pt}
\begin{equation}
    \min_{\mathcal{X}} w_\text{sprs} E_\text{sprs}(\mathcal{X}) + w_\text{pht} E_\text{pht}(\mathcal{X}) + w_\text{geo} E_\text{geo}(\mathcal{X}) 
    \enspace . \label{eq:bf}\\
    \vspace{-3pt} \\
\end{equation}
The term $E_\text{sprs}$ minimises the Euclidean distance for sparse SIFT~\cite{Lowe04IJCV} feature matches across all images. 
Note that this term minimises a 3D distance, not a reprojection error, since the depth of image pixels is known.
The term $E_\text{pht}$ is a photometric loss that ensures a consistent gradient of image luminance across registered images.
Finally, $E_\text{geo}$ optimises a point-to-plane distance of depth maps with projective data association similar to KinectFusion, see Eq.~\ref{eq:kft}.

\noindent \textbf{BundleFusion \GT for 12Scenes.}
Valentin \etal \cite{Valentin20163DV} scanned twelve small-scale indoor environments for their 12Scenes dataset.
They utilized a structure.io depth sensor mounted on an iPad that provided associated color images.
Different from 7Scenes, 12Scenes comes with fully calibrated and synchronized color and depth images 
and depth is registered to the color images. 
Each room was scanned two times, once for training and once for testing, and both scans of each scene were registered manually.

\vspace{-3pt}
\subsection{Pseudo Ground Truth via SfM}
\label{sec:colmap}
\vspace{-3pt}
A common approach to generate \GT
~\cite{Li2010ECCV,Li2012ECCV,Sattler2018CVPR,Torii2019TPAMI,Kendall2015ICCV,Sun2017CVPR} is to use (incremental) SfM 
algorithms~\cite{Schoenberger2016CVPR,Wu133DV,Snavely08IJCV}. 
SfM methods rely on sparse local features such as SIFT~\cite{Lowe04IJCV} to establish feature matches between images, which are then used to recover camera poses and 3D scene structure. 
SfM is usually applied jointly on the test and training images to jointly recover the camera poses of all images~\cite{Sattler2018CVPR,Li2012ECCV,Kendall2015ICCV}. 

SfM algorithms minimise the reprojection error between the estimated 3D points and their corresponding feature measurements in the images, optimising the problem \vspace{-6pt}
\begin{equation}
    \min_{\mathtt{R}_i, \mathbf{\theta}_i, \mathbf{t}_i, \mathbf{X}_j} \sum_i \! \sum_j \! \delta_{ij} \rho\left(||\mathbf{x}_{ij} \! - \! \pi\left(\mathtt{R}_i\mathbf{X}_j \! + \! \mathbf{t}_i, \mathbf{\theta}_i)\right)||^2\right)  \label{eq:ba}\\
    \vspace{-6pt}
\end{equation}
during Bundle Adjustment (BA)~\cite{Triggs2000VATP}. 
Here, $\mathbf{\theta}_i$ are the intrinsic camera parameters, 
$\mathbf{X}_j$ is the $j^\text{th}$ 
3D point, $\delta_{ij} \in \{0, 1\}$ indicates whether the $j^\text{th}$ 
3D point is visible in the $i^\text{th}$ 
image,  $\mathbf{x}_{ij}$ is the corresponding 2D feature position of the $j^\text{th}$ 
3D point in the $i^\text{th}$ 
image, $\pi$ is the projection function, and $\rho$ is a robust cost function~\cite{Triggs2000VATP}. 
SfM only reconstructs the scene up to an arbitrary scaling factor. 
Known  
3D distances
are used to recover the absolute scale of the model.

\noindent \textbf{SfM \GT for 7Scenes and 12Scenes.}
As the basis for our analysis, 
we generate an alternative \GT for 7Scenes and 12Scenes. 
First, we reconstruct the scene with SfM using only the training images. 
Next, we continue the reconstruction process with the test images while keeping the training camera poses fixed. 
This strategy ensures that the training poses are not affected by the test images, as would be the case in practice. 
Finally, we recover the scale by robustly aligning the positions of all cameras to those of the original \GT.
We implement this process with
COLMAP~\cite{Schoenberger2016CVPR}, 
using the same camera intrinsics for all images in a scene. 

This approach 
failed for the office2/5a and %
5b datasets of 12Scenes. 
Both 
depict scenes with highly repetitive structures. 
As a result, the SfM reconstruction collapses, \ie, visually similar but physically different parts of the scene are merged. 
Thus, for both 
scenes, we first triangulate 3D points 
using the original \GT. 
Next, we apply 10 iterations 
consisting of BA followed by merging and completing 3D points: 
nearby 3D points with matching features are merged and new features are added to 3D points if possible. 

Some images of 12Scenes that were not registered by BundleFusion were reconstructed using COLMAP. 
Also, for the office2/5a and 5b scenes, we removed 61 images (out of 3,354 images contained in both scenes together) that we identified as obvious outliers via visual inspection. 

\vspace{-6pt}
\section{Comparison of Pseudo Ground Truths}
\label{sec:comparison_pgt}
\vspace{-3pt}
Given the two versions of \GT for each scene, estimated using (RGB-)D SLAM respectively SfM, a natural question is whether one version is more precise than the other. 
In this section, we quantitatively and qualitatively show that no version of the \GT 
is clearly preferably over the other: 
we first show that the SfM \GT outperforms the (RGB-)D SLAM version according to metrics that are optimised during the SfM process. 
We then show that the (RGB-)D SLAM \GT in turn outperforms the SfM version in terms of dense 3D point alignment, \ie, the metrics optimised by depth-based methods. 
Thus, both versions can be considered as valid \GT for re-localisation experiments. 
Note that our analysis is focused on the two particular datasets. 
For a more general analysis of various reference algorithms, \eg, about the influence of calibration accuracy, we refer to~\cite{Schoeps2019CVPR}.

\begin{table*}[th]
\begin{center}
\setlength{\tabcolsep}{2pt}
\scriptsize{
\begin{tabular}{|cr|c|c|c||c|c|c||c|c|c||c|c|c||c|c|c||c|c|c||c|c|c|}
\cline{1-23} 

\multirow{6}{*}{\begin{sideways}7Scenes\end{sideways}} & & \multicolumn{3}{c||}{Chess} & \multicolumn{3}{c||}{Fire} & \multicolumn{3}{c||}{Heads} & \multicolumn{3}{c||}{Office} & \multicolumn{3}{c||}{Pumpkin} & \multicolumn{3}{c||}{Red Kitchen} & \multicolumn{3}{c|}{Stairs}
\\ \cline{3-23}  
& Ref.~Algo.& orig. & +BA & SfM & orig. & +BA & SfM & orig. & +BA & SfM & orig. & +BA & SfM & orig. & +BA & SfM & orig. & +BA & SfM & orig. & +BA & SfM 
\\ \cline{3-23}
& \#3D & 433k & 204k & 190k & 628k & 316k & 296k & 104k & 73k & {70k} & 515k & 261k & 249k  & 282k & 131k & 150k  & 1.0M & 455k & 472k  & 178k & 119k & 132k 
\\ \cline{3-23}
& \#feat. & 7.4M & 7.7M & \textbf{7.9M} & 9.9M & 10.2M & \textbf{10.4M} & 1.4M & 1.4M & \textbf{1.5M}  & 7.4M & 7.8M & 8.0M  & 4.5M & 5.0M & \textbf{5.2M}& 14.1M & 15.8M & \textbf{16.5M}  & 2.3M & 2.4M & \textbf{2.6M}  
\\ \cline{3-23}
& track & 17.0 & 37.7 & \textbf{41.6} & 15.7 & 32.3 & \textbf{35.0} & 13.4 & 19.8 & \textbf{20.9} & 14.3 & 29.9 & 32.1  & 15.9 & \textbf{38.2} & 35.2  & 13.5 & 34.7 & \textbf{35.0}  & 13.0 & \textbf{20.2} & 19.5
\\ \cline{3-23}
& err. [px] & 1.74 & 1.40 & \textbf{1.25} & 1.54 & 0.95 & \textbf{0.88} & 1.49 & 1.10 & \textbf{1.01}  & 1.68 & 1.25 & 1.12  & 1.76 & 1.40 & \textbf{1.24}  & 1.73 & 1.28 & \textbf{1.13}  & 1.62 & 1.25 & \textbf{1.10}
\\  \hline \hline

\multirow{10}{*}{\begin{sideways}12Scenes\end{sideways}} && \multicolumn{3}{c||}{apt1/kitchen} & \multicolumn{3}{c||}{apt1/living} & \multicolumn{3}{c||}{apt2/bed} & \multicolumn{3}{c||}{apt2/kitchen} & \multicolumn{3}{c||}{apt2/living} & \multicolumn{3}{c||}{apt2/luke} & \multicolumn{3}{c|}{office1/gates362}
\\ \cline{3-23} 
& \#3D & 146k & 106k & 104k & 166k & 112k & 120k & 245k & 201k & 171k & 208k & 121k & 119k & 148k & 116k & 121k & 201k & 135k & 140k & 658k & 424k & 419k 
\\ \cline{3-23}
& \#feat. & 1.3M & 1.3M & \textbf{1.4M} & 1.4M & 1.5M & \textbf{1.6M} & 1.9M & 2.0M & \textbf{2.2M} & 2.8M & 2.9M & \textbf{3.0M}  & 1.2M & 1.2M & \textbf{1.3M} & 1.5M & \textbf{1.7M} & \textbf{1.7M} & 9.7M & 10.1M & \textbf{10.3M} 
\\ \cline{3-23}
& track & 8.6 & 12.6 & \textbf{13.0} & 8.3 & \textbf{13.6} & 13.3 & 7.8 & 11.9 & \textbf{12.7} & 13.3 & 24.0 & \textbf{24.9} & 7.8 & \textbf{10.4} & \textbf{10.4} & 7.3 & \textbf{12.4} & \textbf{12.5} & 14.7 & 23.8 & \textbf{24.7} 
\\ \cline{3-23}
& err. [px] & 1.63 & 1.33 & \textbf{1.25} & 1.72 & 1.38 & \textbf{1.28} & 1.58 & 1.03 & \textbf{0.97} & 1.72 & 1.21 & \textbf{1.07} & 1.59 & 1.18 & \textbf{1.12} & 1.75 & 1.42 & \textbf{1.33} & 1.69 & 1.31 & \textbf{1.18} 
\\  \cline{3-23} \cline{3-23}
& & \multicolumn{3}{c||}{office1/gates381} & \multicolumn{3}{c||}{office1/lounge} & \multicolumn{3}{c||}{office1/manolis} & \multicolumn{3}{c||}{office2/5a} & \multicolumn{3}{c||}{office2/5b} & \multicolumn{6}{c}{ }
\\ \cline{3-17}
& \#3D & 695k & 447k & 471k & 161k & 116k & 120k & 364k & 275k & 273k & 261k & \multicolumn{2}{c||}{202k} & 607k & \multicolumn{2}{c||}{580k} 
\\ \cline{3-17}
& \#feat. & 6.8M & 7.5M & \textbf{7.9M} & 1.4M & 1.4M & \textbf{1.5M} & 3.7M & 3.8M & \textbf{3.9M} & 1.7M & \multicolumn{2}{c||}{\textbf{2.0M}} & 4.0M & \multicolumn{2}{c||}{\textbf{4.6M}} 
\\ \cline{3-17}
& track & 9.9 & \textbf{16.8} & \textbf{16.7} & 8.5 & \textbf{12.4} & \textbf{12.5} & 10.0 & 13.7 & \textbf{14.2} & 6.3 & \multicolumn{2}{c||}{\textbf{9.7}} & 6.7 & \multicolumn{2}{c||}{\textbf{7.9}} 
\\ \cline{3-17}
& err. [px] & 1.60 & 1.17 & \textbf{1.09} & 1.69 & 1.27 & \textbf{1.19} & 1.60 & 1.27 & \textbf{1.19} & 1.57 & \multicolumn{2}{c||}{\textbf{0.83}} & 1.43 & \multicolumn{2}{c||}{\textbf{0.80}} 
\\  \cline{1-17}
\end{tabular}
}
\end{center}
\vspace{-6pt}
\caption{\textbf{SfM statistics for different pseudo ground truth (\GT) versions.} We report the number of 3D points (\#3D), the number of triangulated features (\#feat.), the average track length (track), and the mean reprojection error (err.) in px. Longer tracks generated from more features and a lower reprojection error indicate more accurate poses.  
We show results for the original~\cite{Shotton2013CVPR,Valentin20163DV} training and test poses generated by (RGB-)D SLAM (orig.), the original poses iteratively refined by alternating bundle adjustment and point merging (+BA), and the SfM \GT (SfM). For office2/5a and office2/5b, SfM from scratch failed and the SfM \GT is generated using the +BA strategy.}
\label{tab:stats_sfm}%
\end{table*}

\PAR{Evaluation based on SfM metrics.} 
The first experiment focuses on standard metrics used to evaluate SfM reconstructions~\cite{Schoenberger2017CVPR}. 
We measure the number of 3D points (\#3D), the number of feature observations (\#feat.) used to triangulate the 3D points, the average track length (track), \ie, the average number of features used to triangulate a 3D point, and 
the average reprojection error (err.). 
For the same number of images in a 3D model, 
more observations and longer tracks, 
especially in combination with a lower reproj.~error, indicate higher camera pose accuracy. 
Shorter tracks, \ie, more 3D points, 
indicate that a single physical 3D point is represented by multiple SfM points: 
due to pose inaccuracies, 
no single SfM point  
projects within the 
error threshold
used for robust triangulation
~\cite{Schoenberger2016CVPR} for all its measurements.  

We compare the SfM \GT with point clouds obtained by triangulating the scenes from the original (RGB-)D \GT. For 7Scenes, we adjust the original \GT to account for the offset between RGB camera and depth sensor using  the calibration from 
\cite{wolf2014CVIU}.
We use the same set of matches and the same COLMAP parameters 
for both \GT versions 
and 
use training and test images to calculate the statistics.

\begin{figure}[!t]
    \centering
    \includegraphics[width=0.48\linewidth]{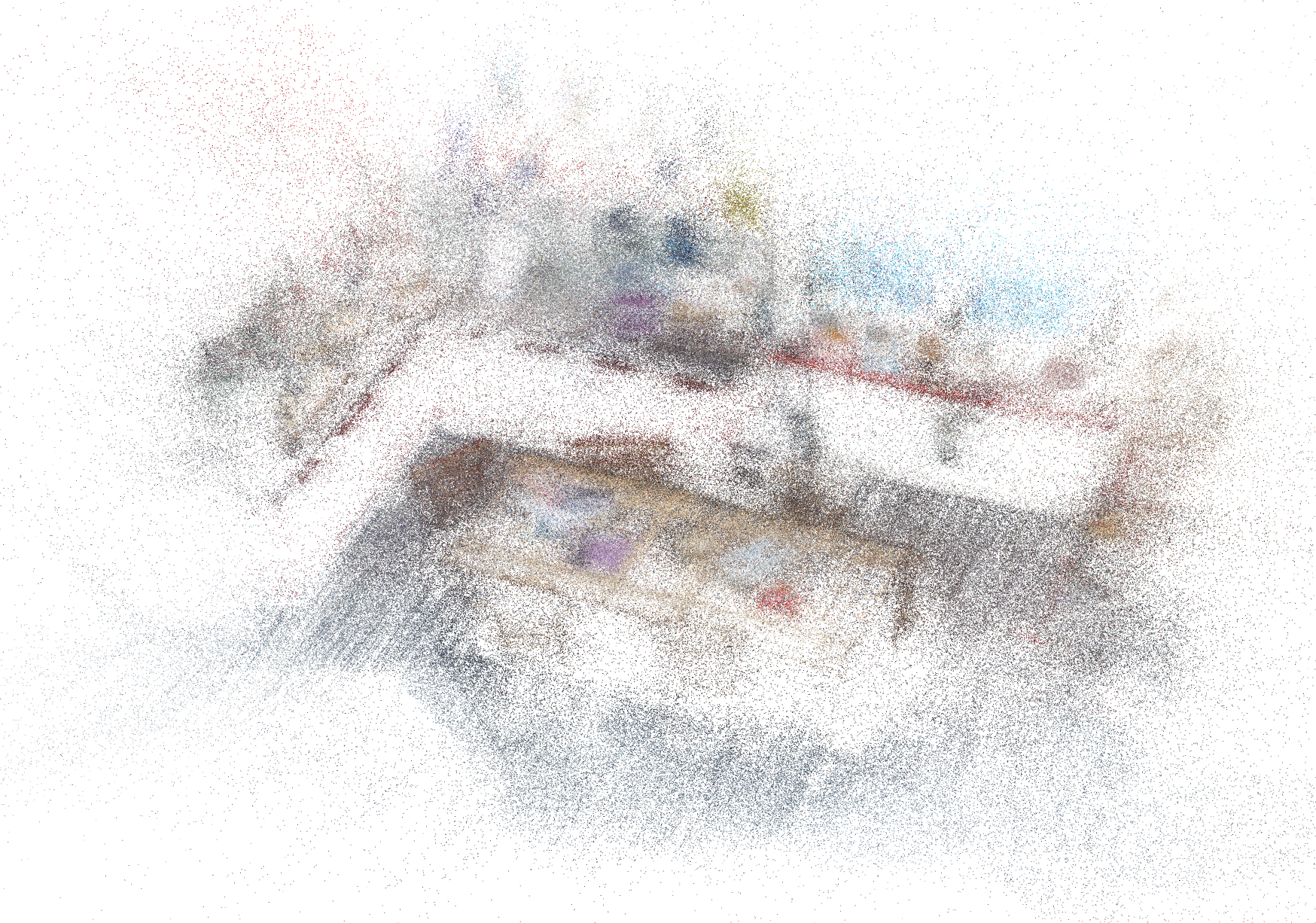}%
    \includegraphics[width=0.48\linewidth]{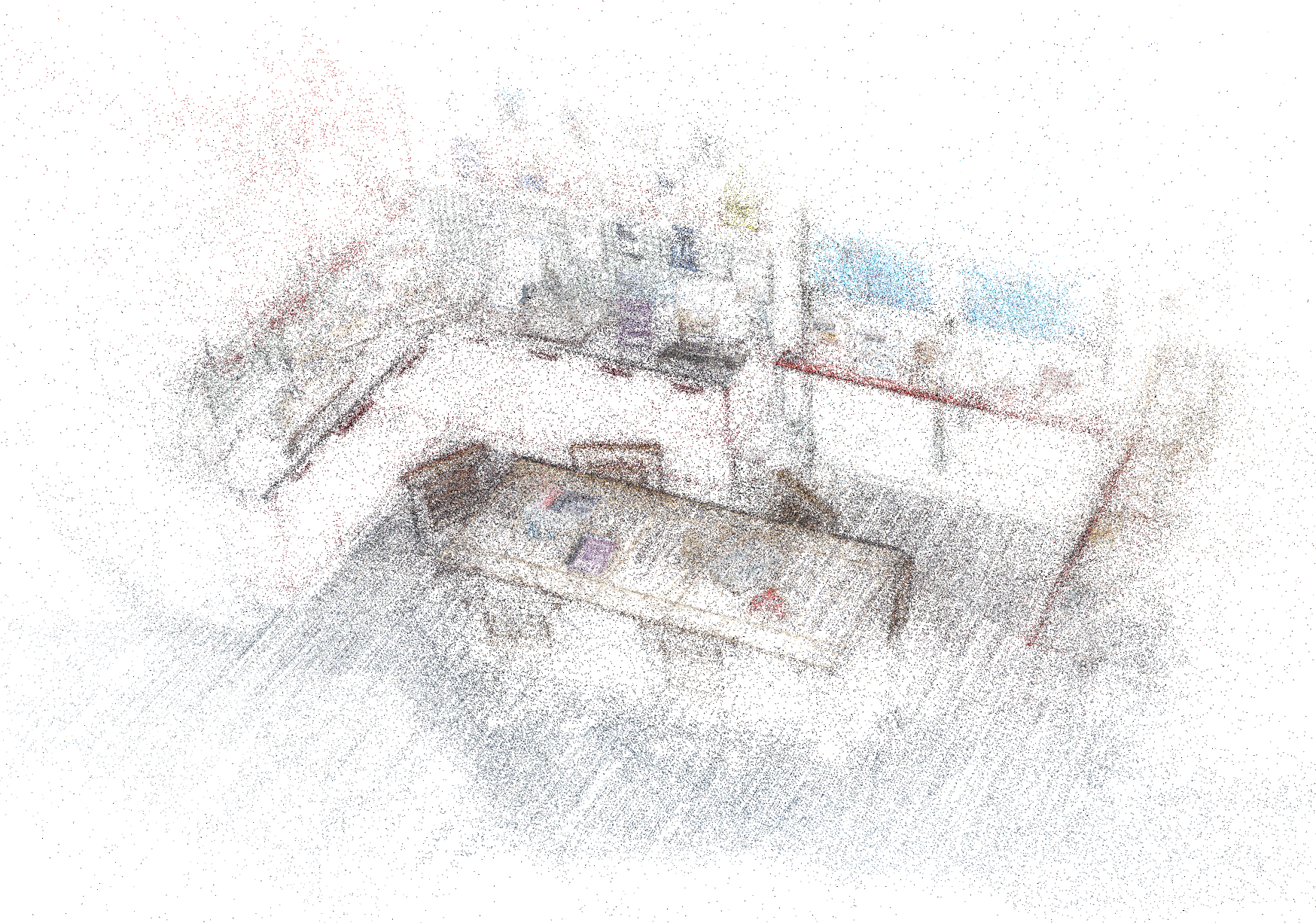}\\%
    \includegraphics[width=0.48\linewidth]{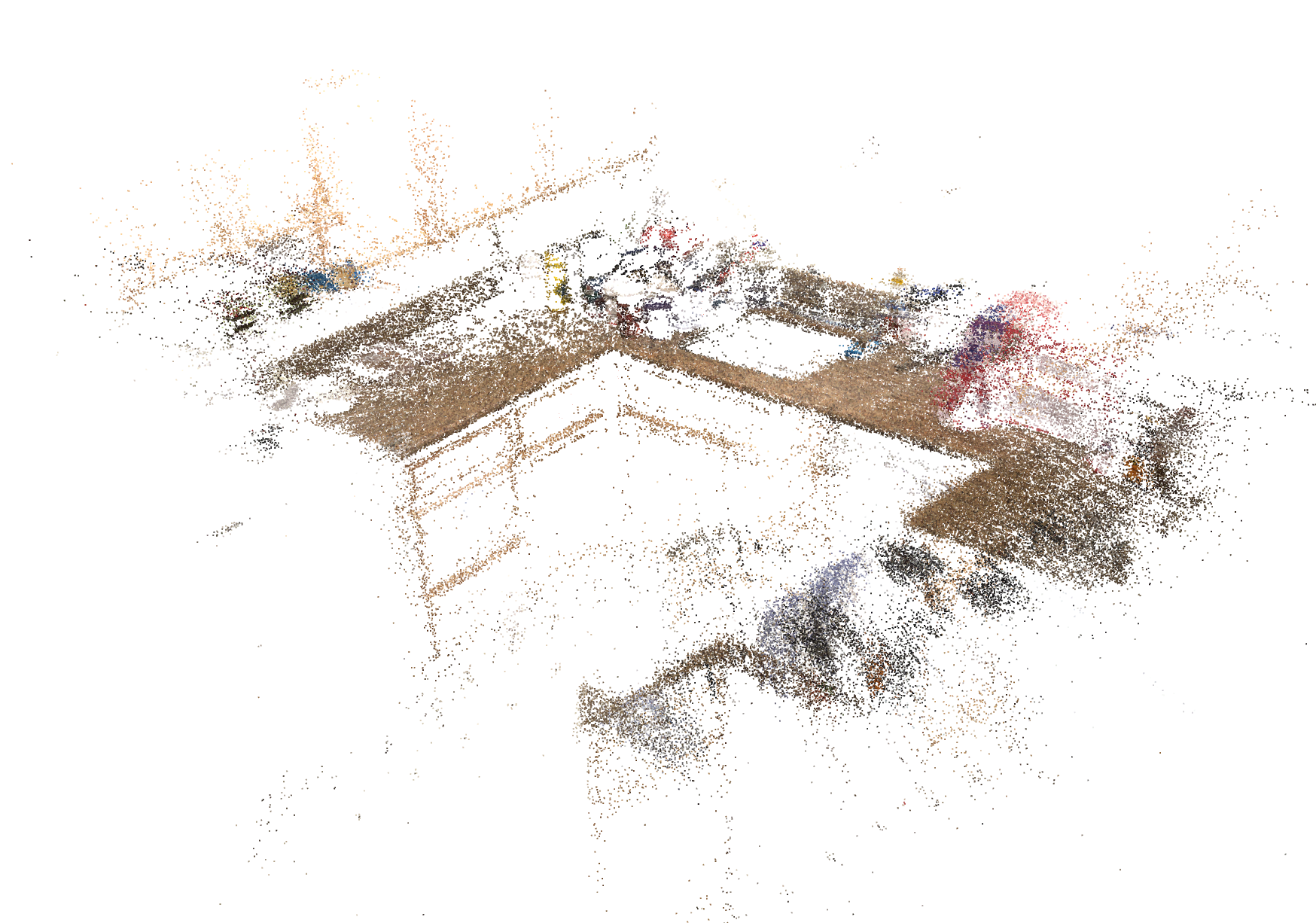}%
    \includegraphics[width=0.48\linewidth]{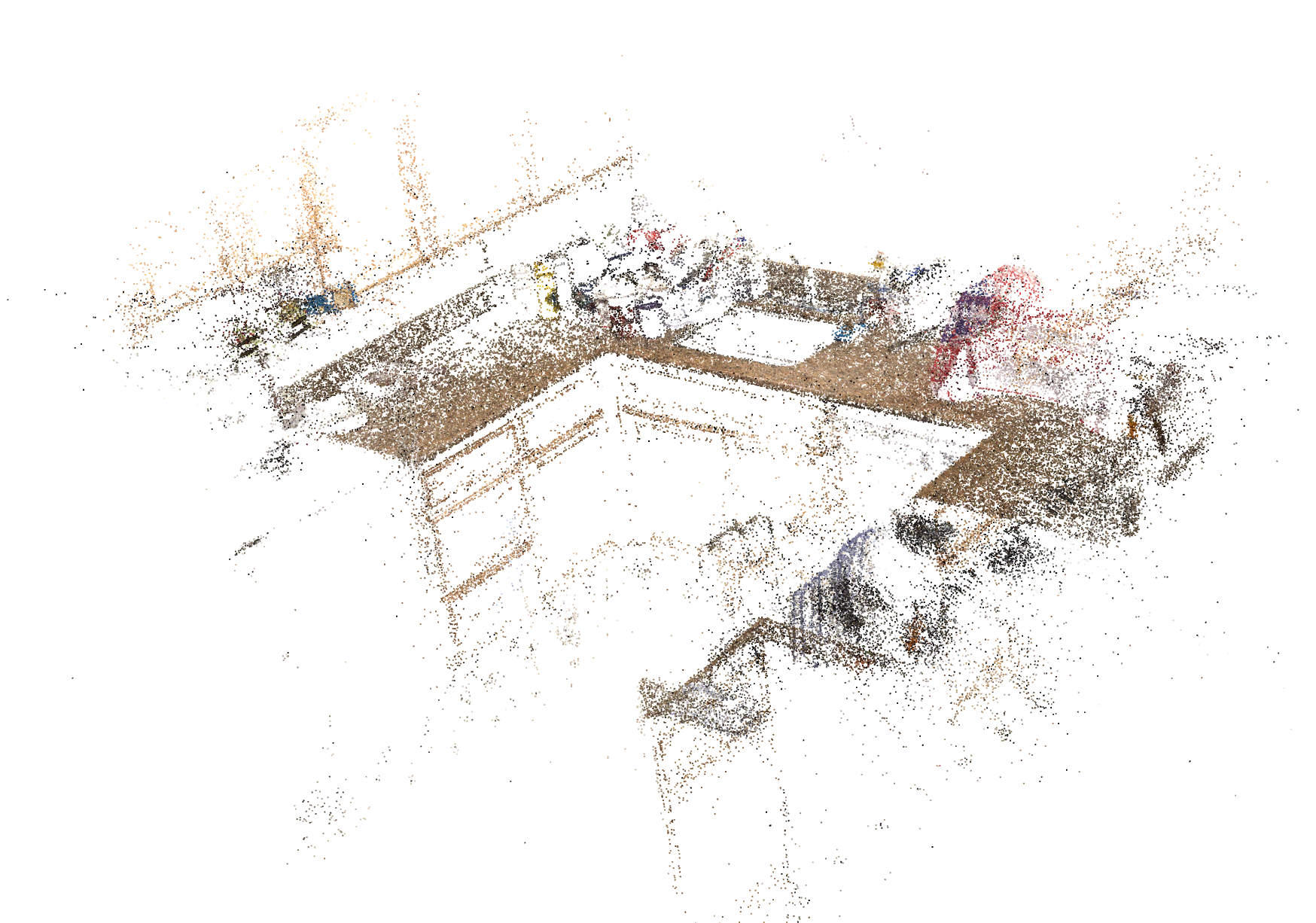}%
    \caption{\textbf{Visualisation of SfM point clouds} for Red Kitchen (7Scenes) and apt2/kitchen (12Scenes) obtained using the original 
    (left) and the SfM \GT (right) poses. The SfM \GT results in less noisy models where structures are more clearly visible.}
    \label{fig:viz_sfm}
\end{figure}

Tab.~\ref{tab:stats_sfm} shows the SfM metrics for 
both datasets. 
The SfM \GT 
clearly outperforms the original (RGB-)D SLAM \GT in the number of 
observations, track length, and reprojection error, especially on 7Scenes. 
We attribute this to the fact that KinectFusion, in contrast to BundleFusion, 
does not perform global optimisation and is thus susceptible to drift~\cite{Valentin20163DV}. 
Fig.~\ref{fig:viz_sfm} qualitatively compares the SfM point clouds obtained with both versions of the \GT, 
showing that the SfM \GT leads to significantly less noisy SfM points.

As way 
to measure 
the similarity of the local optima found by the different \GT algorithms, we generate an ``intermediate" \GT, denoted as +BA in Tab.~\ref{tab:stats_sfm}: 
starting from the original \GT, we alternate between BA of the triangulated 3D model and merging and completing 3D points. 
As for office2/5a and office2/5b, we repeat this process for 10 iterations. 
In case that the local minima found by the \mbox{(RGB-)D} SLAM and SfM algorithms are close-by, we expect this process to result in a similar local optimum for the SfM metrics.\footnote{We compare the ``intermediate" \GT to the other \GT based on SfM metrics rather than comparing pose errors. The alignment between SfM and SLAM \GT introduces a potential error that we cannot easily remove.}
As can be seen in Tab.~\ref{tab:stats_sfm}, the ``intermediate" \GT results in similar or slightly worse statistics compared to the SfM \GT for both datasets. 
This indicates that the difference between poses is not large enough for bundle adjustment to result in significantly different local minima. 

\begin{figure*}[!t]
    \centering
    \includegraphics[width=0.23\linewidth]{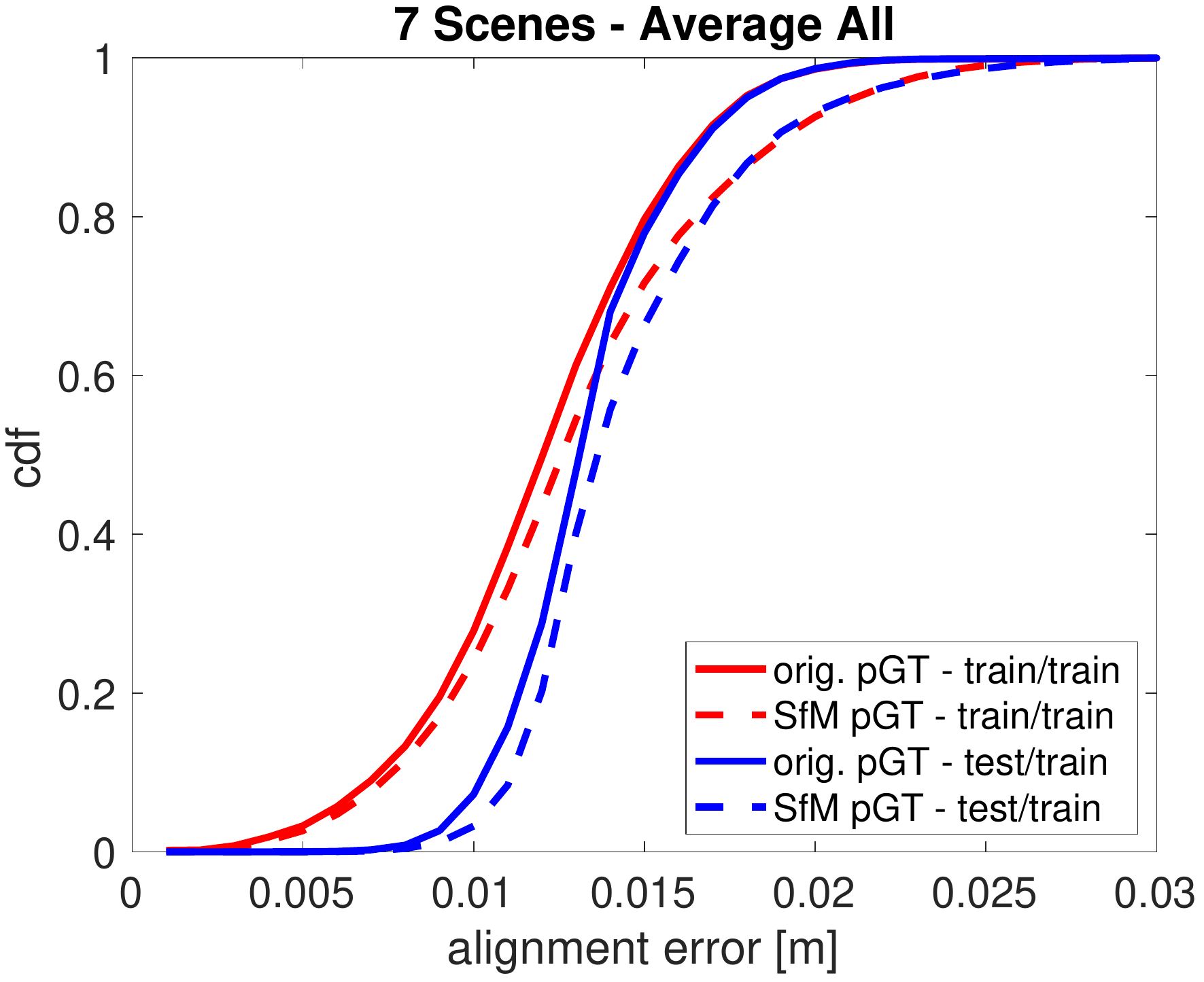}%
    \includegraphics[width=0.224\linewidth]{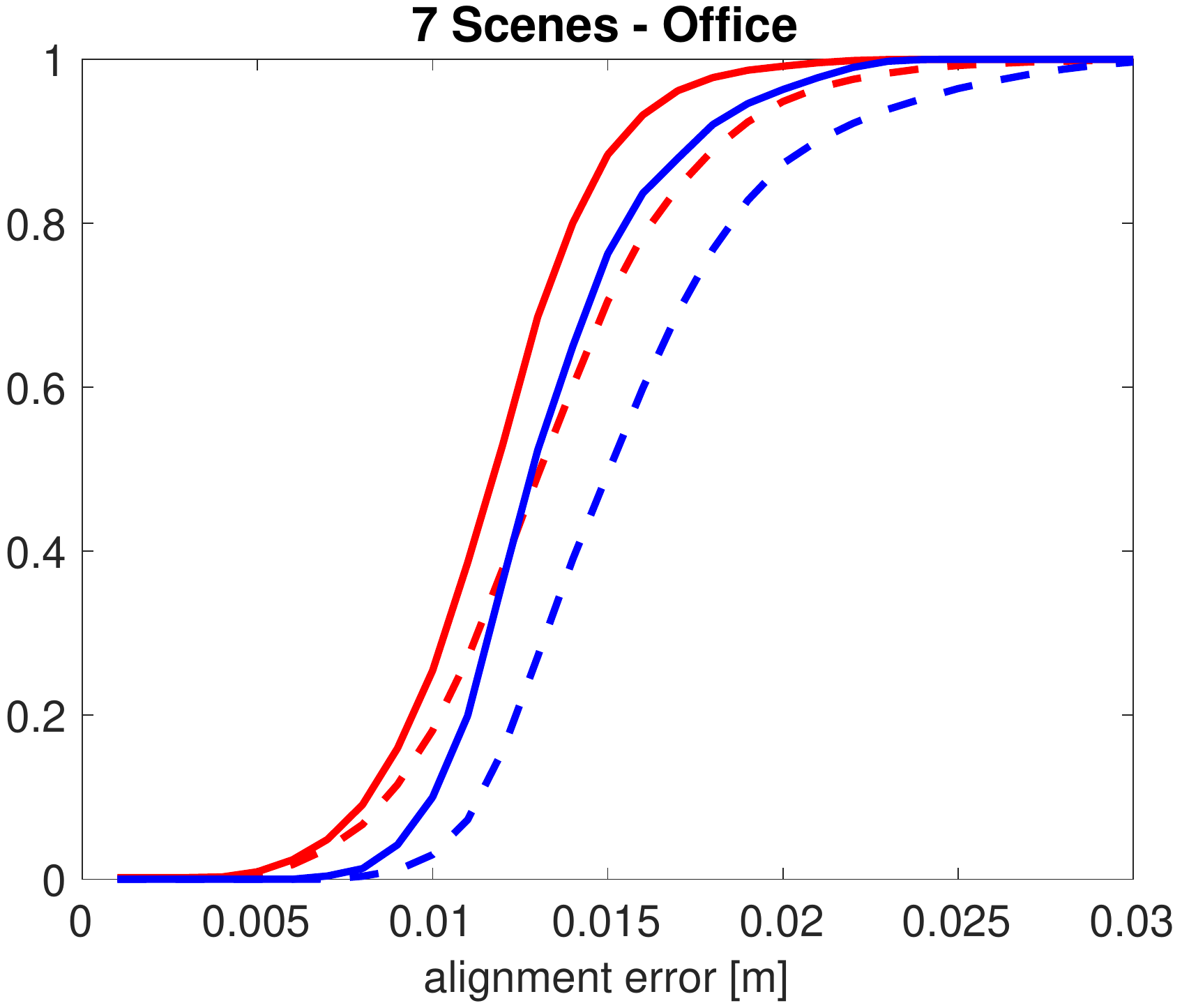}%
    \includegraphics[width=0.224\linewidth]{figures/alignment_12scenes_all.pdf}%
    \includegraphics[width=0.224\linewidth]{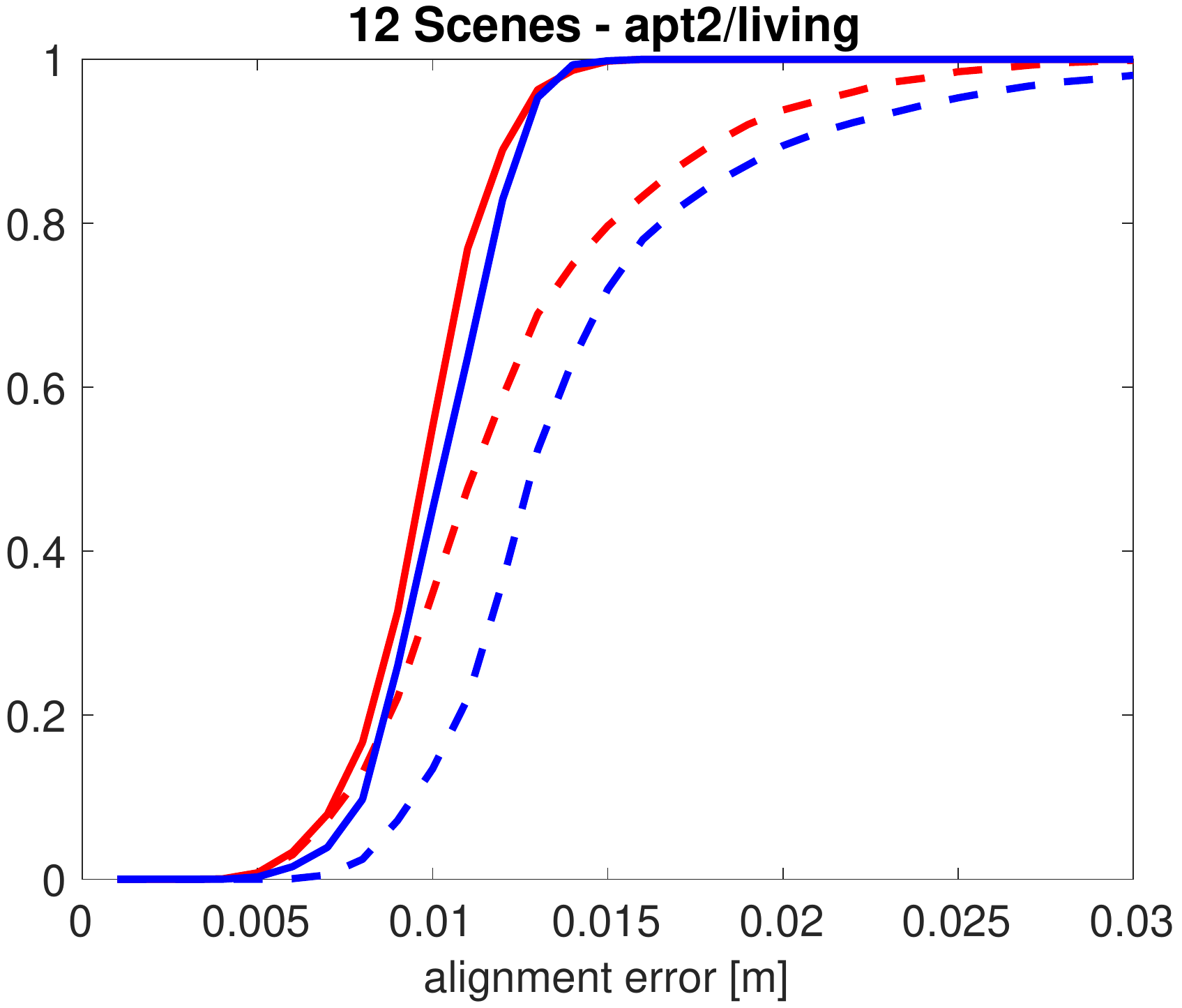}%
    \caption{\textbf{3D alignment statistics} for the 7Scenes and 12Scenes datasets. We show cumulative distributions (cdfs) of the 3D alignment errors between the depth maps of train/train and test/train image pairs with a visual overlap of at least 30\% for the original (RGB-)D SLAM and the SfM pseudo GT. As can be seen, the original pseudo GT produces more accurate alignments of the depth maps.}%
    \label{fig:stats_pc}
\end{figure*}

\PAR{Evaluation based on 3D alignment metrics.} 
We next evaluate how accurately the two \GT versions align the depth maps available for each image. 
For a pair of images $(A, B)$ in a scene, we use the \GT poses to transform their depth maps to 3D point clouds in scene coordinates.
For each 3D point in 
$A$'s depth map, we find the nearest point in 
$B$'s depth map. 
We report the root mean square error (RMSE) of all point correspondences below a 5cm outlier threshold.\footnote{We did not observe image pairs with no correspondences within 5cm.} 
This cost function, implemented in Open3D~\cite{Zhou2018ARXIV}, measures the 3D alignment of the two point clouds and  
replicates the metric minimised by 
algorithms such as KinectFusion~\cite{Newcombe2011ISMAR,Izadi2011UIST} and BundleFusion~\cite{Dai2017TOG}. 

We select image pairs for evaluation based on visual overlap in the SfM \GT~\cite{Radenovic2019PAMI}: 
Let $|P_{AB}|$ be the number of 3D points jointly observed by images $A$ and $B$ and $|P_A|$ and $|P_B|$ be the number of 3D points seen in
$A$ respectively $B$. 
We consider a pair if $|P_{AB}| / \max(|P_A|, |P_B|) \geq 0.3$. 

Fig.~\ref{fig:stats_pc} shows cumulative histograms over the alignment errors for both \GT versions (see Appendix \ref{sec:quantitative_comparisons} for plots of all individual scenes). 
We separately show curves for pairs of training images and pairs containing one test and one training image. 
The former measures the consistency between the training images and the latter measures how well the test images align with the training images. 
Since images are taken in continuous sequences, there are smaller changes in viewpoint between pairs of training images than for pairs containing training and test images. 
As such, there is a larger error for test/train pairs than for train/train pairs.

\begin{figure*}[!t]
    \centering
    \includegraphics[width=0.85\linewidth]{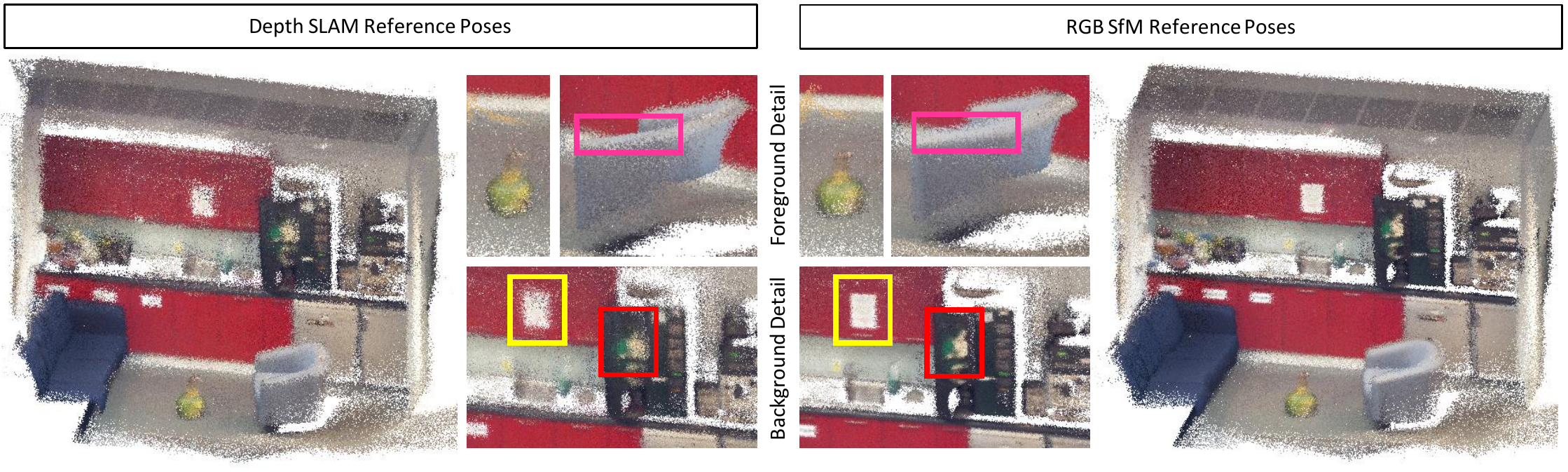}
    \caption{\textbf{Consistency of depth map fusion for 7Scenes \cite{Shotton2013CVPR}}. We reproject depth maps of the \emph{Pumpkin} scene and accumulate the resulting 3D point clouds in world space using the \GT reference poses. While the depth SLAM \GT leads to sharper reconstructions of foreground objects like the pumpkin and chair, SfM \GT leads to better alignment of distant objects like the paper note and coffee machine.}
    \label{fig:7sdrift}
\end{figure*}

\begin{figure*}[!t]
    \centering
    \includegraphics[width=0.85\linewidth]{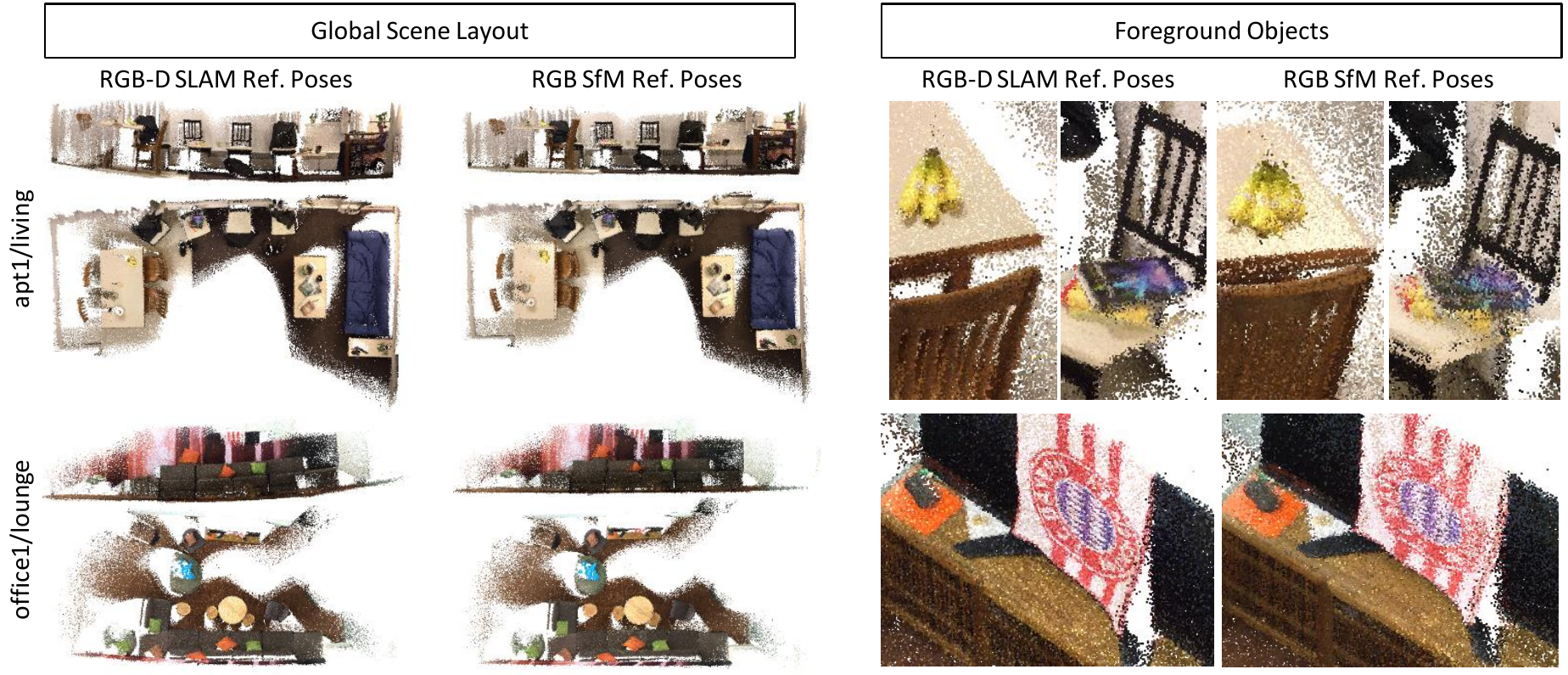}
    \caption{\textbf{Consistency of depth map fusion for 12Scenes \cite{Valentin20163DV}.} 
    \textbf{Left:} RGB-D SLAM pseudo GT leads to a significant warp of the global scene layout, which is clearly reduced by the SfM pseudo GT. \textbf{Right:} Foreground objects appear blurred for SfM pseudo GT poses compared to the sharper depth alignment generated by RGB-D SLAM \GT.}%
    \label{fig:viz_pc}
\end{figure*}

Fig.~\ref{fig:stats_pc} shows smaller alignment errors for the orig. \GT. 
We also show dense point clouds obtained by fusing individual depth maps using the different \GT poses in 
Fig.~\ref{fig:7sdrift} and Fig.~\ref{fig:viz_pc}. 
While the SfM \GT leads to globally more consistent geometry with less drift\footnote{Note that global consistency might not always be  necessary, \eg, AR applications where a user observes only a small part of the scene.}, fine details of foreground objects are better recovered with the original \GT. 
This confirms the results from Fig.~\ref{fig:stats_pc}, which show a more precise relative alignment of depth maps for the orig. \GT. 

\vspace{-6pt}
\section{Re-localisation Evaluation}
\vspace{-0.1cm}
\label{sec:reloc_eval}
Sec.~\ref{sec:comparison_pgt} showed that neither the original (RGB-)D SLAM \GT nor the SfM \GT is clearly better than the other. 
Thus, both \GT versions are valid choices for evaluating re-localisation algorithms. 
Their differences are in the order of centimeters.
However, this is typically the range used to measure localisation accuracy. 
This section thus investigates how different \GT versions affect the performance of re-localisers. 
We show that RGB-D baselines fare better on \GT generated with (RGB-)D SLAM. 
Baselines that minimise a reprojection error perform better on the SfM \GT.

\PAR{Evaluation measures.} 
We report the percentage of images localised within $X$cm and $X^\circ$ of the respective \GT~\cite{Shotton2013CVPR,Valentin20163DV}. 
We also report the Dense Correspondence Reprojection Error (DCRE)\cite{Wald2020ECCV}: 
for each test image, we back-project the depth map into a 3D point cloud using its \GT pose. 
We 
project each 3D point into the 
image using 
the estimated and the \GT pose and measure the 2D distance between both projections. 
We report the maximum DCRE per test image below and the mean DCRE per test image in Appendix \ref{sec:localization}.

\PAR{Baselines.} We evaluate classical, feature-based as well as learning-based re-localisers on the two versions of the \GT. 
Learning-based methods are re-trained on each \GT, and feature-based methods use each version of the \GT to create their map.
Please see Appendix \ref{sec:implementation_details} for details.

\colorlet{clr1}{green!100}
\colorlet{clr2}{green!66!yellow}
\colorlet{clr3}{green!33!yellow}
\colorlet{clr4}{yellow}
\colorlet{clr5}{yellow!66!orange}
\colorlet{clr6}{yellow!33!orange}

\definecolor{likesfm}{RGB}{197,224,180}
\definecolor{inter}{RGB}{222,235,247}
\definecolor{likeslam}{RGB}{250,190,212}

\begin{table*}[th]
\vspace{-0.2cm}
\begin{center}
\setlength{\tabcolsep}{1.8pt}
\scriptsize{
\begin{tabular}{|r|c|c|c||c|c|c||c|c|c||c|c|c||c|c|c||c|c|c||c|c|c||c|c|c|}
\cline{1-25} 

& \multicolumn{3}{c||}{Chess} & \multicolumn{3}{c||}{Fire} & \multicolumn{3}{c||}{Heads} & \multicolumn{3}{c||}{Office} & \multicolumn{3}{c||}{Pumpkin} & \multicolumn{3}{c||}{Red Kitchen} & \multicolumn{3}{c||}{Stairs}& \multicolumn{3}{c|}{\textbf{Average}}\\ \cline{2-25}
Pseudo GT & orig. & +BA & SfM & orig. & +BA & SfM & orig. & +BA & SfM & orig. & +BA & SfM & orig. & +BA & SfM & orig. & +BA & SfM & orig. & +BA & SfM & orig. & +BA & SfM \\ \hline

\cellcolor{likesfm} Active Search & 
86.4 & 98.6 & 99.9 &
86.3 & 96.6 & \textbf{99.8} &
95.7 & 96.5 & \textbf{100} &
65.6 & 76.7 & 98.6 &
34.1 & 44.9 & 99.6 &
45.1 & 61.1 & \textbf{99.8} &
67.8 & 82.0 & 91.9 &
\cellcolor{clr6} 68.7 & \cellcolor{clr6} 79.5 & \cellcolor{clr1} \textbf{98.5} \\ \hline

\cellcolor{likesfm} hLoc & 
94.2 & 99.6 & \textbf{100} &
93.7 & 97.0 & 99.4 &
99.7 & 99.9 & \textbf{100} &
83.2 & 88.7 & \textbf{100} &
55.2 & 65.5 & \textbf{100} &
61.9 & 72.7 & 98.6 &
49.4 & 58.1 & 72.0 &
\cellcolor{clr5} 76.8 & \cellcolor{clr4} 83.1 & \cellcolor{clr3} 95.7 \\ \hline

\cellcolor{likesfm} DVLAD+R2D2 & 
94.0 & 97.6 & \textbf{100} &
95.3 & 94.8 & 99.1 &
95.6 & 95.6 & 97.0 &
78.8 & 84.1 & 99.7 &
59.2 & 69.5 & 98.8 &
61.2 & 69.6 & 98.4 &
59.2 & 69.6 & 76.9 &
\cellcolor{clr4} 77.6 & \cellcolor{clr5} 83.0 & \cellcolor{clr3} 95.7 \\ \hline

\cellcolor{inter} DVLAD+R2D2(+D) & 
93.1 & 98.2 & \textbf{100} &
91.3 & 95.3 & 98.5 &
96.5 & 96.4 & 96.3 &
81.2 & 86.1 & 96.6 &
58.5 & 68.0 & 91.6 &
72.6 & 78.5 & 97.8 &
68.0 & 72.8 & 69.5 &
\cellcolor{clr3} 80.2 & \cellcolor{clr3} 85.0 & \cellcolor{clr5} 92.9 \\ \hline

\cellcolor{inter} DSAC* & 
97.8 & 99.2 & 99.9 & 
94.5 & \textbf{98.7} & 98.9 & 
98.8 & 99.6 & 99.8 & 
83.9 & 89.8 & 98.1 & 
62.0 & 73.9 & 99.0 & 
65.5 & 79.1 & 97.0 & 
77.7 & \textbf{91.5} & \textbf{92.0} & 
\cellcolor{clr2} 82.9 & \cellcolor{clr2} 90.3 & \cellcolor{clr2} 97.8 \\ \hline

\cellcolor{likeslam} DSAC*(+D) & 
\textbf{99.4} &\textbf{ 99.7} & 99.6 & 
\textbf{98.9} & 98.4 & 96.9 & 
\textbf{99.9} & \textbf{100} & 99.5 & 
\textbf{98.9} & \textbf{96.4} & 95.3 & 
\textbf{80.9} & \textbf{77.6} & 90.9 & 
\textbf{92.4} & \textbf{94.9} & 96.4 & 
\textbf{92.6} & 85.5 & 88.4 &
\cellcolor{clr1} \textbf{94.7} & \cellcolor{clr1} \textbf{93.2} & \cellcolor{clr4} 95.3\\ \hline

\end{tabular}
}
\end{center}
\vspace{-6pt}
\caption{\textbf{Re-localisation results on 7Scenes~\cite{Shotton2013CVPR}} as the percentage of images localised within 
5cm and 5$^\circ$ of the original \GT generated by Depth~SLAM (\emph{orig.}), local SfM minima obtained by fixing the orig.~\GT training poses and optimising the test poses (\emph{+BA}; see Sec.~\ref{sec:reloc_eval} for details), and the SfM \GT (\emph{SfM}). We visualise the ranking of methods from \colorbox{clr1}{best}  to \colorbox{clr6}{worst} depending on the \GT. 
We color-code methods based on their similarity to the reference algorithm: \colorbox{likesfm}{similar to SfM}, \colorbox{likeslam}{similar to D-SLAM}, or \colorbox{inter}{intermediary}.}
\label{tab:results_7scene}%
\end{table*}

\noindent \textbf{DSAC*}~\cite{Brachmann2017CVPR, Brachmann2018CVPR,  brachmann2020ARXIV} is a learning-based scene coordinate regression approach, where a neural network predicts for each pixel the corresponding 3D point in scene space.
DSAC* uses a PnP\cite{gao2003complete} solver and RANSAC\cite{Fischler81CACM} on top of the \mbox{2D-3D} matches.
Its RGB-D variant, \textbf{DSAC* (+D)}, uses image depth to establish 3D-3D matches and a Kabsch\cite{kabsch1976solution} solver. 
\noindent \textbf{hLoc}~\cite{Sarlin2019CVPR} combines image retrieval with  SuperPoint~\cite{DeTone2018CVPRWorkshops} features and  SuperGlue~\cite{Sarlin2020CVPR} for matching, followed by P3P+RANSAC-based pose estimation. 
\noindent \textbf{DenseVLAD+R2D2}~\cite{Torii2015CVPR,revaud2019r2d2,HumenbergerX20Kapture} uses DenseVLAD~\cite{Torii2015CVPR} for retrieving image pairs and R2D2 features for matching. 
The training images and poses are used to construct a 3D SfM map, and test images are localised using 2D-3D matches and P3P+RANSAC. 
Instead of triangulating point matches, \textbf{DenseVLAD+R2D2 (+D)} constructs the 3D map by projecting R2D2 keypoints to 3D space using depth maps.
\noindent \textbf{Active Search (AS)}~\cite{Sattler2012ECCV, Sattler2017PAMI} is a classical feature-based approach that establishes 2D-3D correspondences based on prioritized SIFT~\cite{Lowe04IJCV} matching. 
AS estimates the camera pose with a P3P solver~\cite{Kneip11CVPR,Haralick94IJCV} inside a RANSAC loop~\cite{Fischler81CACM}. 

\PAR{Results.} Tab.~\ref{tab:results_7scene} reports the percentage of test images localised within 5cm and 5$^\circ$ of the \GT for the 7Scenes dataset. 
For the original \GT, depth-based \emph{DSAC* (+D)} clearly outperforms all other methods. 
Depth-based \emph{DenseVLAD+R2D2 (+D)} achieves the best results among all sparse feature-based methods. 
\emph{AS}, using classical SIFT features, achieves the lowest accuracy using the original \GT. 

\begin{figure*}[!t]
    \centering
    \includegraphics[width=1\linewidth]{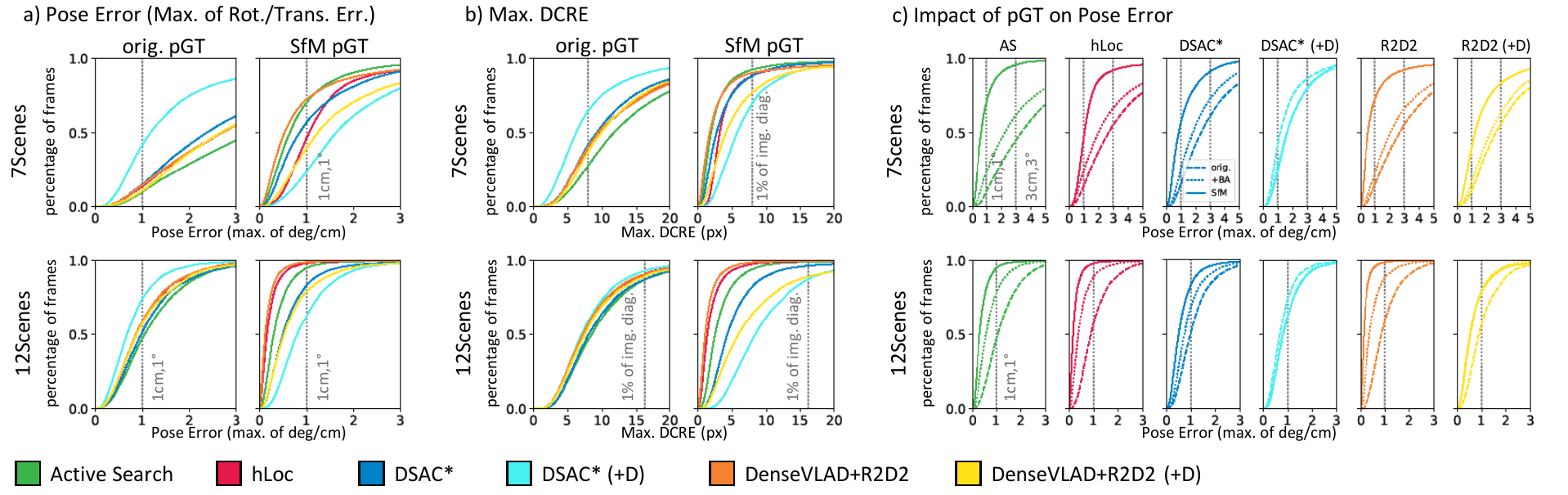}
    \caption{\textbf{Pose error and Dense Correspondence Reprojection Error (DCRE)\cite{Wald2020ECCV}. a) }
    Cum.~distributions of pose error (max.~of rotation and position error) for 7Scenes (top row) and 12Scenes (bottom row), averaged across scenes. 
    \textbf{b)} Cum.~distribution of the maximum DCRE per test image.
    We mark 1\% of the image diagonal in the DCRE plots.
    The relative ordering of the methods changes when switching between {\GT}s. 
    \textbf{c)} Changes in the cum.~pose error per method and \GT (
    orig.~\GT, locally bundle-adjusted \GT (+BA), and SfM \GT). 
    }
    \label{fig:acc_repro}
\end{figure*}

The ranking changes drastically using the SfM \GT.
\emph{AS} jumps from last to first place with an absolute difference of +29.8 in pose accuracy, outperforming all learning-based and depth-based competitors. 
Particularly notable are the results on Pumpkin and Red Kitchen, where AS improves from localising less than 50\% within the 5cm,  5$^\circ$ threshold to localizing more than 99\% of the images. 
For both scenes, Tab.~\ref{tab:stats_sfm} shows a significant difference in the SfM statistics between the two \GT versions.
See Appendix \ref{sec:visualizations} for a visual analysis on Pumpkin.
The previously-leading depth-based \emph{DSAC* (+D)} and \emph{DenseVLAD+R2D2 (+D)} drop to the last places of the ranking. 
Both methods are outperformed by their 
RGB-only counterparts when using the SfM \GT.

We can correlate these observations with each method's similarity to the respective reference algorithm (\cf column 1 of Tab.~\ref{tab:results_7scene} for a coarse classification). 
We regard methods that optimise a reprojection error over sparse features as \emph{similar to SfM} and 
methods that optimise a dense 3D-3D error 
as \emph{similar to (RGB-)D SLAM}.
The RGB variant of \emph{DSAC*} optimises a dense reprojection error. 
\emph{DVLAD+R2D2(+D)} optimises a sparse reprojection error but incorporates depth when building the 3D map. 
Thus, we classify those two methods as \emph{intermediary}.
Among methods \emph{similar to SfM}, 
AS shows the largest improvement under the SfM \GT as it re-uses the SIFT features from SfM. 
Fig.~\ref{fig:acc_repro}(a) shows cumulative distributions over the fraction of images localised within Xcm, X$^\circ$ of the \GT for tighter thresholds than used before. 
This is particularly interesting for
12Scenes, 
where the accuracy of all methods saturates under the 5cm, 5$^\circ$ threshold.
Poses predicted by \emph{DSAC* (+D)} better align with the original 
(RGB-)D SLAM \GT than with the SfM \GT. 
At the same time, poses predicted by RGB-based methods better align with the SfM \GT. 
There are larger differences 
between the methods for finer thresholds. 
For 12Scenes, \emph{hLoc} and \emph{DenseVLAD+R2D2} achieve the highest accuracy under a 1cm, 1$^\circ$ threshold.

Fig.~\ref{fig:acc_repro}(b) shows cumulative distributions for the max.~DCRE. 
Since the DCRE depends on the pose accuracy, we observe the same behavior as before, \ie, methods more similar to SfM 
outperform depth-based methods on the SfM \GT while performing worse on the original (RGB-D) \GT. 
Yet, this does not necessarily imply that such 
methods are 
superior. 
They closely resemble the SfM ref.~algorithm, and they use the 3D points triangulated by the SfM pipeline from the training images for pose estimation. 
Thus, it seems likely that feature-based methods ``overfit" to the SfM \GT by being able to closely replicate SfM behavior. 
To further illustrate the issue, we created an ``intermediate" \GT: 
starting with the original \GT poses, we triangulate the scene and use bundle adjustment followed by point merging to optimise the test poses while keeping the training poses fixed. 
Intuitively, the resulting poses, denoted as ``+BA", approximate the ``optimal" test poses for the original training image \GT under the reprojection error metric. 
Fig.~\ref{fig:acc_repro}(c) and 
Tab.~\ref{tab:results_7scene} show results obtained using the +BA test poses. 
The +BA poses significantly improve the evaluation scores of RGB-based methods such as AS, but less so for depth-based methods such as DSAC* (+D) or DenseVLAD+R2D2 (+D). 
The closer the \GT is to the cost function optimised by both methods, the better they perform. 
In contrast, depth-based methods typically either perform similar or worse under these \GT poses. 
Our results indicate that learning-based methods might have some capacity to adjust to
the \GT since {DSAC*} ranks well across all \GT versions. 
Still, {DSAC*} is always outperformed by methods more similar to the ref. algorithm.

\vspace{-0.2cm}
\section{Conclusion}
\vspace{-0.2cm}

Re-localisation benchmarks usually rely on a reference algorithm to create pseudo ground truth for evaluation.  
As such, 
they do not measure absolute pose accuracy but rather how well a given method is able to reproduce the reference output.
Our paper points out an important implication: 
different cost functions optimised by reference algorithms lead to different local minima. 
This affects re-localisation evaluation as methods that optimise a similar cost function as the reference algorithm 
better replicate the local minima and imperfections of the \GT, to a degree that relative rankings can be (nearly) completely inverted. 
This issue is fundamental, and we do not see a solution to this problem. 
However, there are ways to address the issue, as shown in Sec.~\ref{sec:reloc_eval}: 
new benchmark datasets could provide multiple pGT versions to enable a more concise evaluation that takes the impact of the \GT into account. 
\Eg, although DSAC* does not perform \emph{best} under any \GT, it performs \emph{well} under all \GT versions. 
If multiple \GT versions are not available, 
localisation algorithms can be grouped based on their similarity to the reference algorithm (\cf color-coding in Tab.~\ref{tab:results_7scene}) and only be compared within but not between groups. 
Another approach is to choose 
evaluation thresholds that are large enough that the difference in pGT will not affect the measured performance, \eg, 5cm, 5$^\circ$ for 12Scenes. 
Such an approach will likely have to explicitly account for the uncertainties in the estimated poses, which itself is a complex problem~\cite{Foerstner2016Book}. 
Still, knowledge about pose uncertainties would allow us to determine when a dataset is solved. 
Another direction is a task-specific evaluation of re-localisation methods, \eg, measuring their performance in the context of AR, robotic navigation, \etc 
Again, understanding the impact of the \GT on such evaluations is an interesting and open problem.

{
\small{
\PAR{Acknowledgements.} This work has received funding from the EU Horizon 2020 project RICAIP (grant agreement No 857306) and the European Regional Development Fund under project IMPACT (No.~CZ.02.1.01/0.0/0.0/15$\_$003/0000468).} 
}

\appendix

\section{Implementation Details}
\label{sec:implementation_details}
In the following, we detail how we adjusted the source code 
of Active Search~\cite{Sattler2017PAMI}, hLoc~\cite{Sarlin2019CVPR,Sarlin2020CVPR}, R2D2~\cite{HumenbergerX20Kapture}, and DSAC*~\cite{brachmann2020ARXIV} and provide training details for the latter.

\subsection{Active Search (AS)}

We use the source code 
of~\cite{Sattler2017PAMI}, but replace the original RANSAC method with the LO-RANSAC~\cite{Lebeda2012BMVC} implementation from~\cite{Sattler2019Github}. 
Local optimisation is implemented by minimising the sum of squared reprojection errors over a subset of the inliers of the best pose found so far. 
In addition, we perform non-linear optimisation of the pose by minimising the sum of squared reprojection errors over all inliers after LO-RANSAC. 
In both cases, Ceres~\cite{ceres-solver} is used to implement the optimisation. 
Based on preliminary experiments, both modifications significantly improve performance. 

We set the inlier threshold for LO-RANSAC to 1\% of the image diagonal\footnote{While we observed better results when tuning the threshold per scene, we want to avoid overfitting to the test set and thus use the same setting for all scenes.} and use 10k visual words trained on an unrelated outdoor dataset for prioritization. 
For the SfM \GT, which provides an estimate of the radial distortion of the test images, we undistort the SIFT~\cite{Lowe04IJCV} feature positions in the test images before RANSAC-based pose estimation. 

AS requires a SfM model of the scene for 2D-3D matching. 
We use COLMAP to build these models by triangulating the 3D structure of the scene from the known \GT poses of the training images. 
To establish the matches required for triangulation, we use COLMAP's image retrieval pipeline~\cite{Schoenberger2016ACCV} to match each training image against the top-100 retrieved other training images. 
In addition, we match each training image against each other training image that has a \GT pose difference below 2m and 45$^\circ$. 
For the original \GT of 7Scenes, we obtained better results by relaxing the thresholds COLMAP uses for triangulation. 
We account for the transformation between the depth and the RGB cameras when building the SfM models for the original 7Scenes \GT.\footnote{Note that the SfM \GT directly provides poses for the RGB images and it is not necessary to account for the transformation.}

\subsection{Hierarchical Localization (hLoc)}

Similar to Active Search, hLoc~\cite{Sarlin2019CVPR,Sarlin2020CVPR} is based on local features. 
Whereas Active Search relies on SIFT~\cite{Lowe04IJCV}, hLoc employs SuperPoint features~\cite{DeTone2018CVPRWorkshops}, a modern learned alternative. 
Active Search directly matches features extracted from the test image against descriptors associated with the 3D points. 
In contrast, hLoc first employs an image retrieval stage to identify a set of training images that potentially show the same part of the scene as the test image. 
The features found in the test image are then only matched against the 3D points visible in the top-$k$ retrieved training images. 
For matching, the SuperGlue~\cite{Sarlin2020CVPR} approach is used to improve matching quality. 
The resulting 2D-3D matches are then used to estimate the camera pose by applying a P3P solver inside a RANSAC loop. 

We use the source made publicly available by the authors and use their default settings. 
While the original publication describing the hierarchical localization pipeline~\cite{Sarlin2019CVPR} uses NetVLAD~\cite{Arandjelovic2016CVPR} descriptors, we use DenseVLAD~\cite{Torii2015CVPR} descriptors instead. 
DenseVLAD is a non-learned alternative to NetVLAD, where densely extracted RootSIFT~\cite{Arandjelovic2012CVPR} features are pooled into a VLAD~\cite{Jegou-CVPR10} descriptor. 
We chose DenseVLAD as it, in our experience, performs better for the 7Scenes and 12Scenes datasets than NetVLAD and use the top-20 retrieved images. 

\subsection{DenseVLAD+R2D2}

DenseVLAD+R2D2~\cite{HumenbergerX20Kapture} follows the workflow of image retrieval as well as structure-based methods where, first, the most similar training images are retrieved using global image representations, and second, these image pairs are used for local feature matching.
Same as for our hLoc experiments (we use exactly the same retrieval results), for image retrieval during localisation, we use DenseVLAD~\cite{Torii2015CVPR} features and for mapping, we use a list for matching training images that was obtained as a result of finding co-observations of reconstructed 3D points (using the AS map as basis).
For local feature matching, and in addition to Active Search and hLoc (which use SIFT resp.~SuperPoint), here we use R2D2~\cite{revaud2019r2d2} features.
DenseVLAD+R2D2 uses COLMAP for both, 3D point triangulation of the map and image registration using 2D-3D correspondences.
The matches are obtained using the nearest neighbors in descriptor space (L2-norm), cross-validation, and geometric verification.

Instead of triangulating keypoint matches using the camera poses, for \textbf{DenseVLAD+R2D2 (+D)}, we construct the 3D map by projecting the keypoints to 3D space using the provided and registered~\cite{wolf2014CVIU} depth maps.
For localization, we follow the same method as described above.

\subsection{DSAC*}

We use the public code of DSAC* \cite{brachmann2020ARXIV} with default parameters.
DSAC* supports different training modes utilising varying degrees of supervision.
To achieve best results, we follow Brachmann and Rother \cite{brachmann2020ARXIV} and initialize the DSAC* network using scene coordinate ground truth.
Brachmann and Rother render ground truth scene coordinates using 3D models of each scene provided in the 7Scenes and 12Scenes datasets, respectively. 
Next to the pseudo ground truth camera poses of these datasets, the 3D models are an additional output of (RGB-)D SLAM.
Hence, these 3D models would add an additional, non-trivial dependency of DSAC* training to the underlying dataset reference algorithm.
To restrict the influence of the reference algorithm to \GT poses alone, we train DSAC* using ground truth scene coordinates that we obtain from the measured depth map of each image.
We backproject the depth map to 3D using the camera calibration parameters, and transform them to scene space using the \GT pose.
For 7Scenes, we manually register depth maps to RGB images using the calibration parameters provided by \cite{wolf2014CVIU}.

Since the DSAC* code does not support a camera model with radial distortion, we instead undistort RGB images using COLMAP before passing it to the DSAC* pipeline.
We only do this for experiments with the SfM \GT since the (RGB-)D SLAM \GT assumes zero radial distortion.

We follow Brachmann and Rother \cite{brachmann2020ARXIV} and train DSAC* for 1.1M iterations (initialization + end-to-end). 
This took approximately 16 hours per scene on a GeForce RTX 2080 Ti.
Compared to the results published in \cite{brachmann2020ARXIV}, we observe slightly reduced accuracy, \eg 82.9\% versus 85.2\% for DSAC* (RGB) on 7Scenes (averaged over all scenes). 
We attribute this slight difference to our use of measured depth maps in the initialization training stage, which are more noisy and contain holes as well as large areas of invalid depth compared to the rendered ground truth scene coordinates used in \cite{brachmann2020ARXIV}.

\section{Visual Comparisons of pGT}
\label{sec:visualizations}

\begin{figure*}[!t]
    \centering
    \includegraphics[width=1\linewidth]{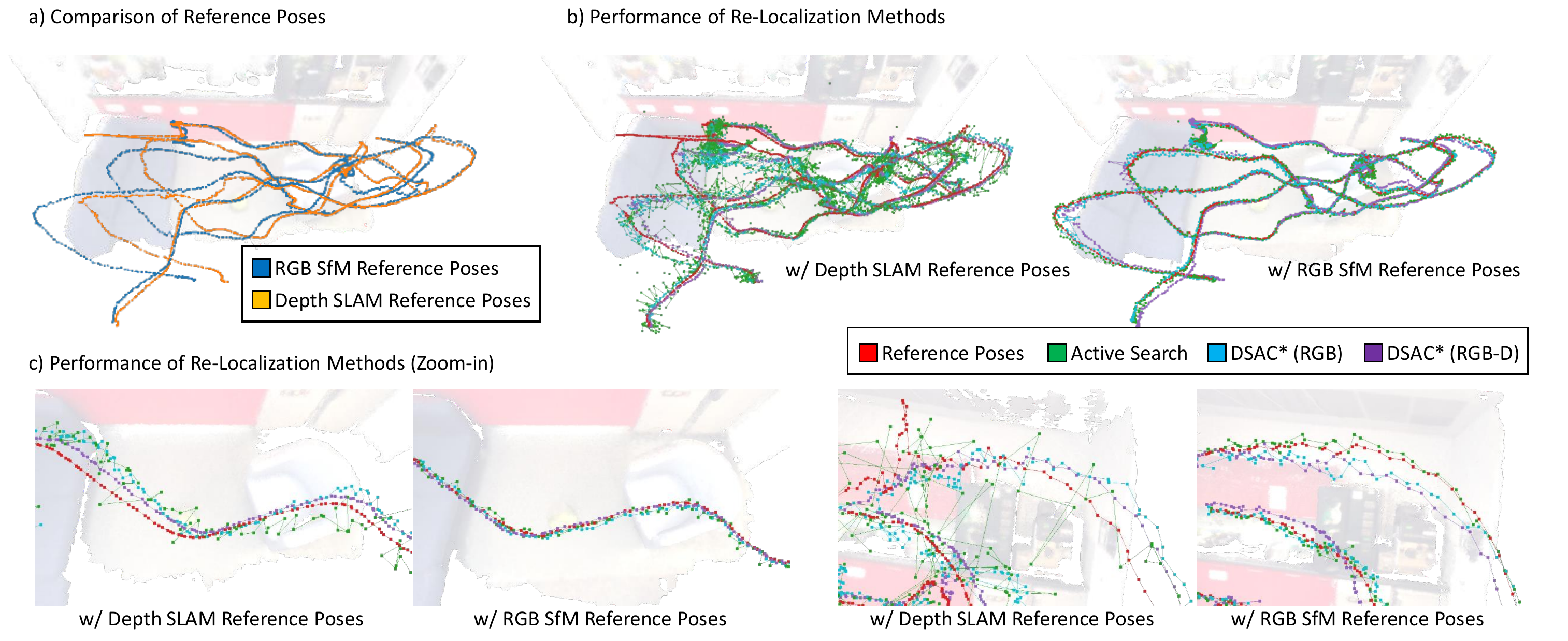}
    \caption{\textbf{Comparison of \GT on 7Scenes Pumpkin. a)} SfM-based reference poses (blue) versus SLAM-based reference poses (orange). \textbf{b)} We show estimated camera positions for Active Search (green), DSAC* w/ RGB inputs (cyan) and DSAC* w/ RGB-D inputs (purple). The respective reference poses, SLAM \GT on the left and SfM \GT on the right, are shown in red. We show close-up views in \textbf{c)}. }
    \label{fig:7s_pumpkin}
\end{figure*}

We plot depth-based SLAM \GT versus RGB-based SfM \GT for the Pumpkin scene of 7Scenes in Fig.~\ref{fig:7s_pumpkin}. 
For this scene, we observe the largest visual drift between both versions of the \GT.
We also show estimated camera trajectories for Active Search, DSAC* and \mbox{DSAC* (+D)}, the top-performing methods depending on the \GT version, for both versions of the \GT. 
While the depth-based SLAM \GT on this scene seems to have defects that make it hard for all re-localization methods to follow the ground truth trajectory, results look smoother for the SfM \GT. 
Still, both DSAC* re-localizers fail to follow the SfM \GT exactly, exhibiting small, consistent offsets \wrt the pseudo ground truth trajectory. 
We observe similar, yet less pronounced, patterns for other scenes of 7Scenes and the scenes of 12Scenes.

\begin{figure*}[!t]
    \centering
    \includegraphics[width=0.315\linewidth]{figures/alignment_7scenes_all.pdf}%
    \includegraphics[width=0.3\linewidth]{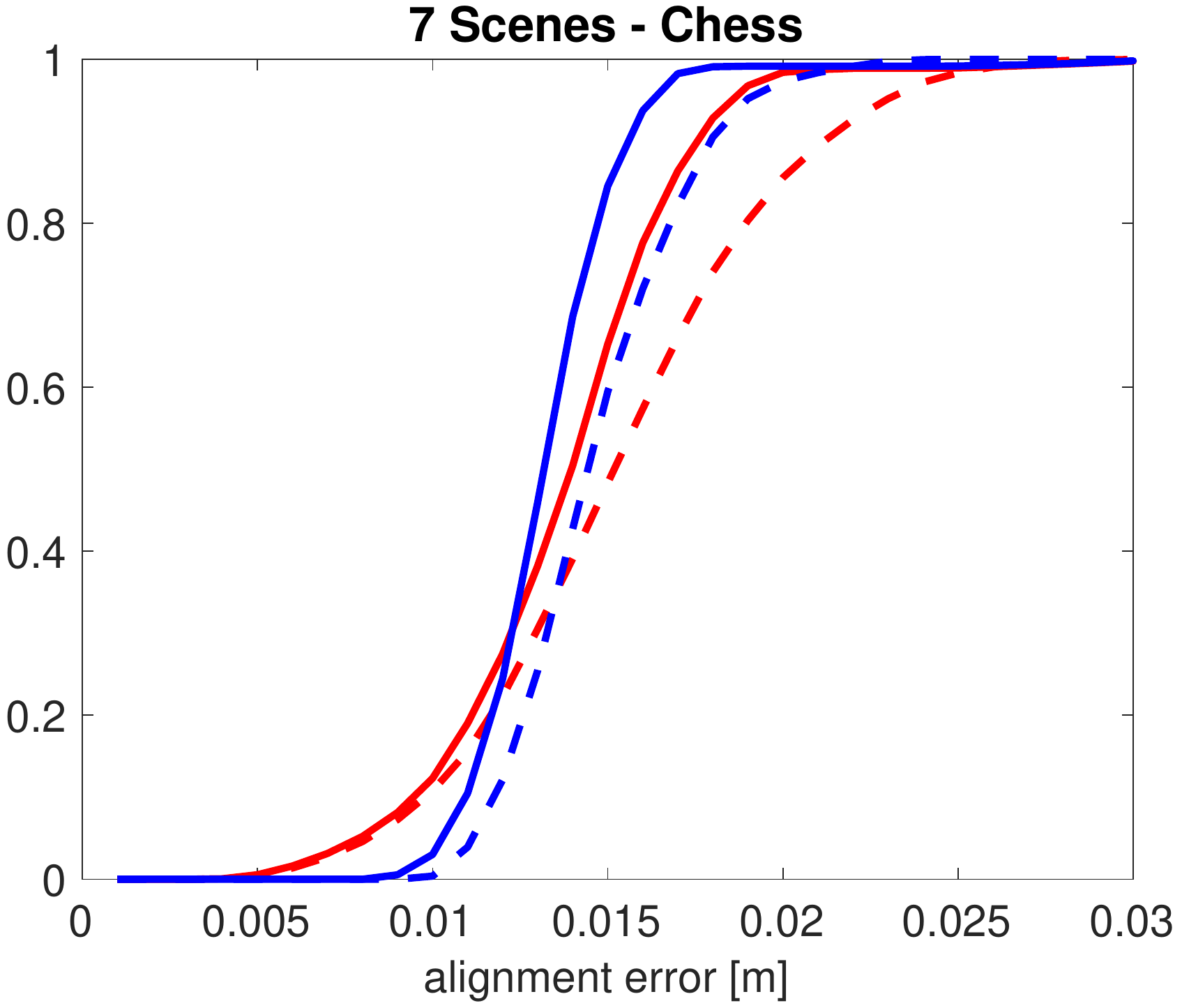}%
    \includegraphics[width=0.3\linewidth]{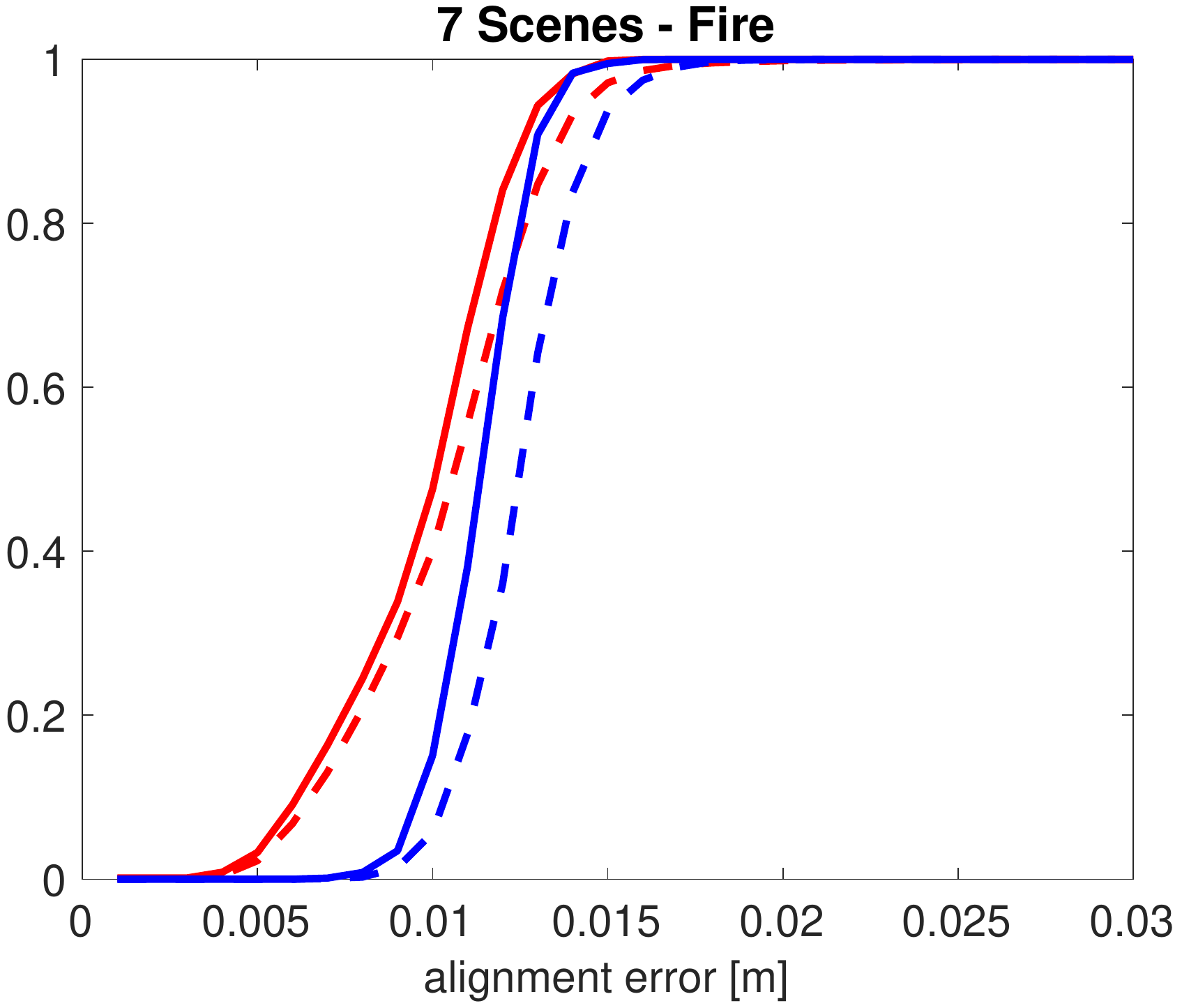}\\%
    \includegraphics[width=0.3\linewidth]{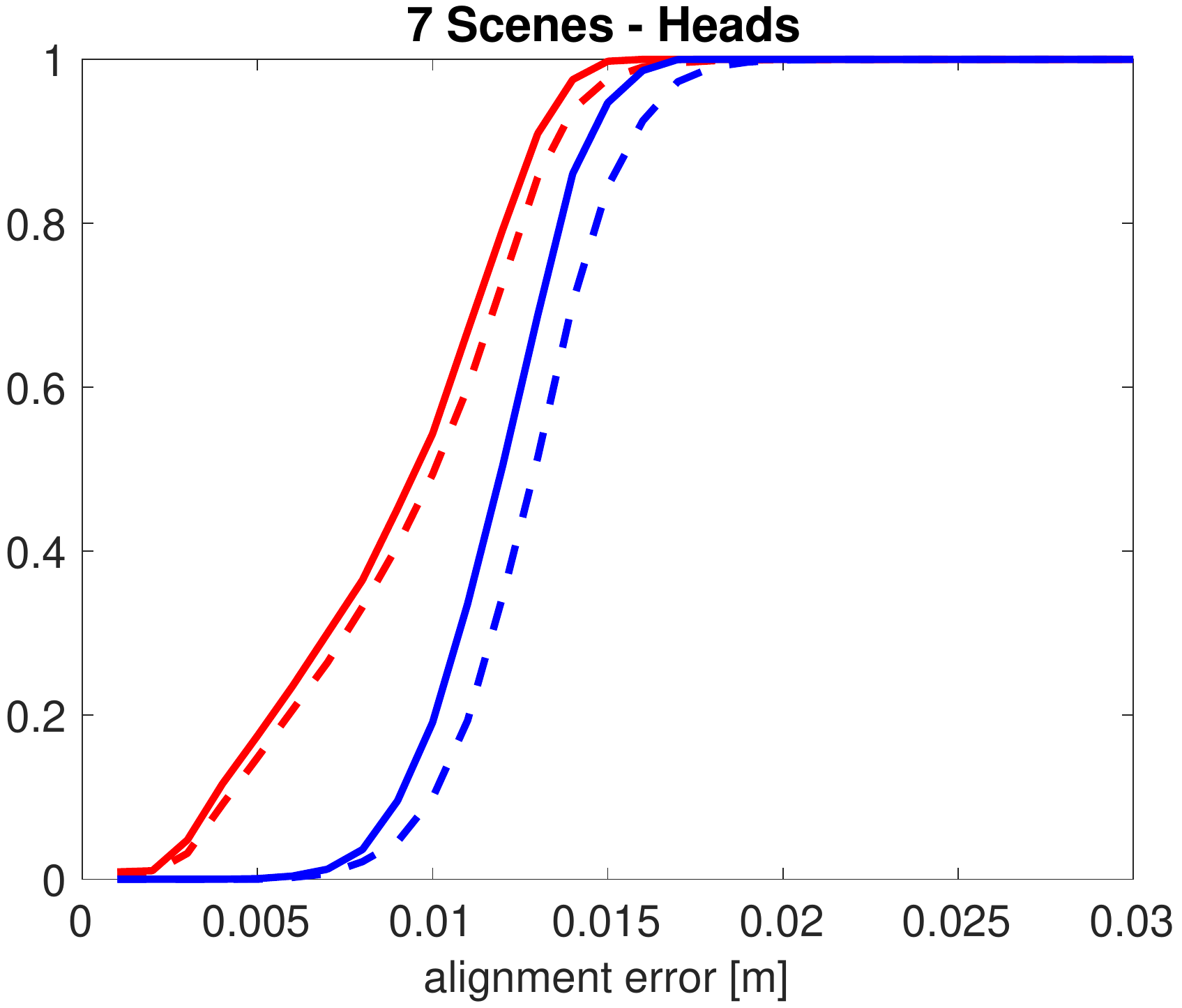}%
    \includegraphics[width=0.3\linewidth]{figures/alignment_7scenes_office.pdf}%
    \includegraphics[width=0.3\linewidth]{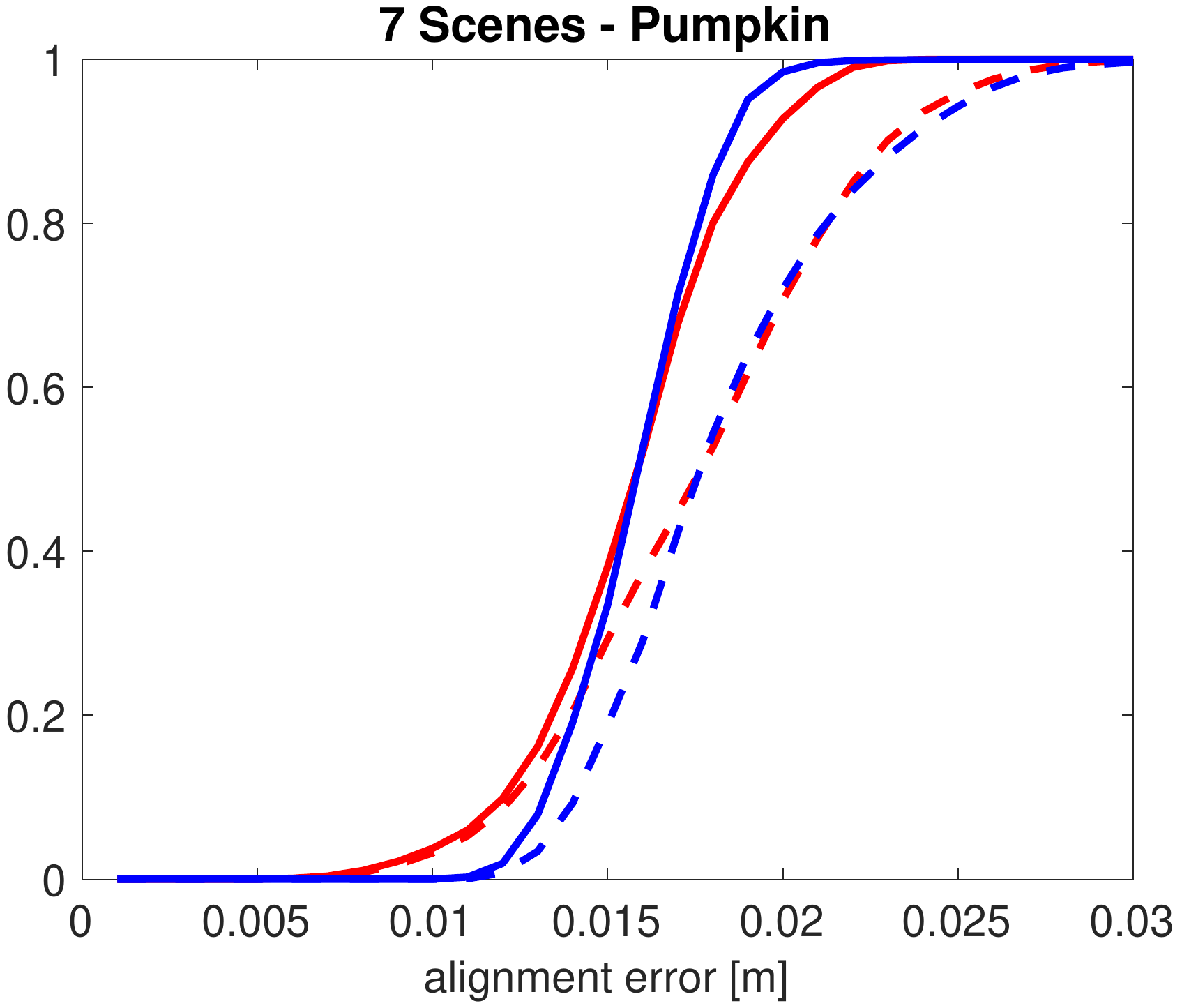}\\%
     \includegraphics[width=0.3\linewidth]{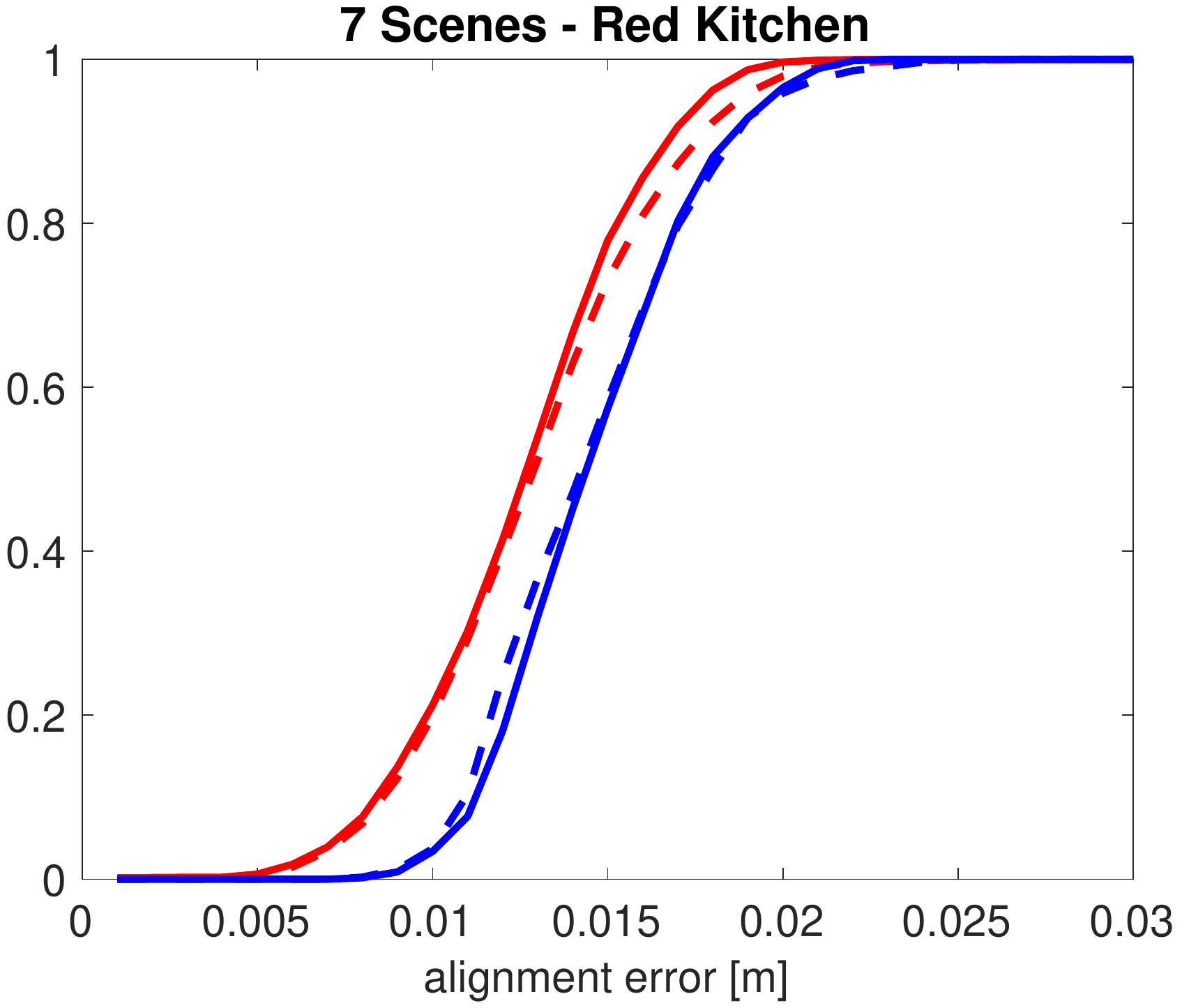}%
    \includegraphics[width=0.3\linewidth]{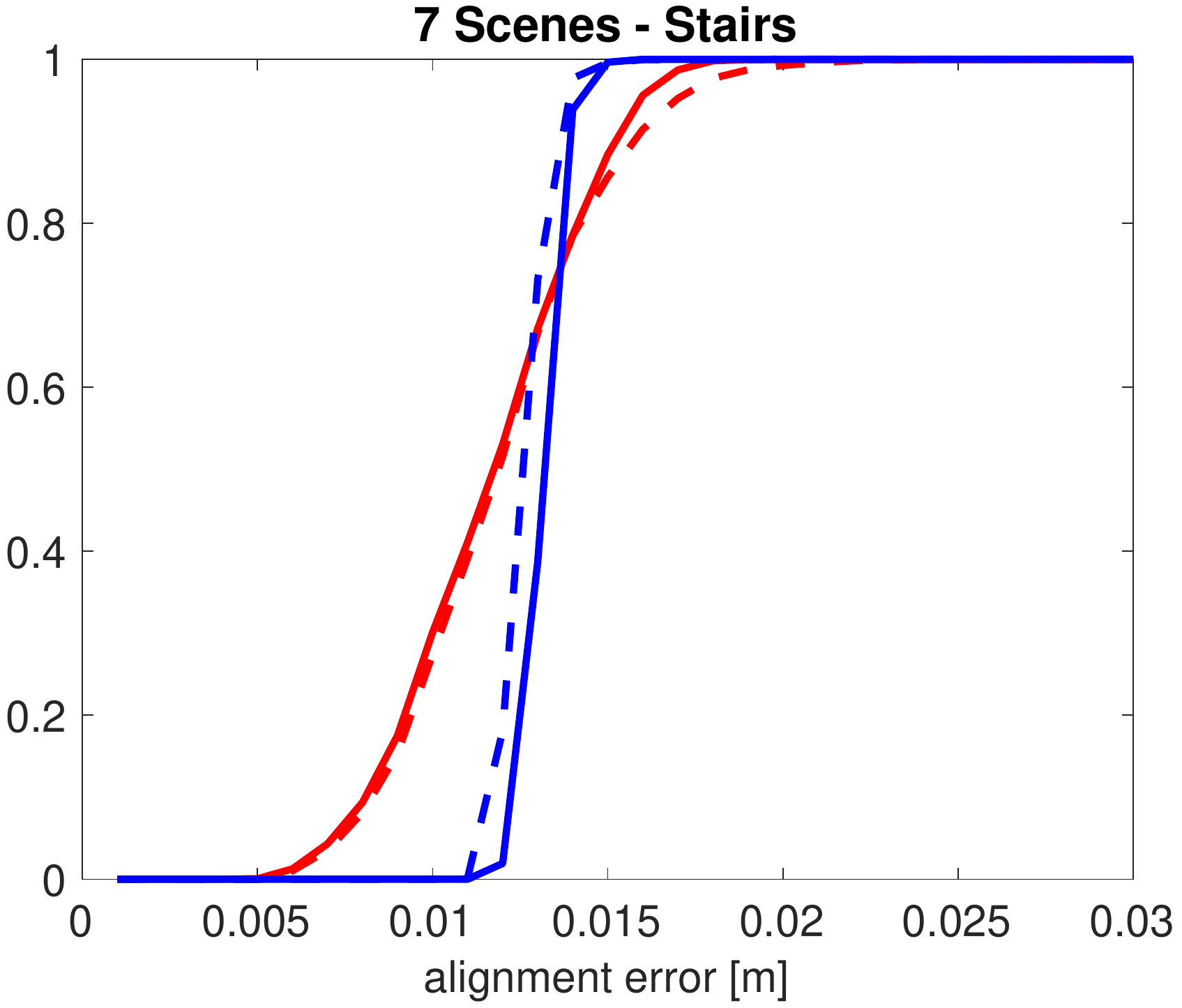}%
    \caption{\textbf{3D alignment statistics for the 7Scenes~\cite{Shotton2013CVPR} dataset.} We show cumulative distributions (cdfs) of the 3D alignment errors between the depth maps of train/train and test/train image pairs with a visual overlap of at least 30\% for the original (RGB-)D SLAM and the SfM pseudo GT. }
    \label{fig:7scenes_alignment}
\end{figure*}

\begin{figure*}[!t]
    \centering
    \vspace{-0.5cm}
    \includegraphics[width=0.3\linewidth]{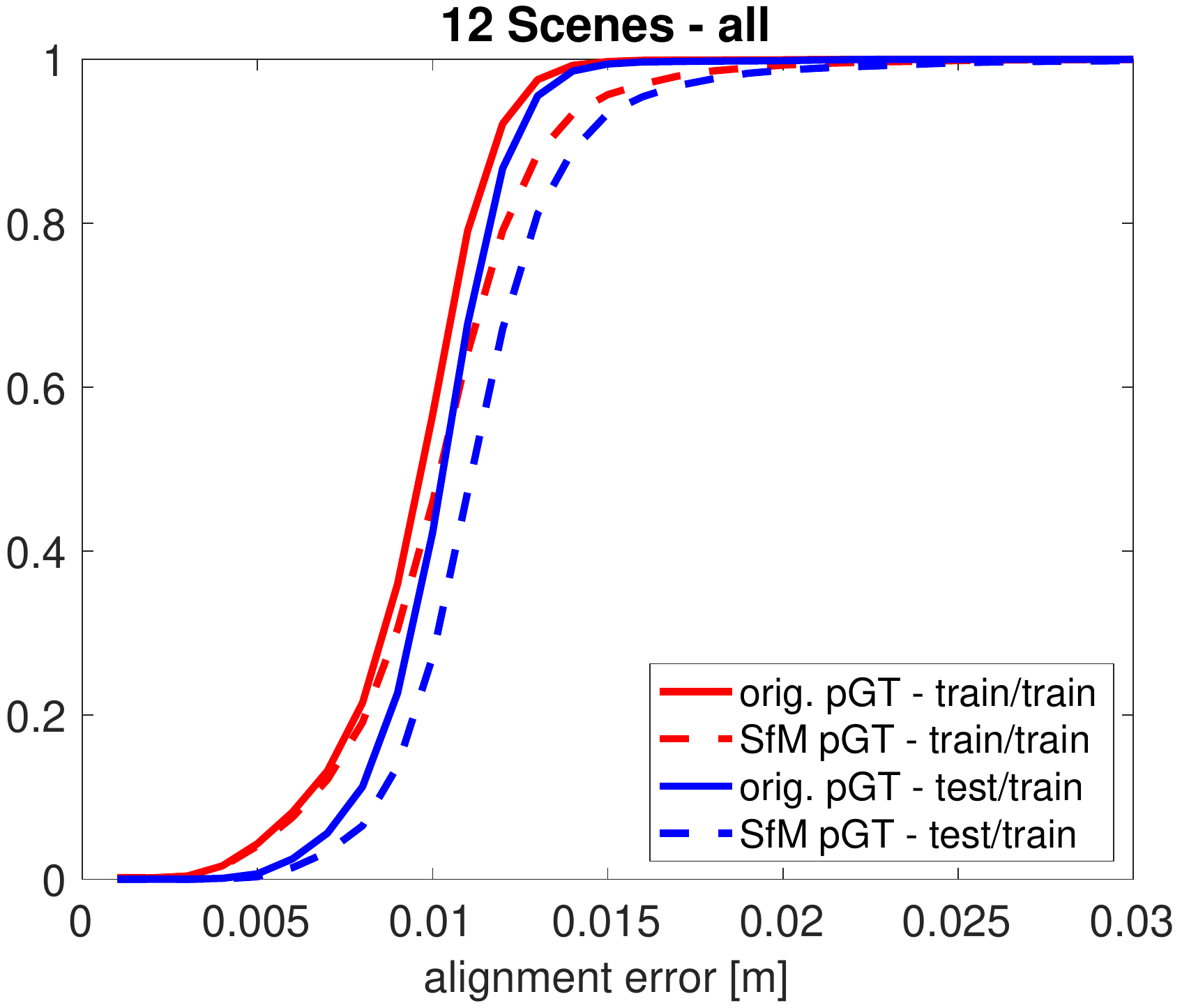}%
    \includegraphics[width=0.3\linewidth]{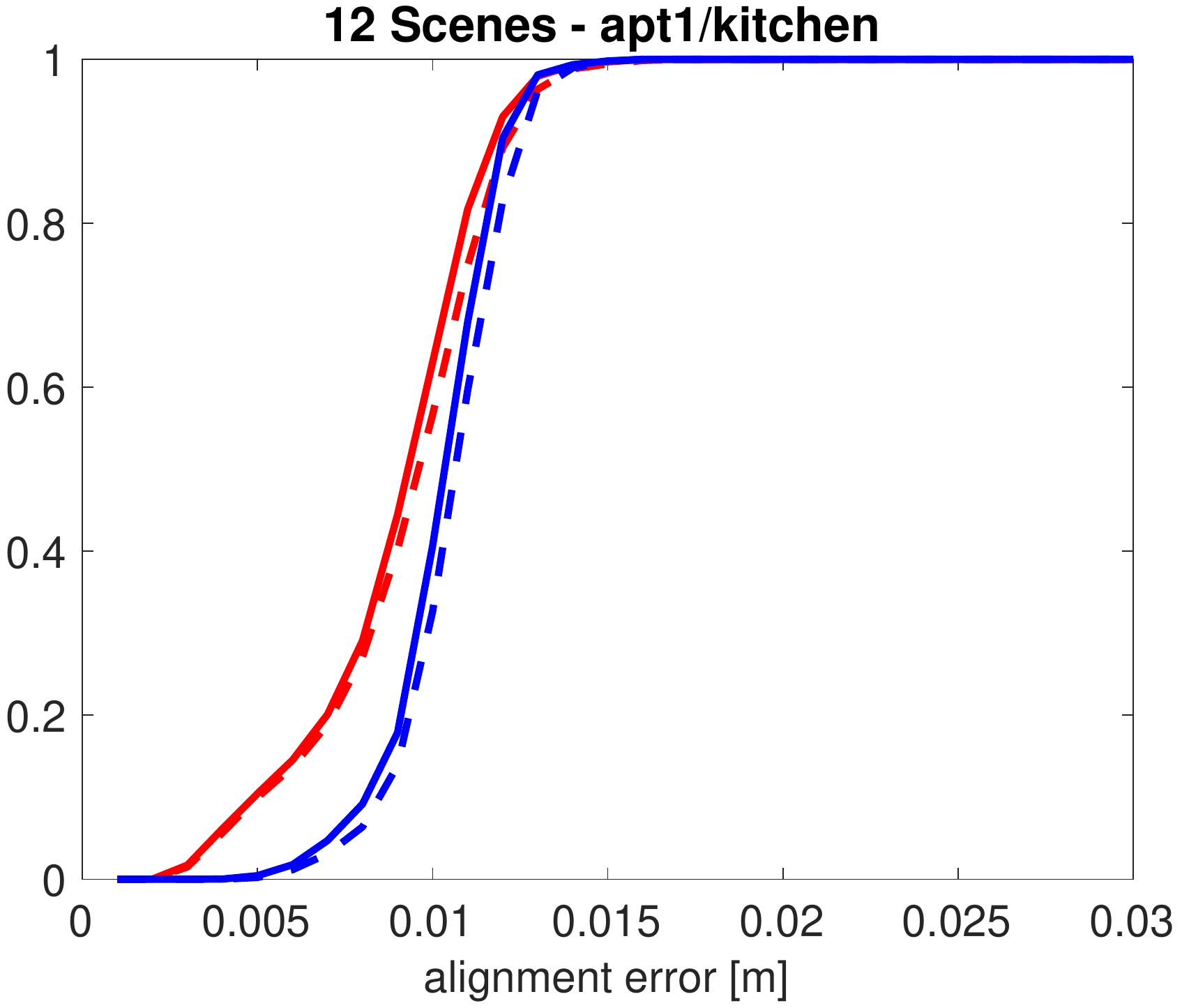}%
    \includegraphics[width=0.3\linewidth]{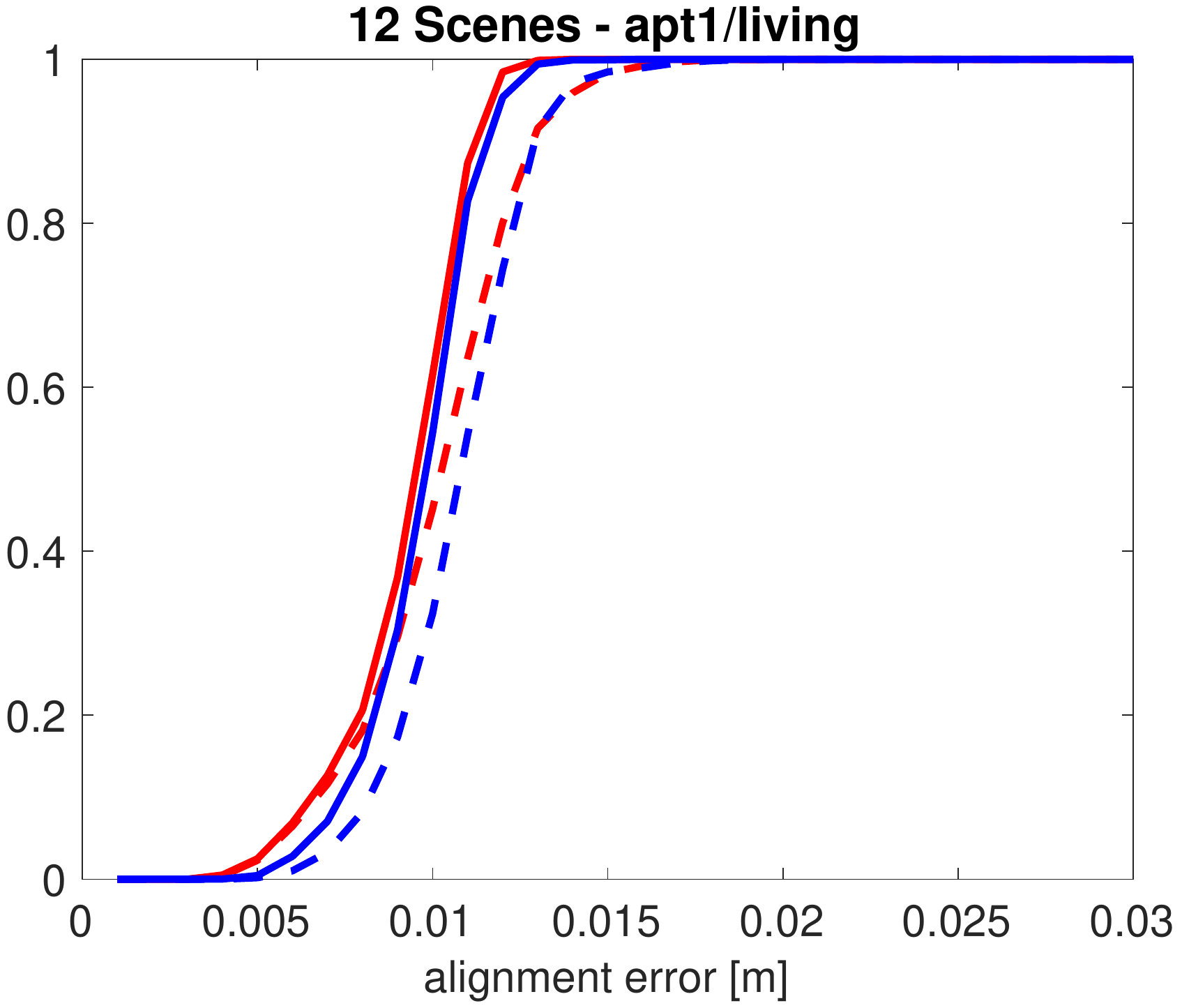}\\%
    \includegraphics[width=0.3\linewidth]{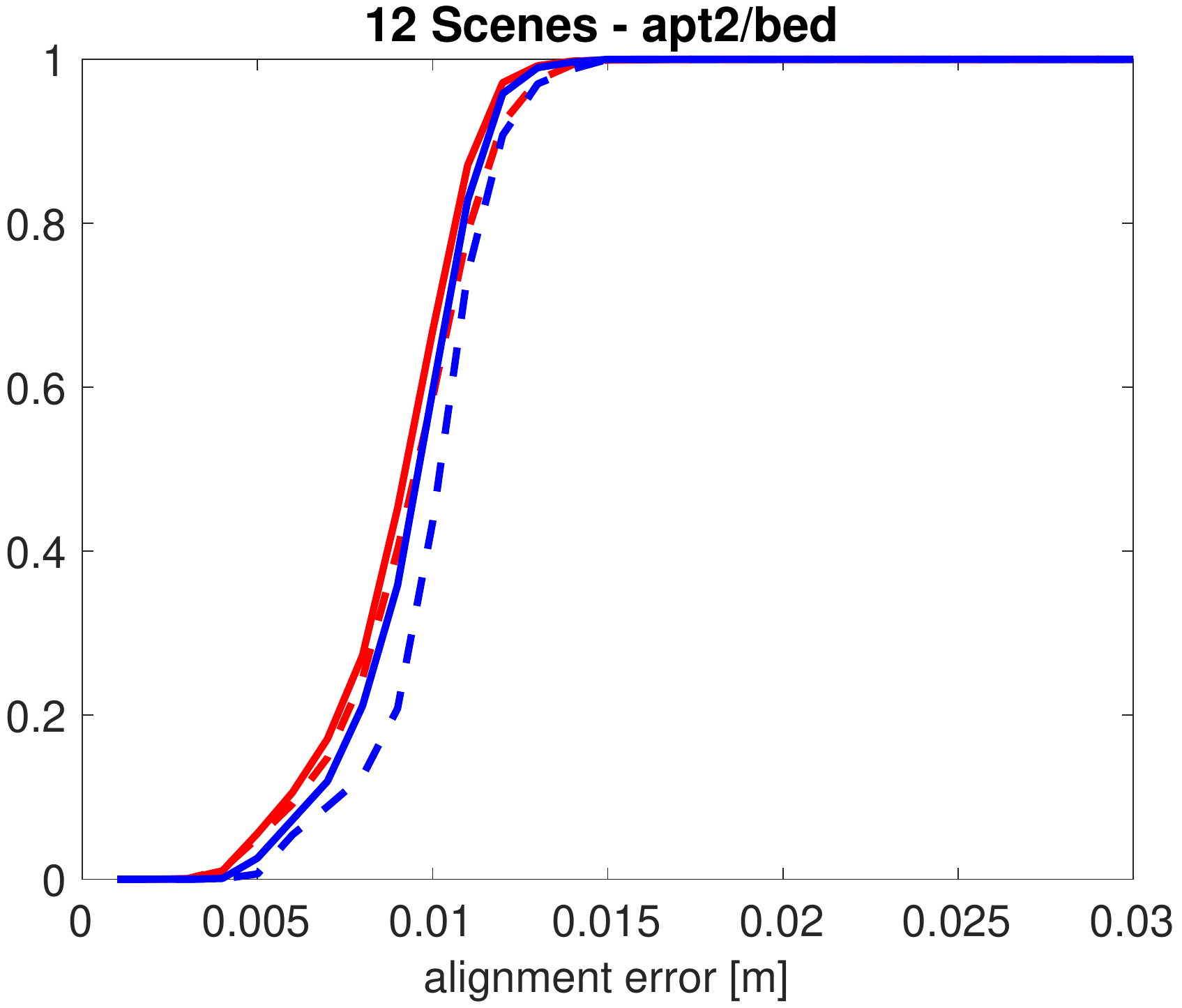}%
    \includegraphics[width=0.3\linewidth]{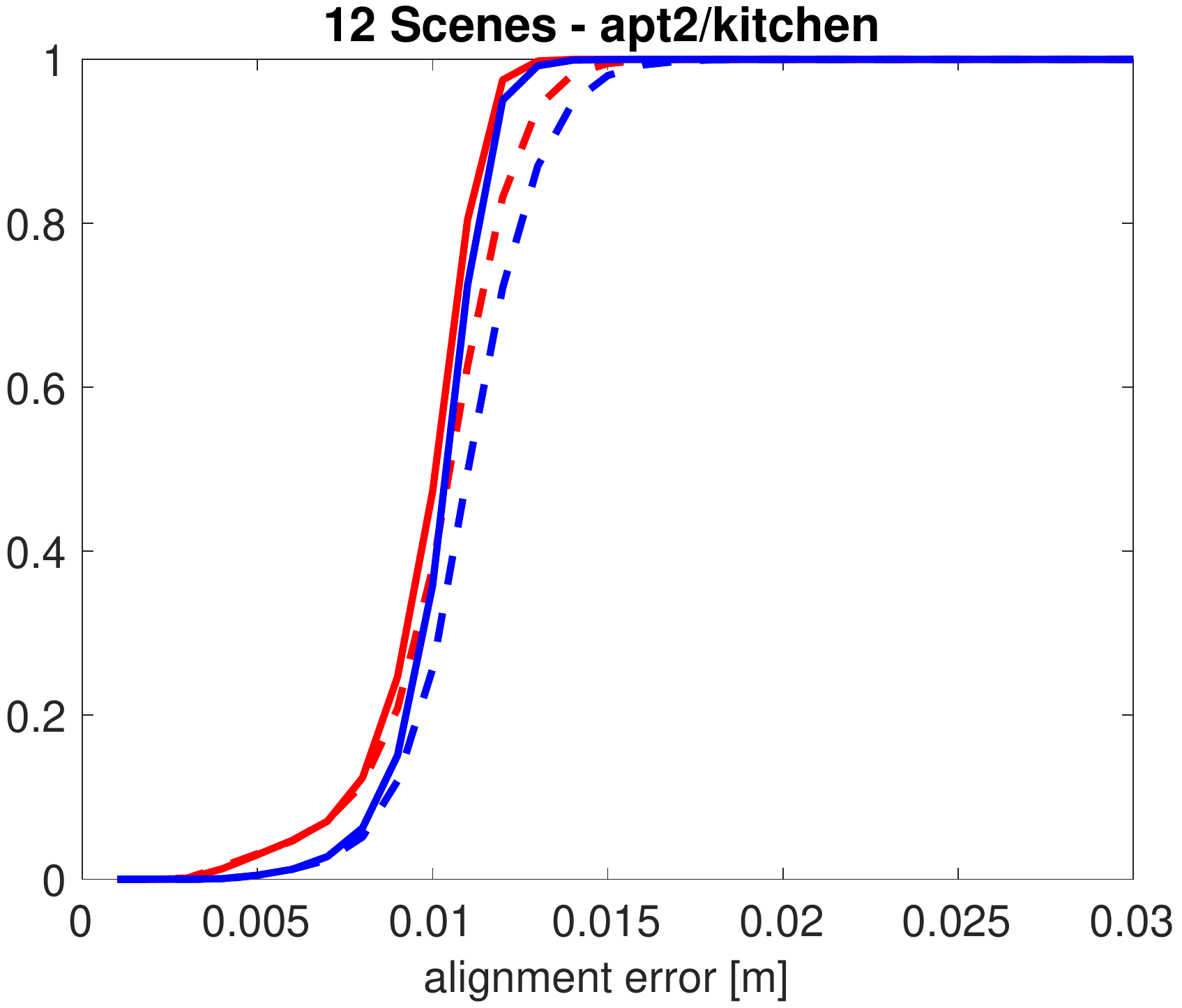}%
    \includegraphics[width=0.3\linewidth]{figures/alignment_12scenes_apt2_living.pdf}\\%
    \includegraphics[width=0.3\linewidth]{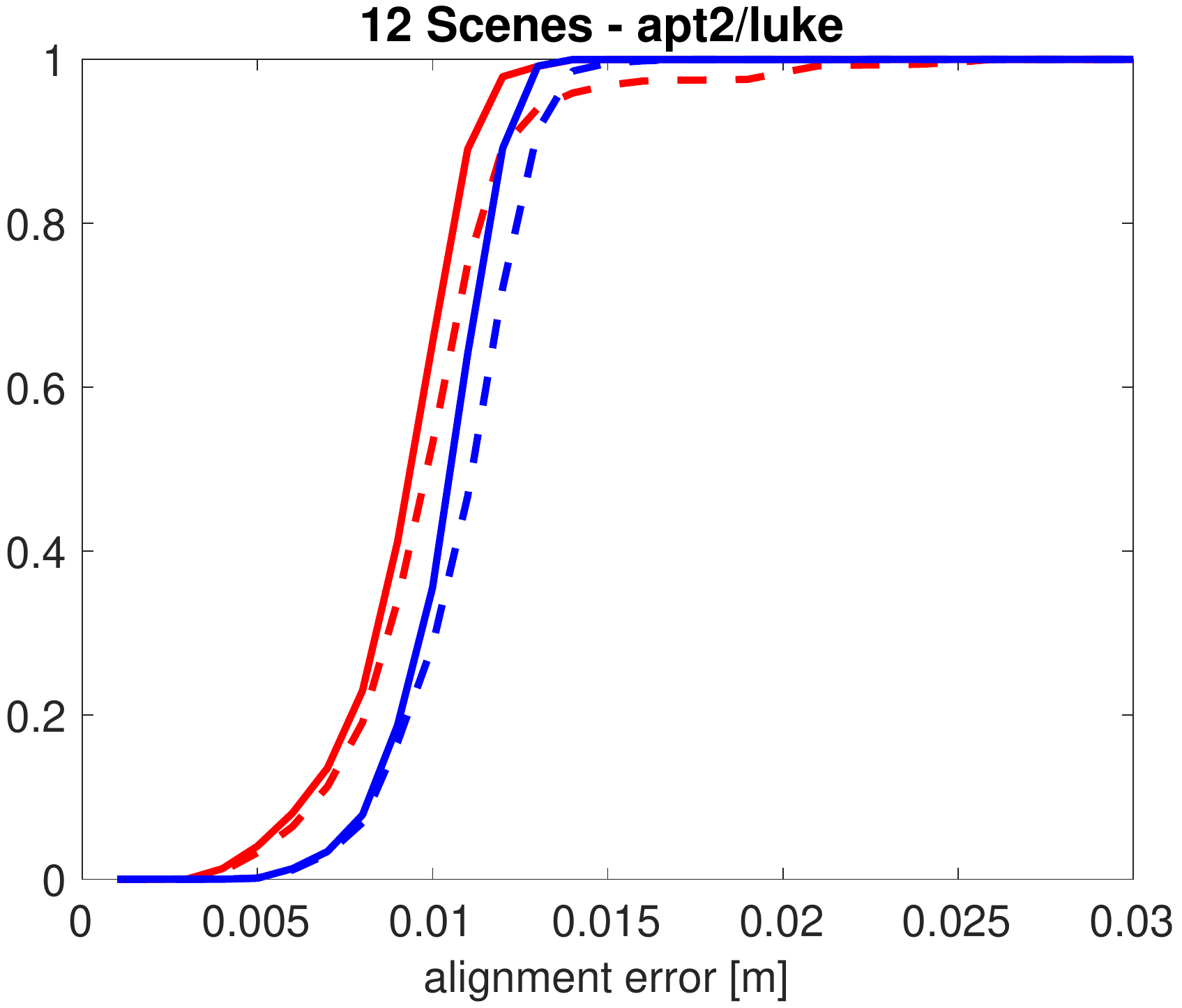}%
    \includegraphics[width=0.3\linewidth]{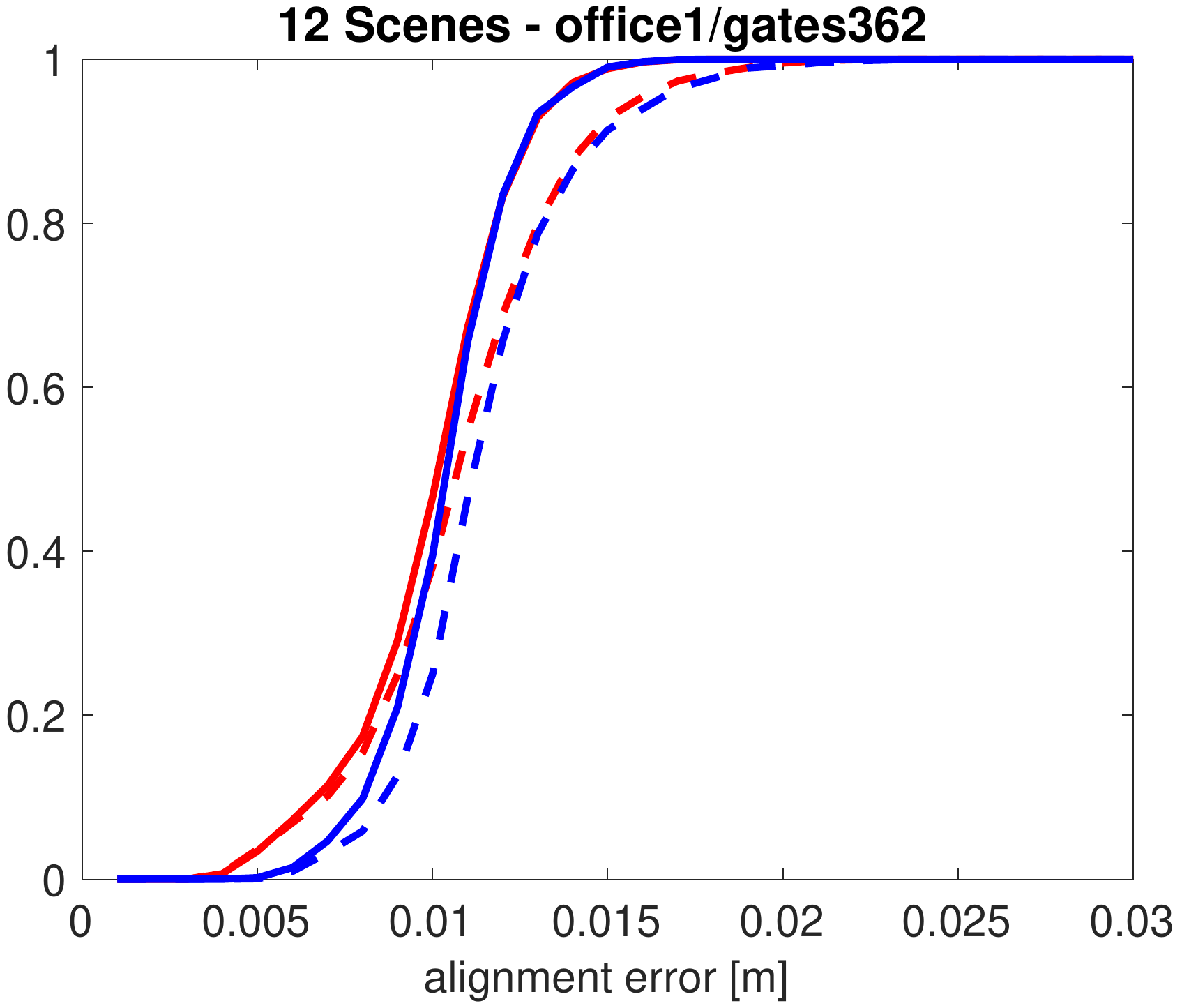}%
    \includegraphics[width=0.3\linewidth]{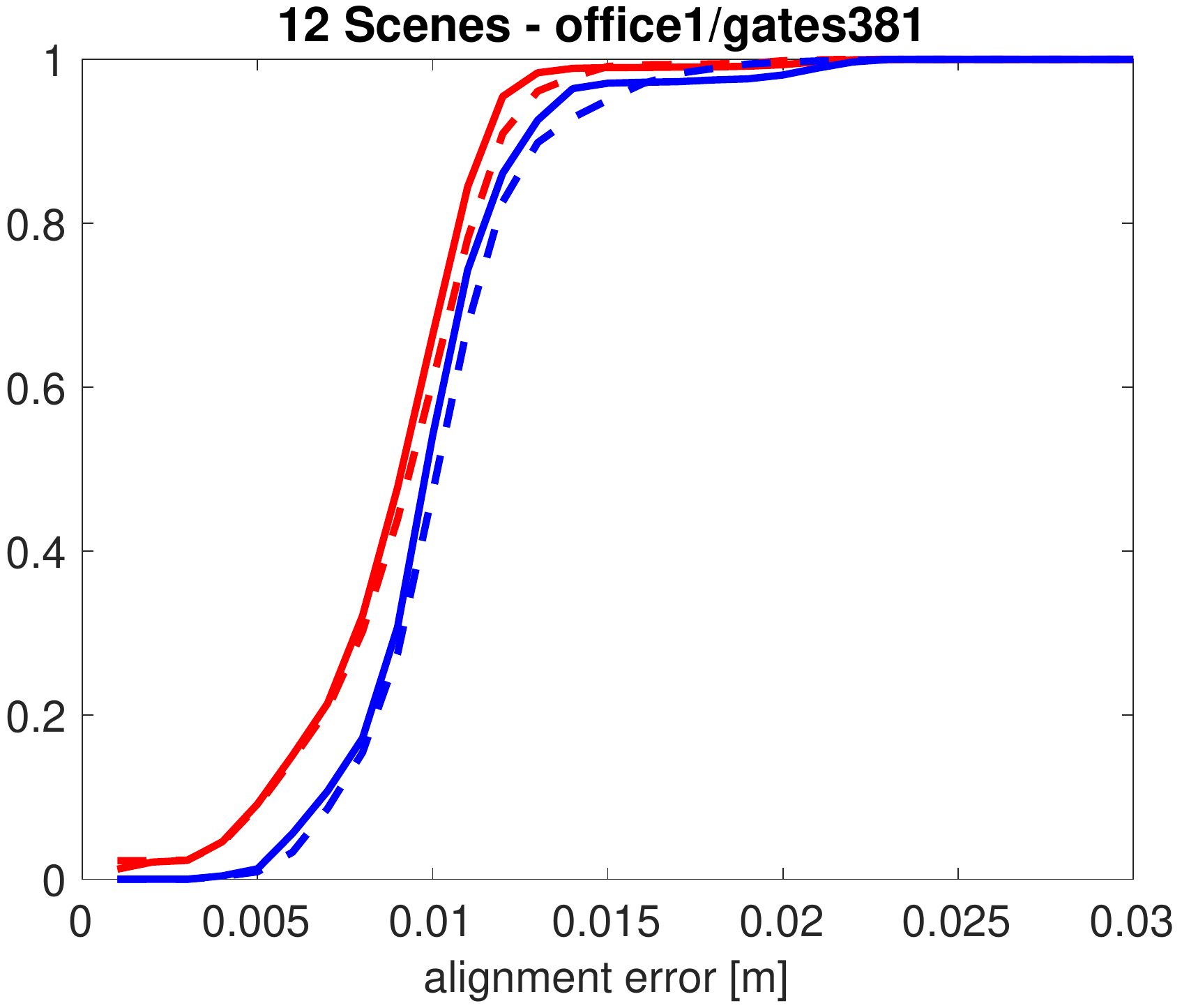}\\%
    \includegraphics[width=0.3\linewidth]{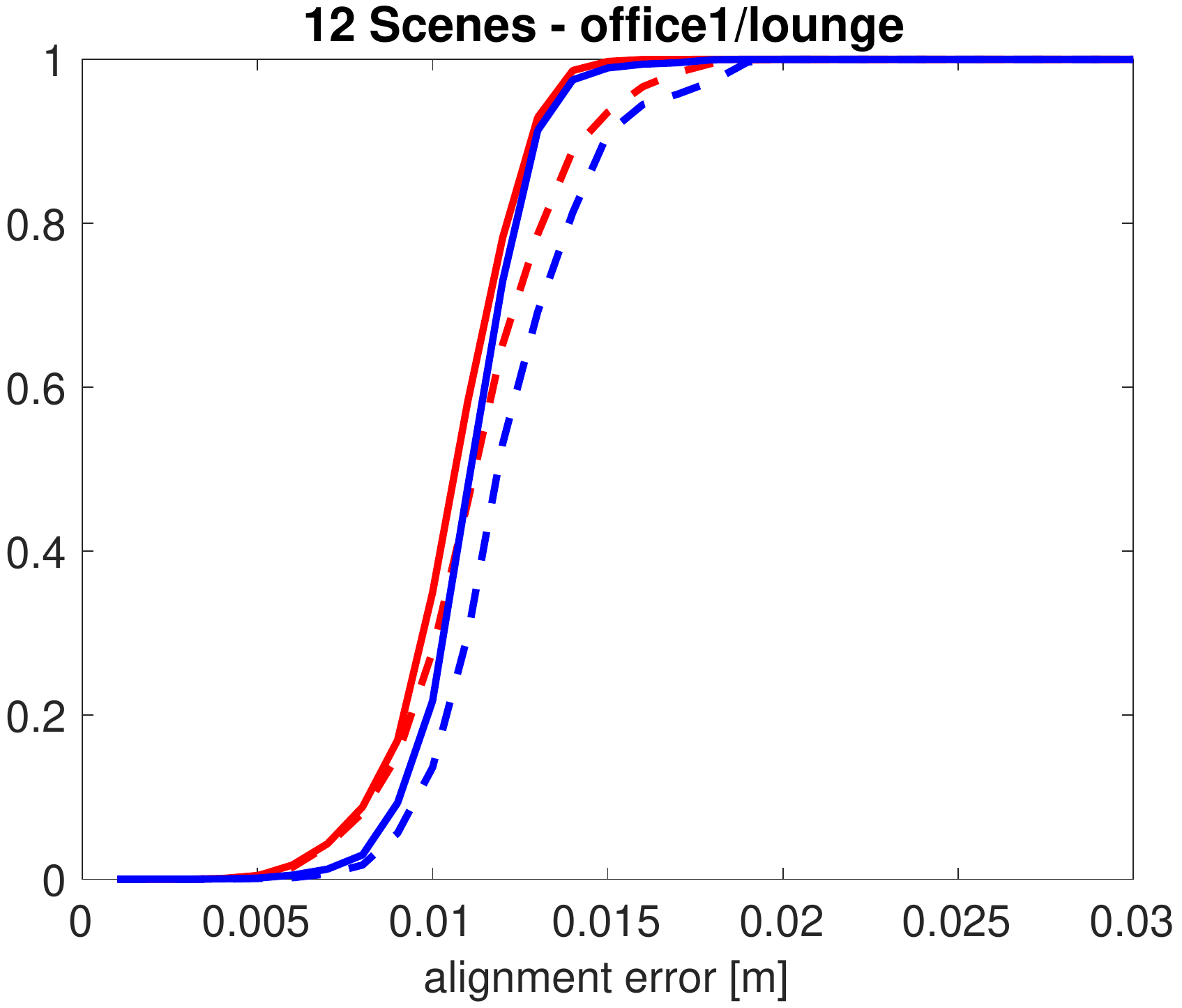}%
    \includegraphics[width=0.3\linewidth]{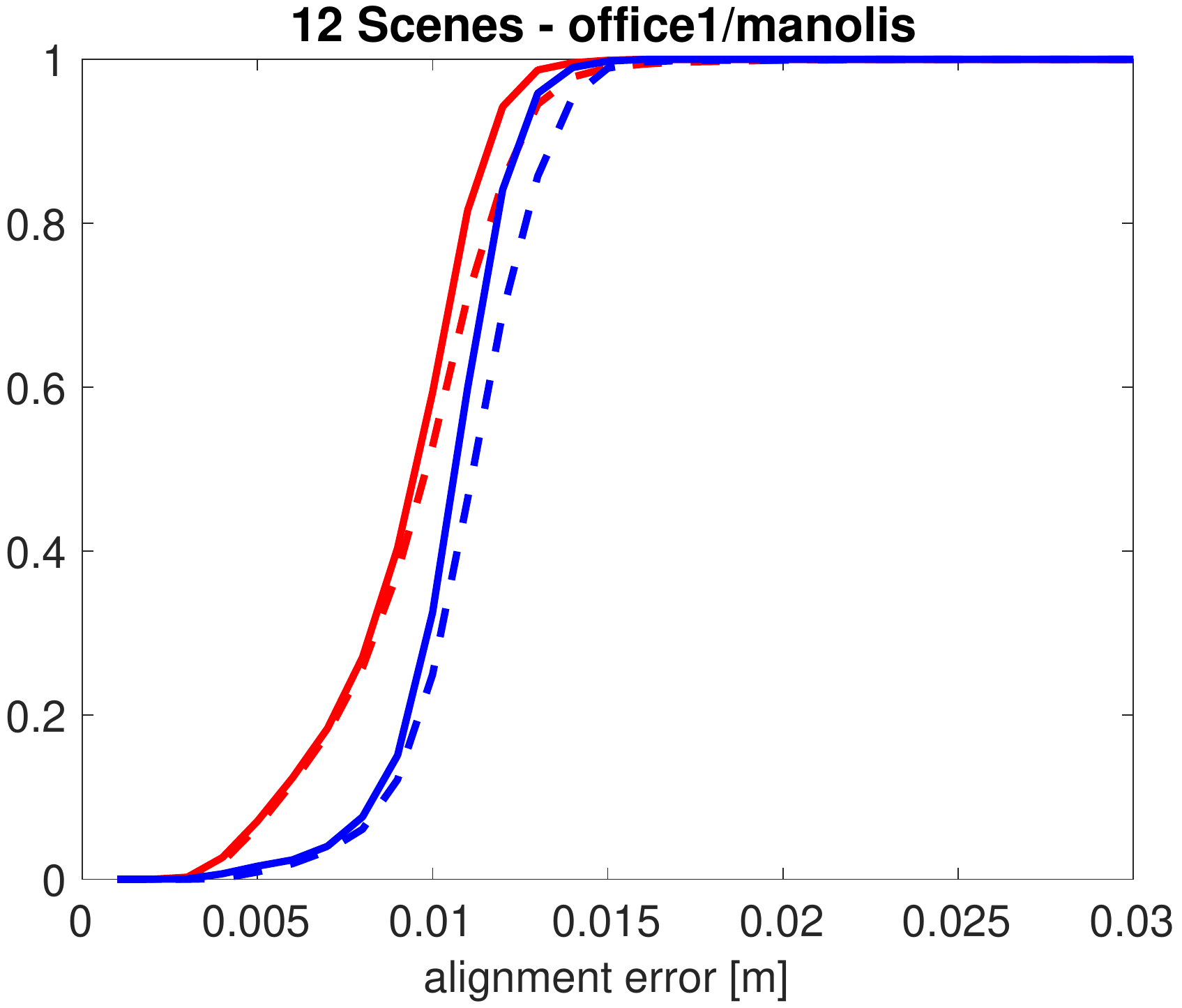}%
    \includegraphics[width=0.3\linewidth]{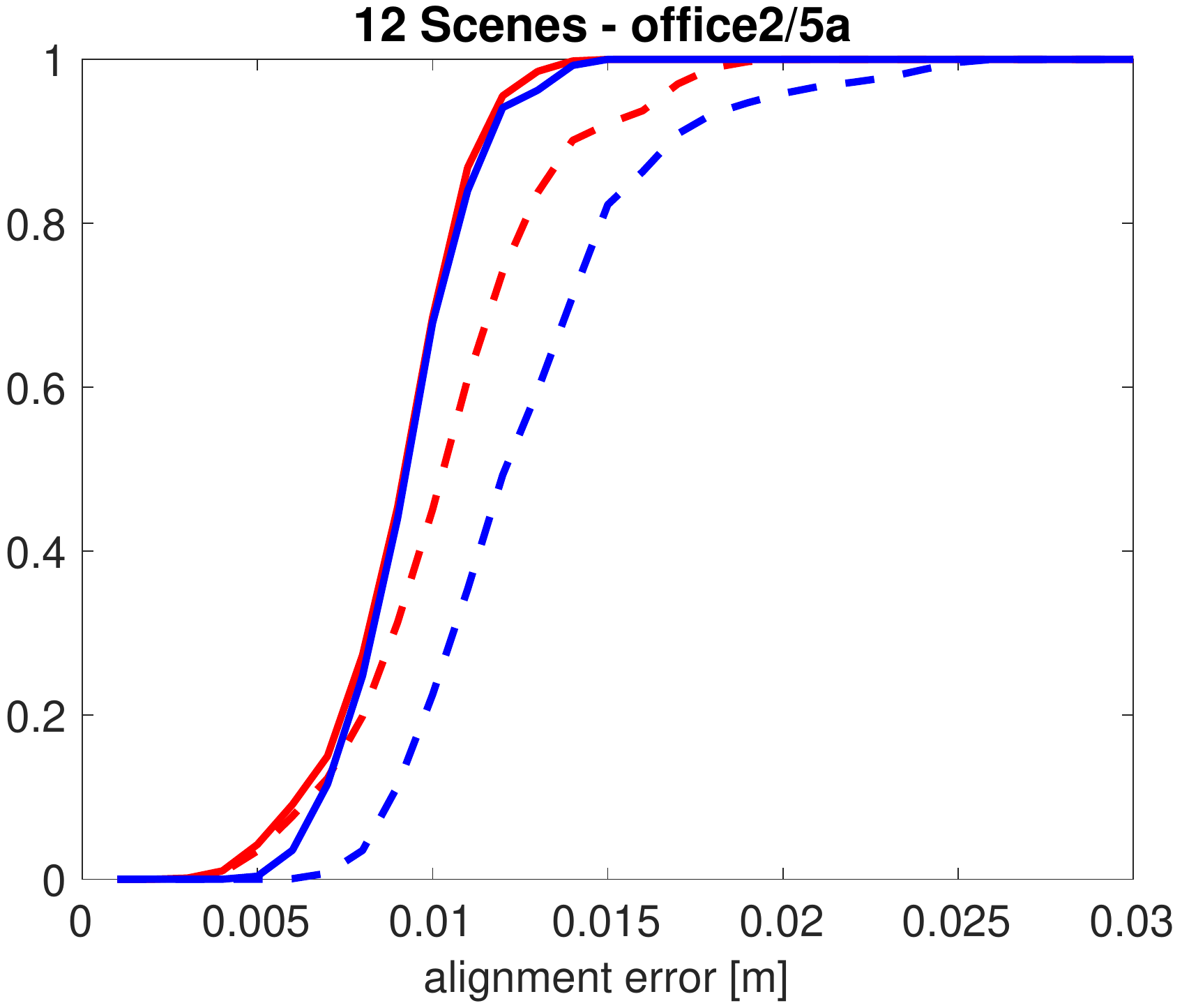}\\%
    \includegraphics[width=0.3\linewidth]{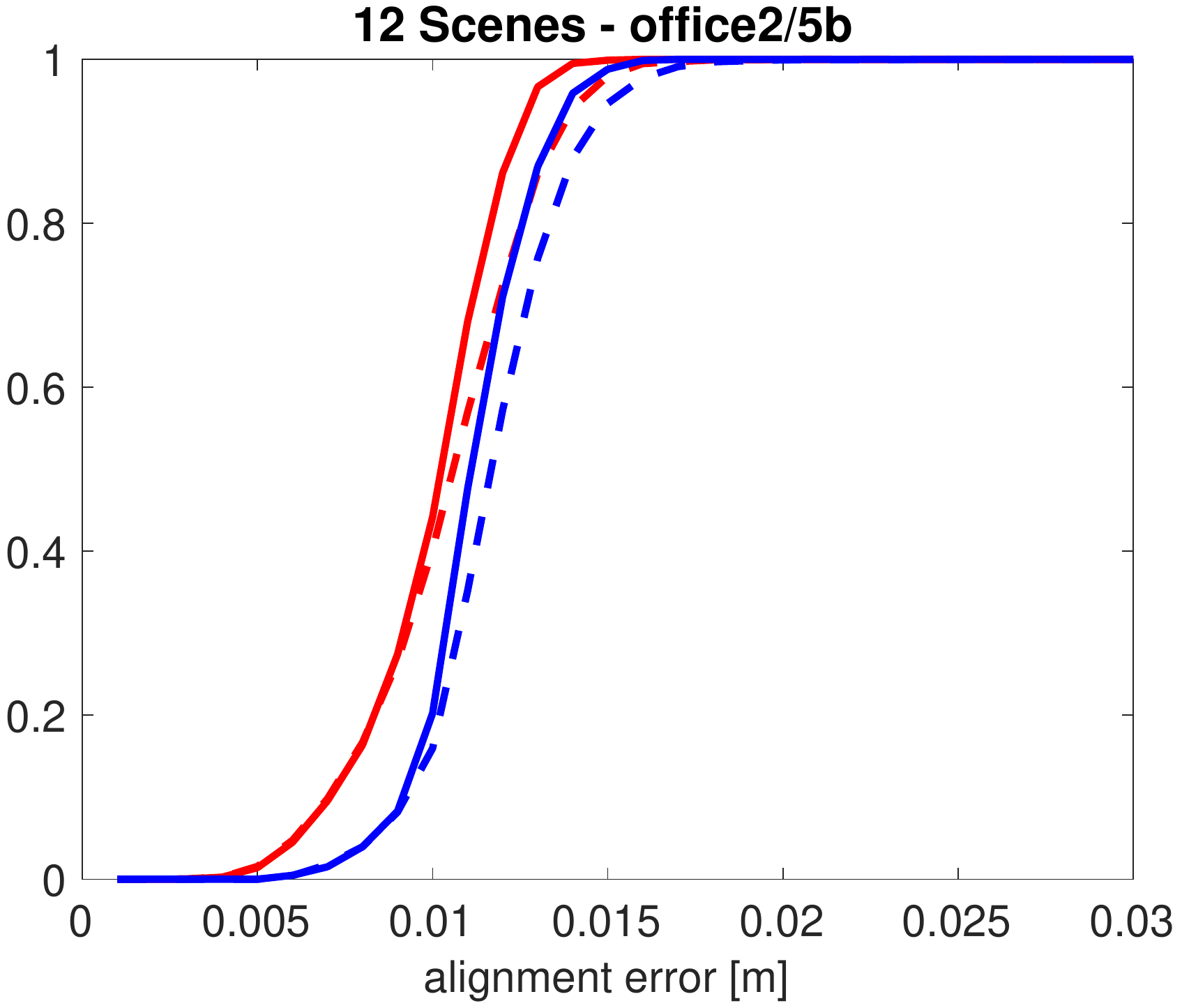}%
    \caption{\textbf{3D alignment statistics for the 12Scenes~\cite{Valentin20163DV} dataset.} We show cumulative distributions (cdfs) of the 3D alignment errors between the depth maps of train/train and test/train image pairs with a visual overlap of at least 30\% for the original (RGB-)D SLAM and the SfM pseudo GT.}
    \label{fig:12scenes_alignment_1}
\end{figure*}

\section{Quantitative Comparisons of pGT}
\label{sec:quantitative_comparisons}
Fig.~\ref{fig:stats_pc} shows cumulative distributions over 3D alignment statistics for the 7Scenes and 12Scenes datasets (\cf Sec.~\ref{sec:comparison_pgt} for details). 
While Fig.~\ref{fig:stats_pc} shows average statistics over all scenes in each dataset and one selected scene per dataset, here we show the distributions for all scenes of the two datasets.

Fig.~\ref{fig:7scenes_alignment} shows the cumulative distributions for all scenes of the 7Scenes dataset. 
As can be seen, the original (RGB-)D SLAM \GT results in a more accurate alignment for most scenes compared to the SfM \GT. 
For the Red Kitchen and Stairs scenes, there is little difference between the two versions of the \GT and the SfM \GT produces a (slightly) more accurate alignment for the test/train pairs.  

Similarly, Fig.~\ref{fig:12scenes_alignment_1} shows the cumulative distributions for all scenes of the 12Scenes dataset. 
Again, we observe that the original (RGB-)D SLAM \GT results in more accurate 3D alignments compared to the SfM \GT. 
However, for most scenes, the difference between both versions of the \GT is smaller than for the 7Scenes dataset. 

\section{Visual Re-Localization Evaluation}
\label{sec:localization}

\begin{figure*}[!t]
    \centering
    \vspace{-0.5cm}
    \includegraphics[width=0.33\linewidth]{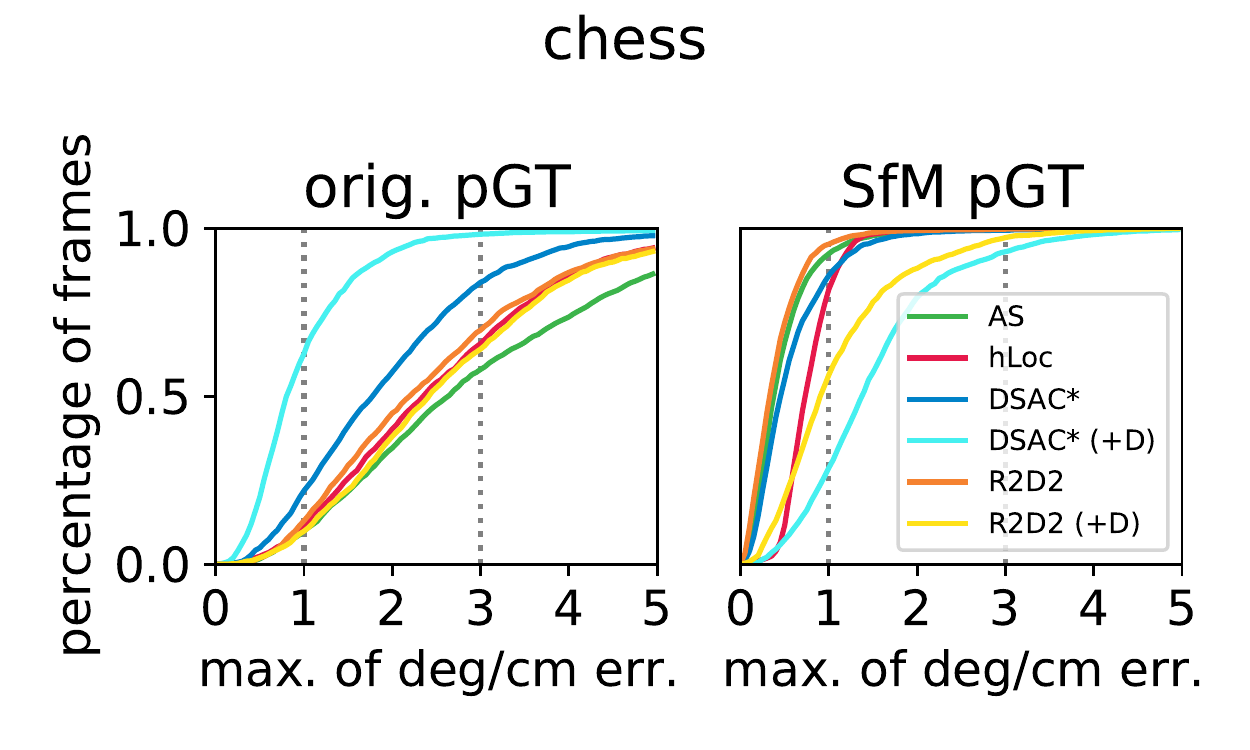}%
    \includegraphics[width=0.33\linewidth]{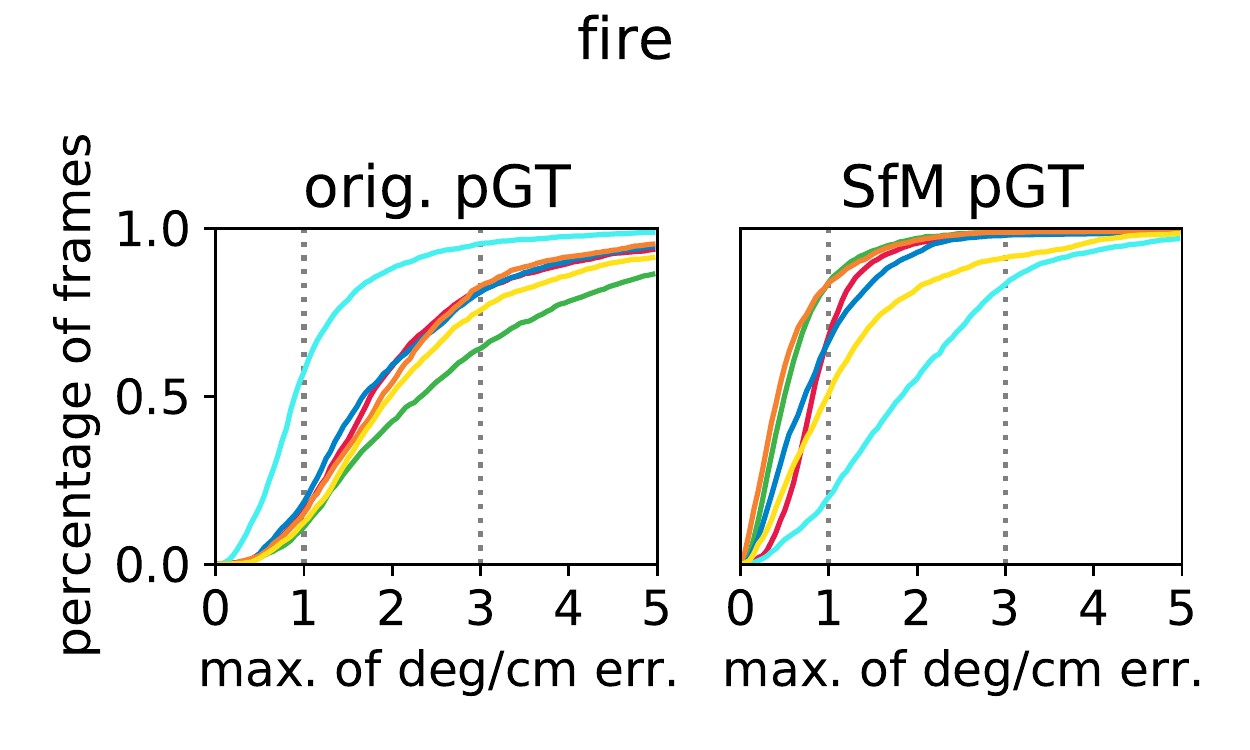}%
    \includegraphics[width=0.33\linewidth]{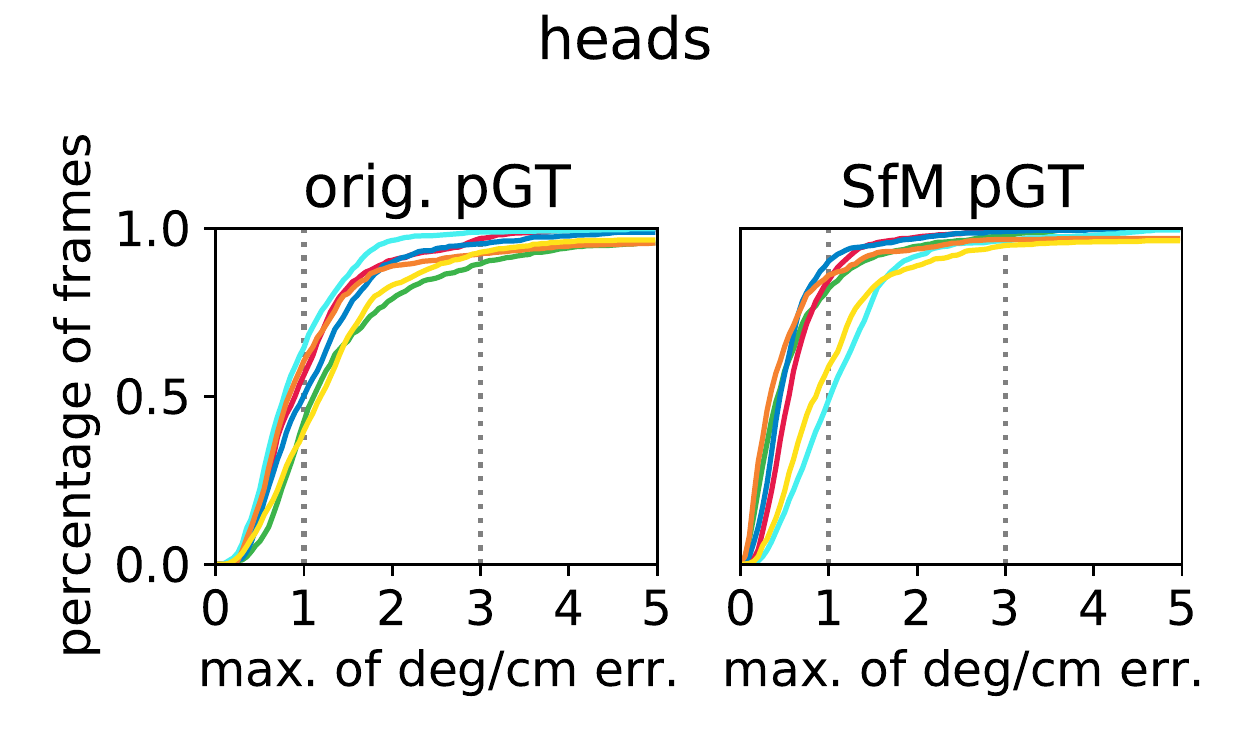}\\%
    \includegraphics[width=0.33\linewidth]{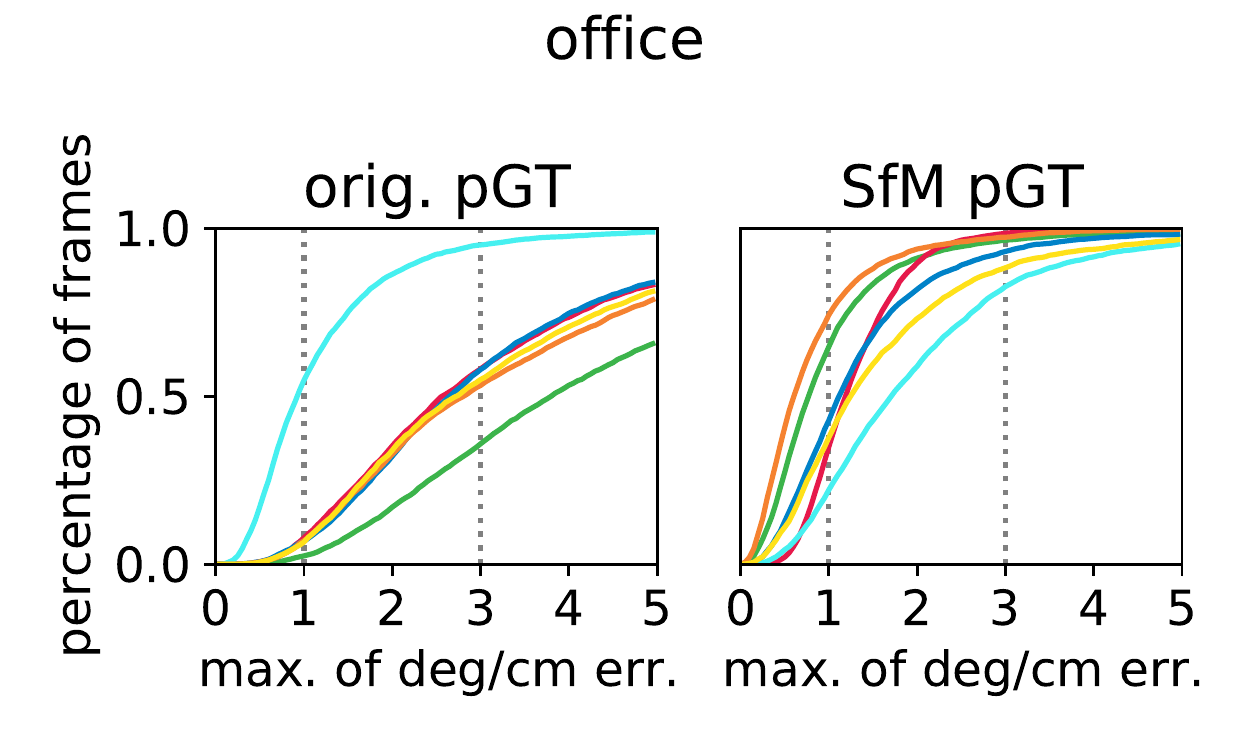}%
    \includegraphics[width=0.33\linewidth]{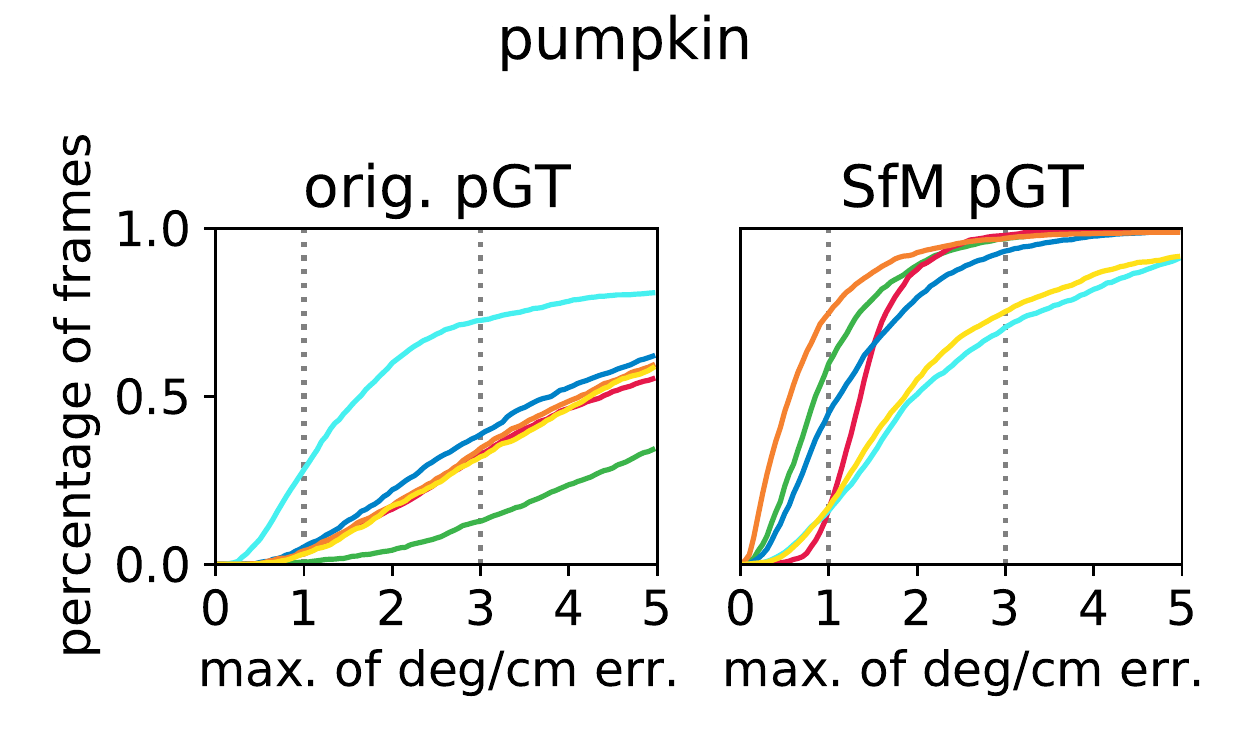}%
    \includegraphics[width=0.33\linewidth]{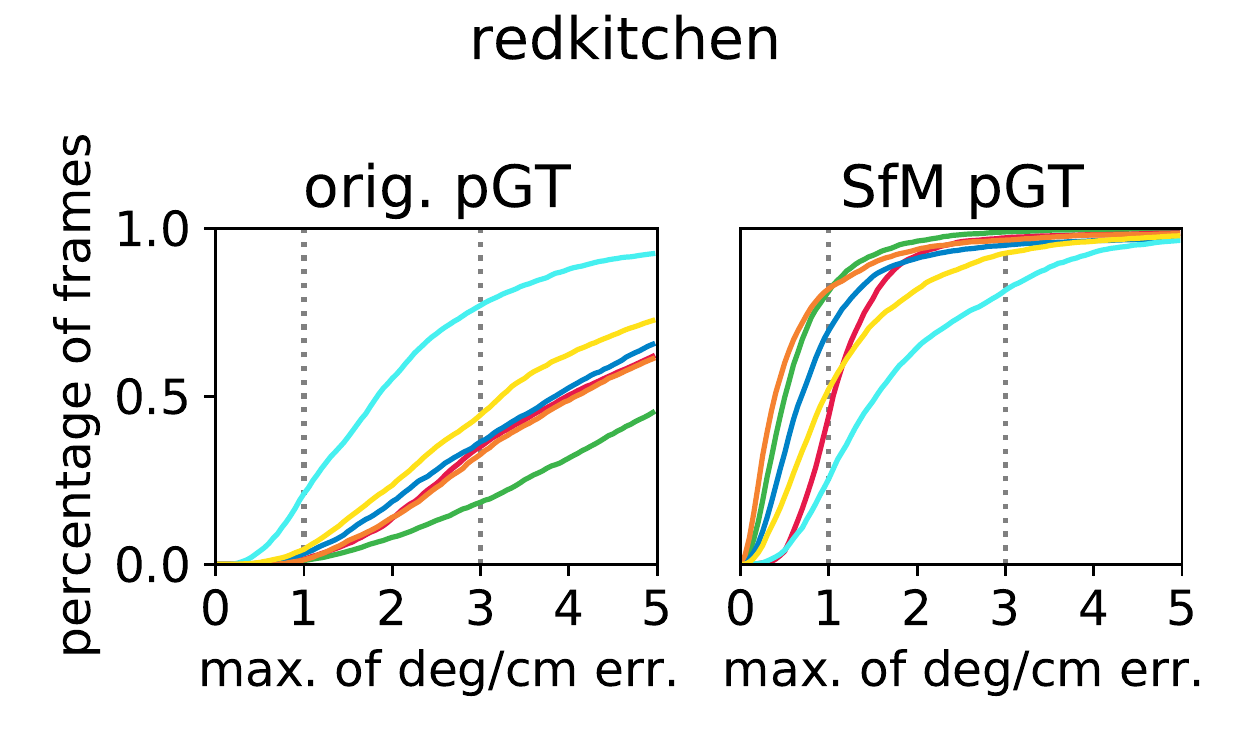}\\%
    \includegraphics[width=0.33\linewidth]{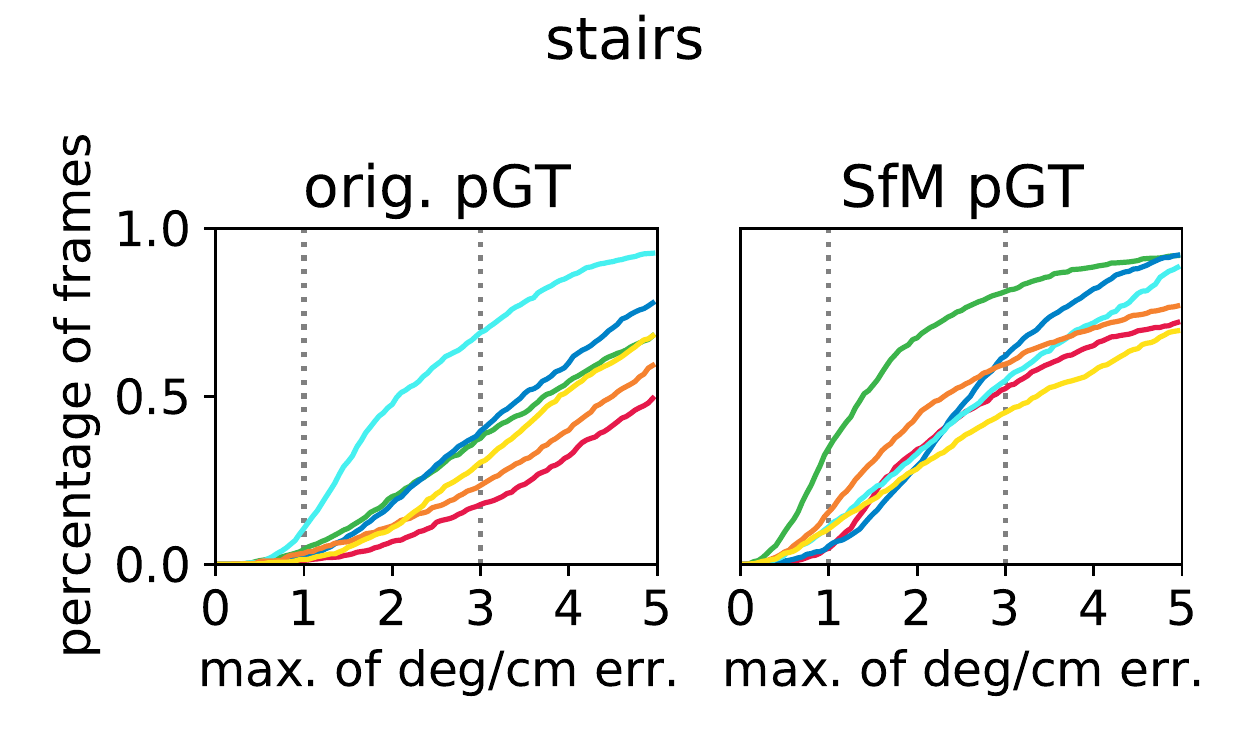}%
    \caption{\textbf{Pose error for 7Scenes.} We show cum.~distributions of the pose error (max.~of rotation and translation error) for all scenes of 7Scenes \cite{Shotton2013CVPR}. Dotted vertical lines correspond to 1cm,1$^\circ$ and 3cm,3$^\circ$ thresholds for reference.}
    \label{fig:7s_pose_plots}
\end{figure*}

\begin{figure*}[!t]
    \centering
    \includegraphics[width=0.33\linewidth]{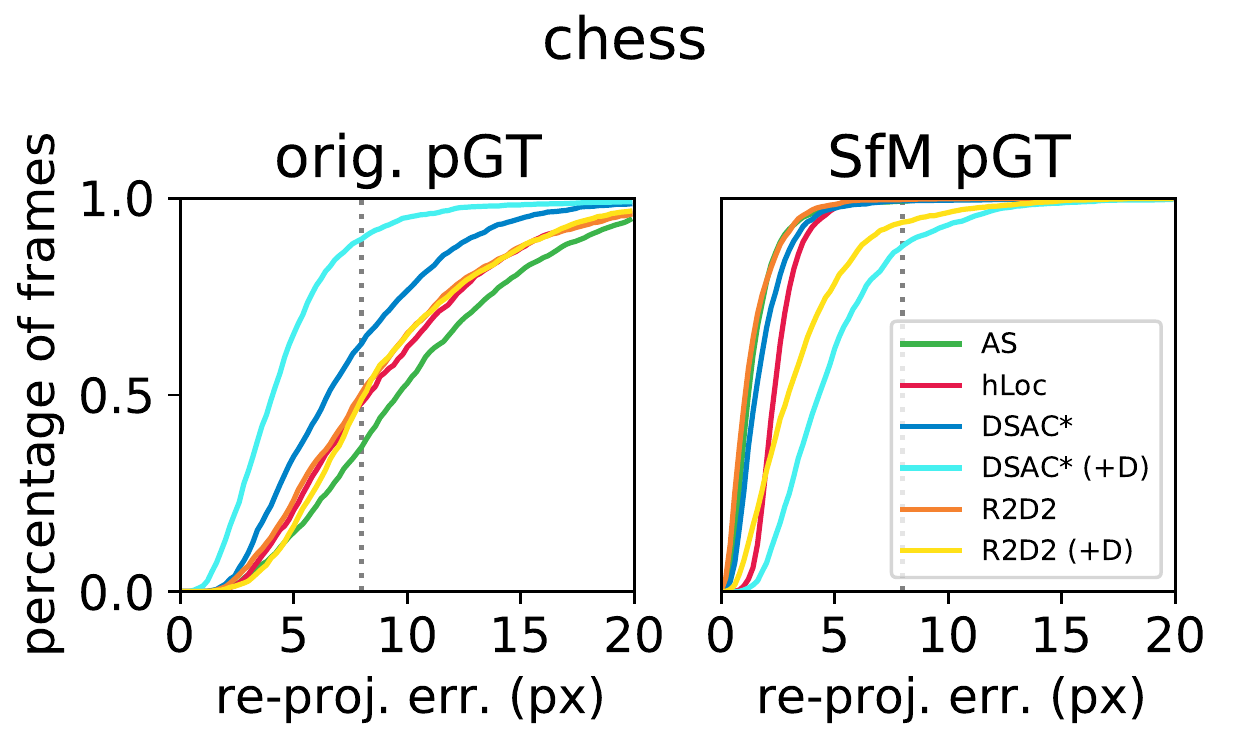}%
    \includegraphics[width=0.33\linewidth]{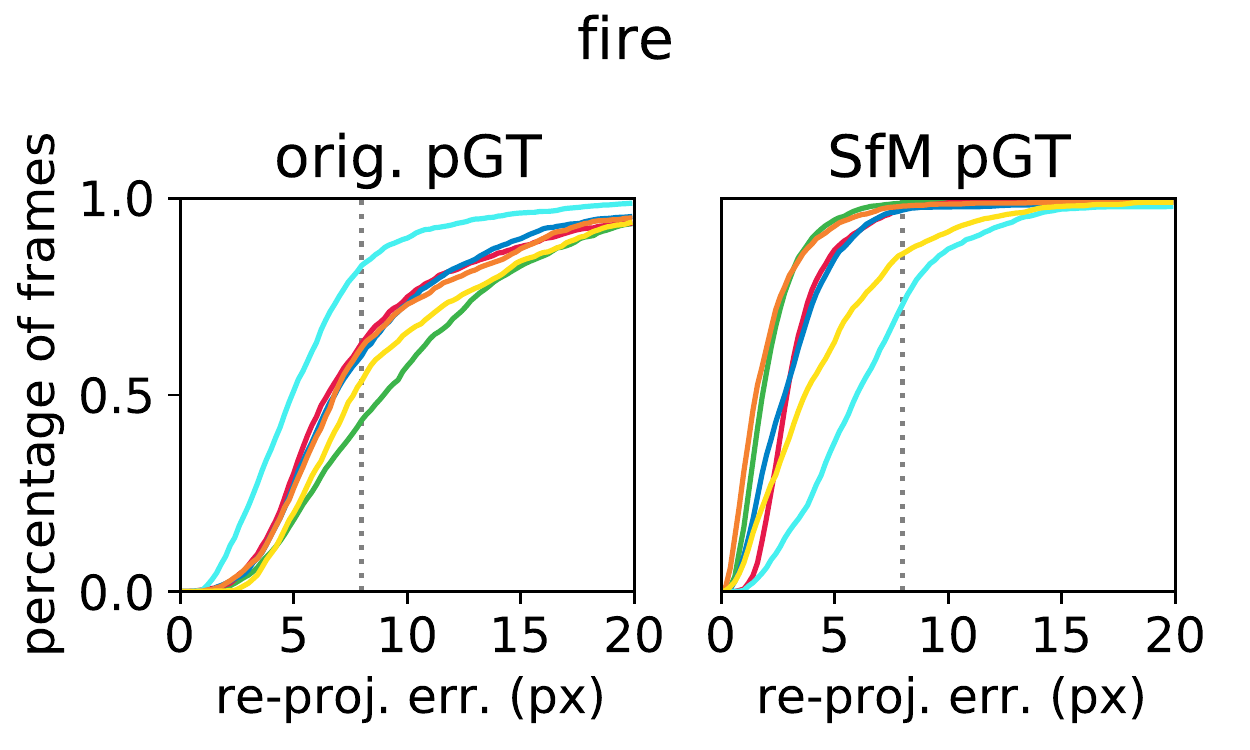}%
    \includegraphics[width=0.33\linewidth]{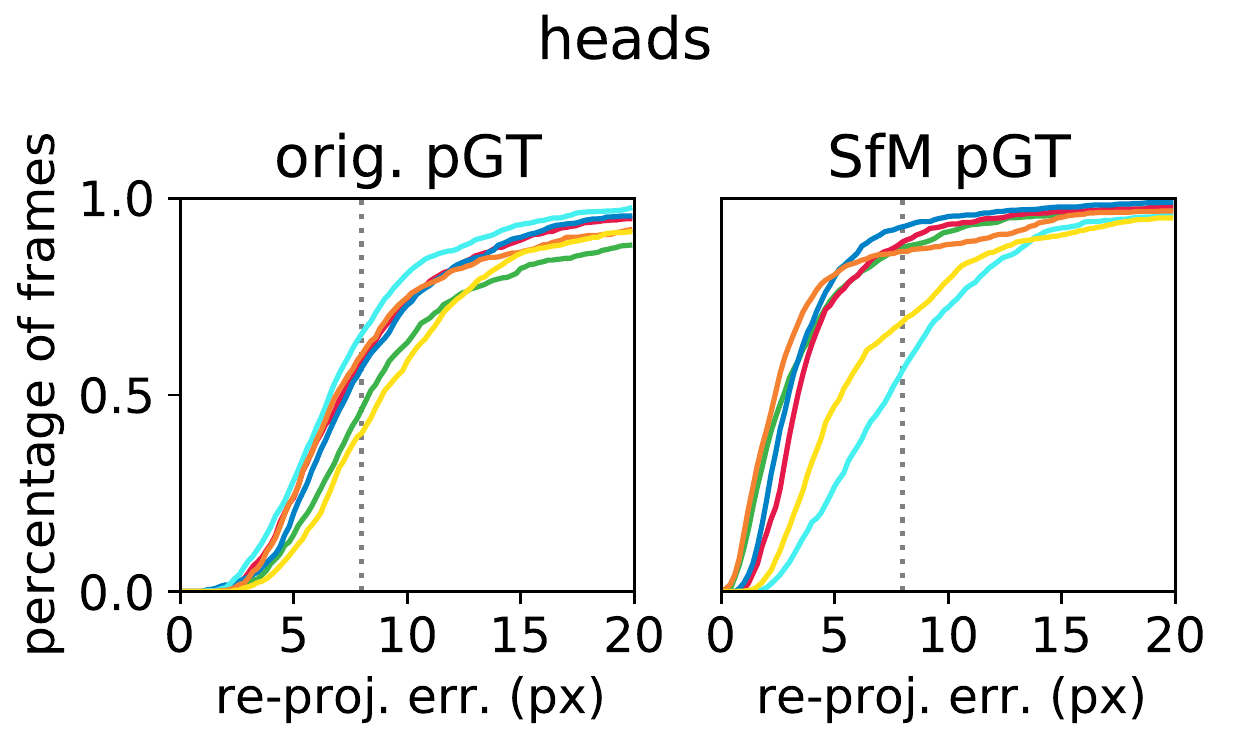}\\%
    \includegraphics[width=0.33\linewidth]{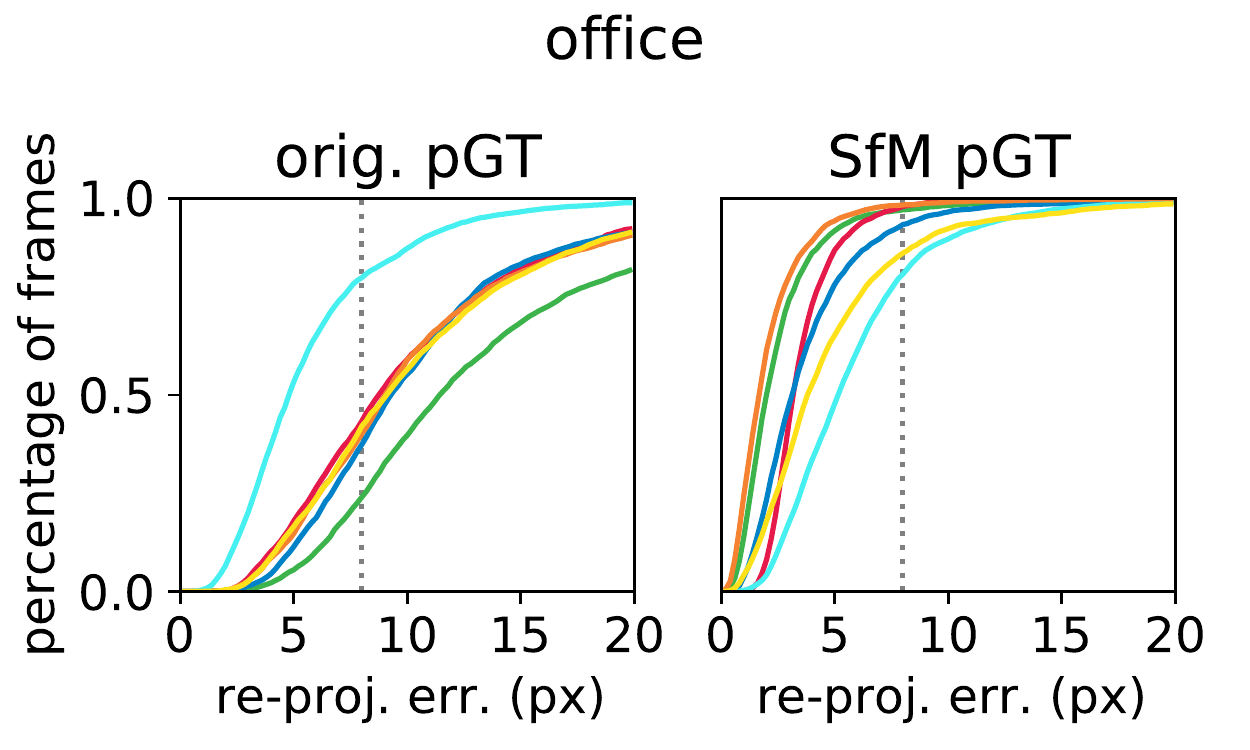}%
    \includegraphics[width=0.33\linewidth]{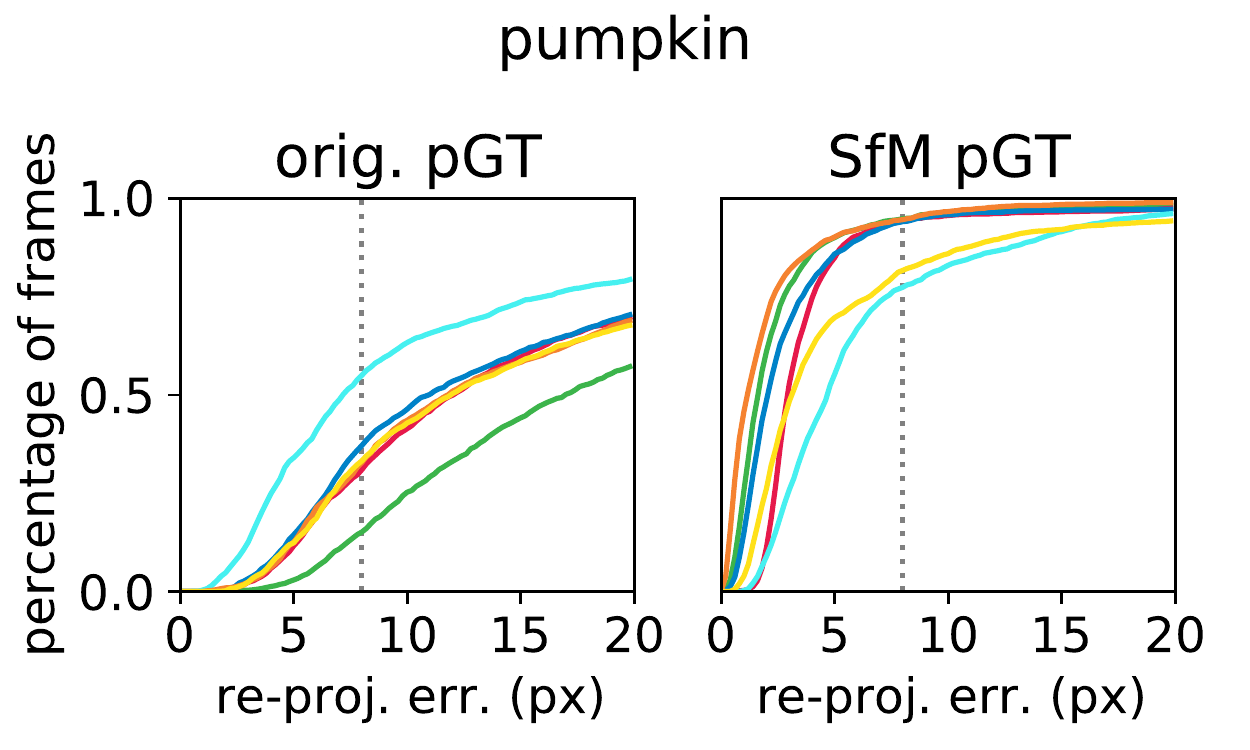}%
    \includegraphics[width=0.33\linewidth]{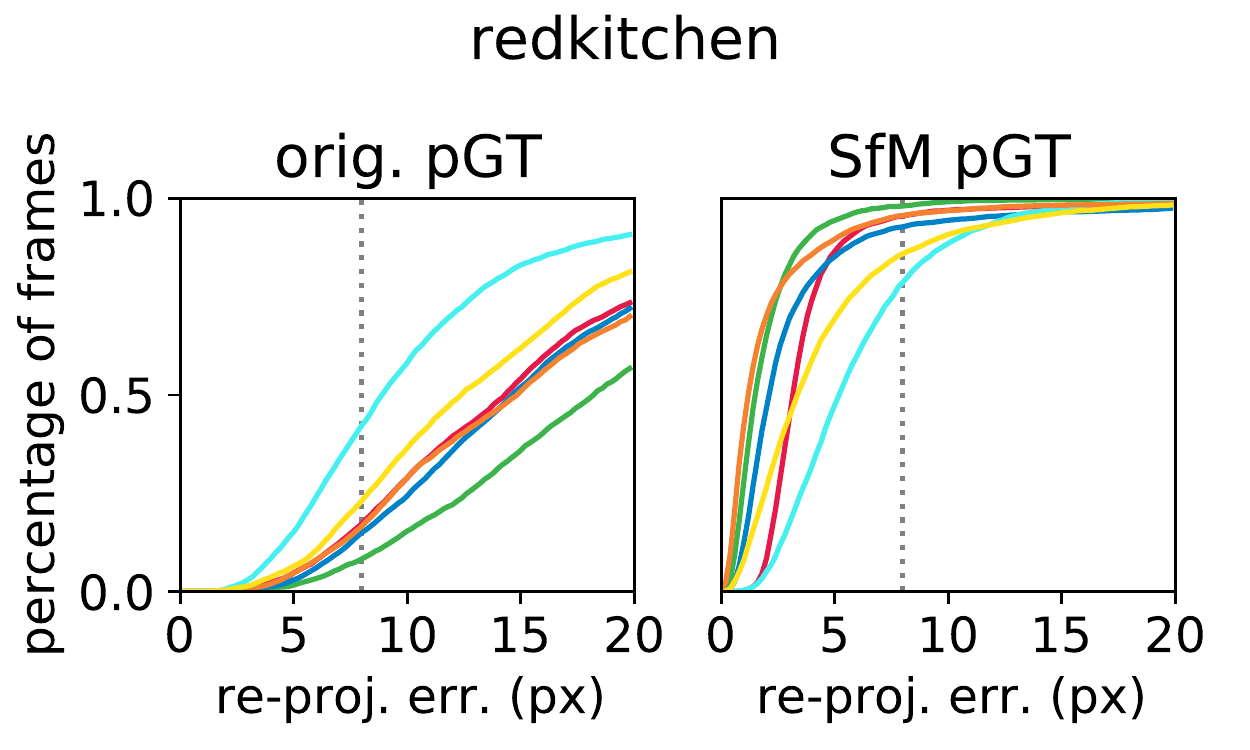}\\%
    \includegraphics[width=0.33\linewidth]{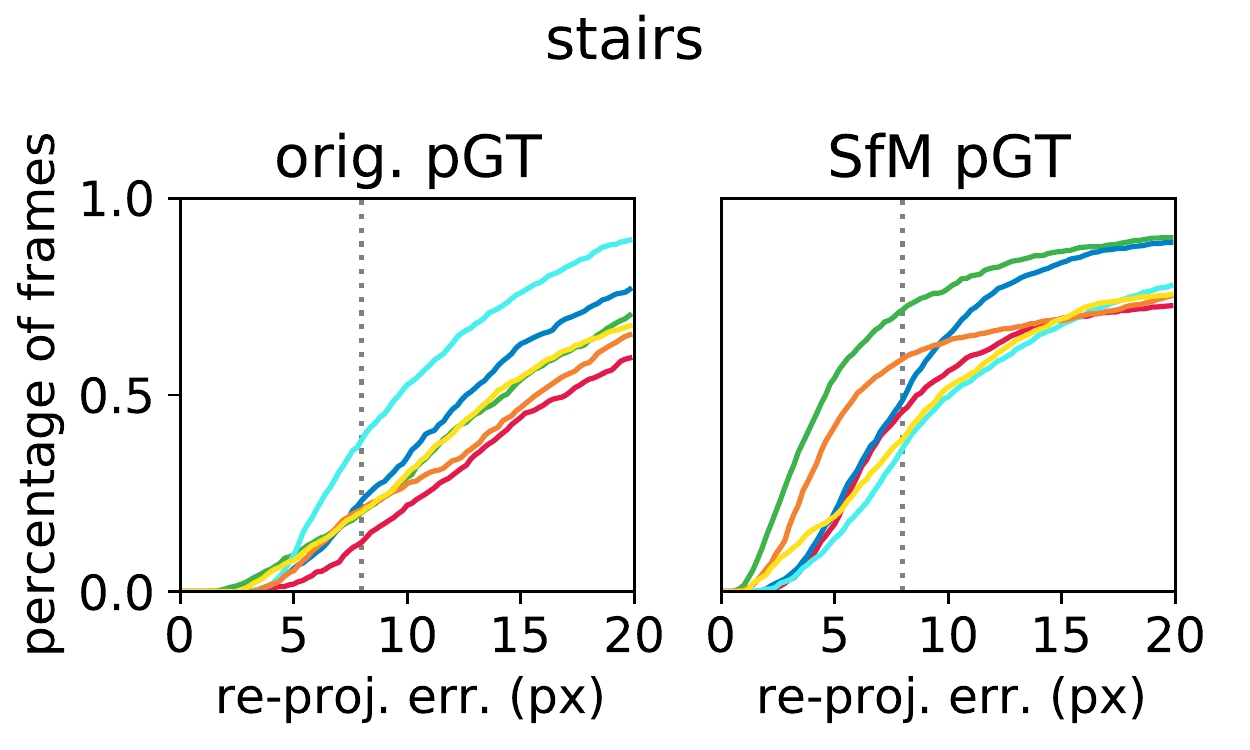}%
    \caption{\textbf{Max.~DCRE for 7Scenes.} We show cum.~distributions of the DCRE (Dense Correspondence Re-Projection Error \cite{Wald2020ECCV}) for all scenes of 7Scenes \cite{Shotton2013CVPR}, taking the max.~re-projection error per test image. The dotted line corresponds to 1\% of the image diagonal.}
    \label{fig:7s_max_plots}
\end{figure*}

\begin{figure*}[!t]
    \centering
    \includegraphics[width=0.33\linewidth]{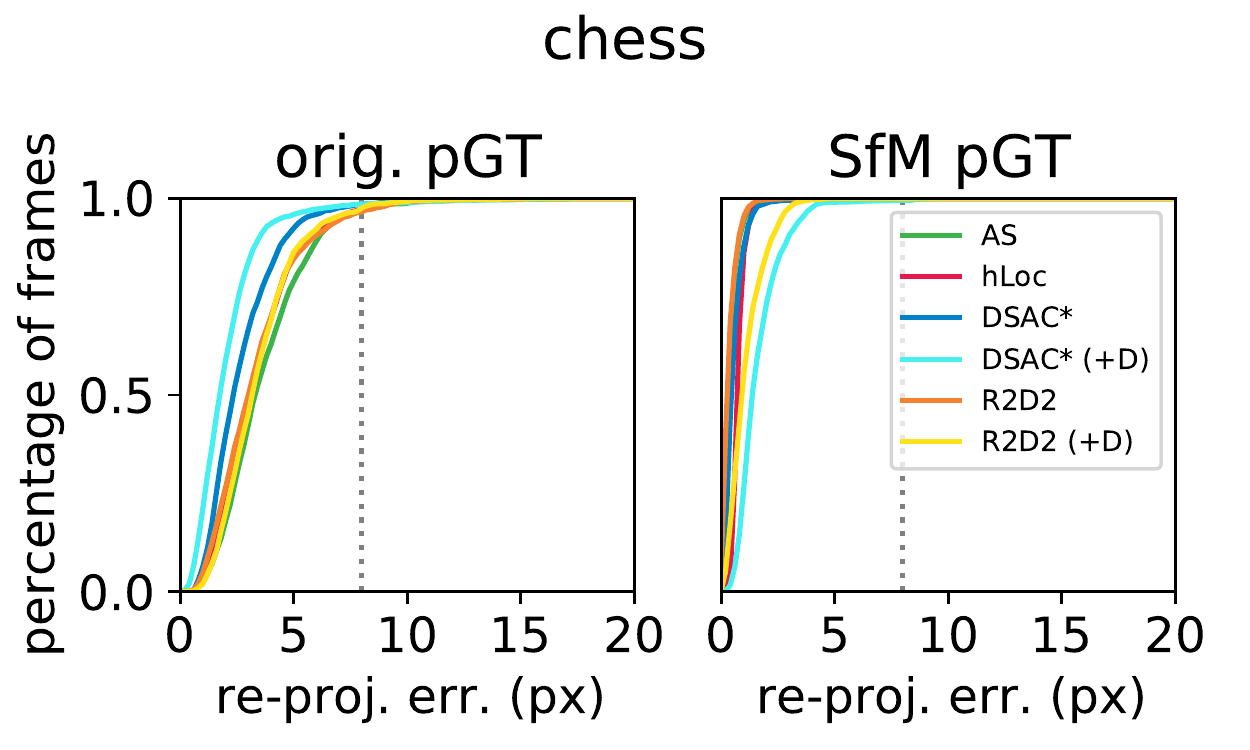}%
    \includegraphics[width=0.33\linewidth]{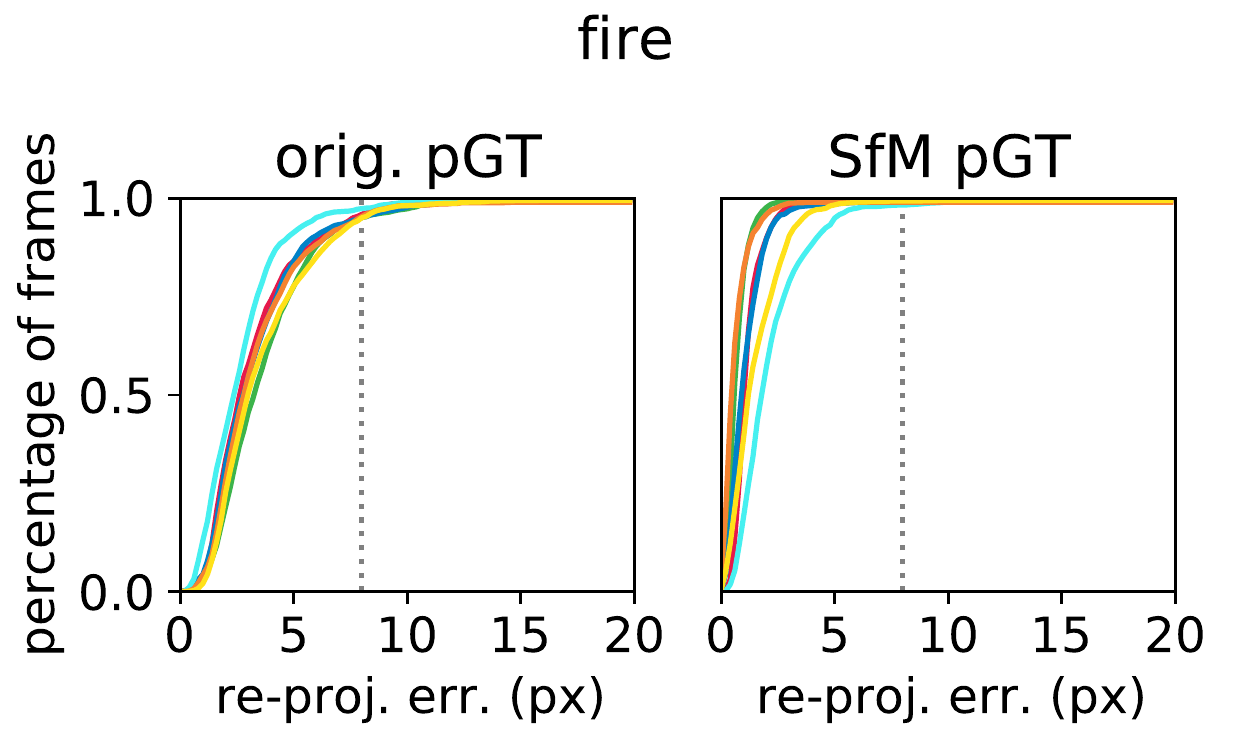}%
    \includegraphics[width=0.33\linewidth]{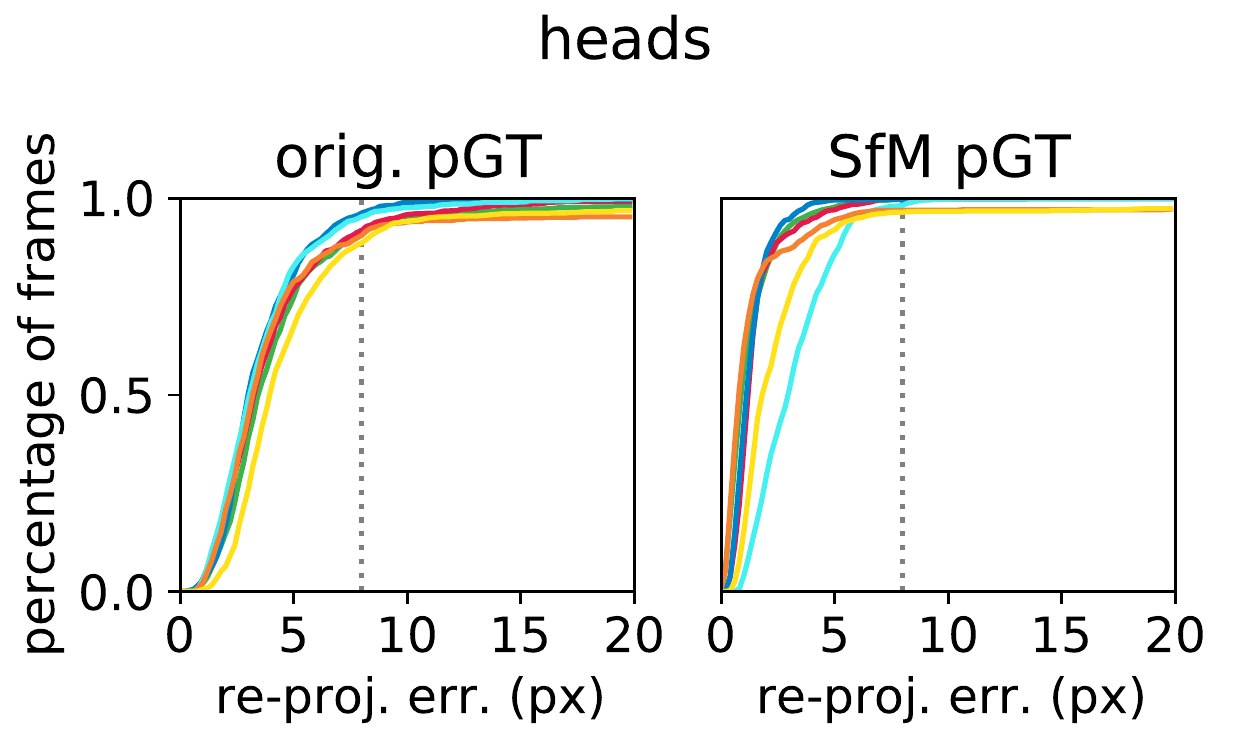}\\%
    \includegraphics[width=0.33\linewidth]{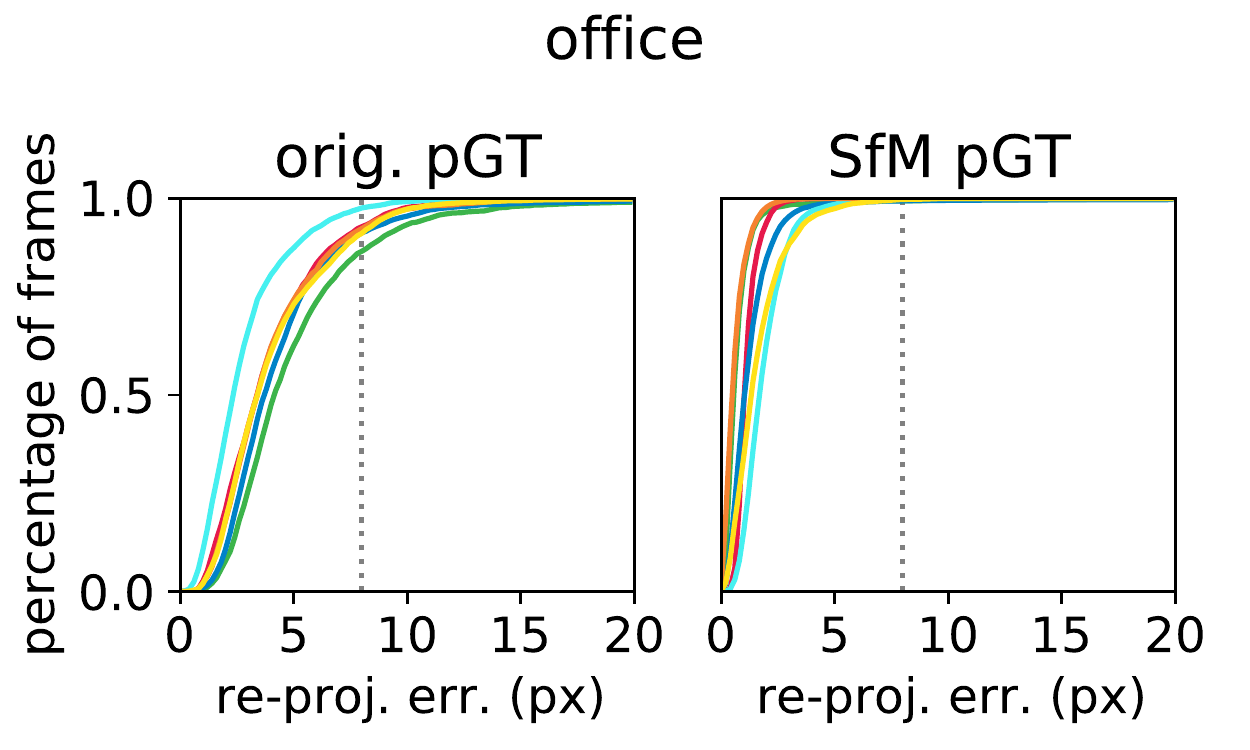}%
    \includegraphics[width=0.33\linewidth]{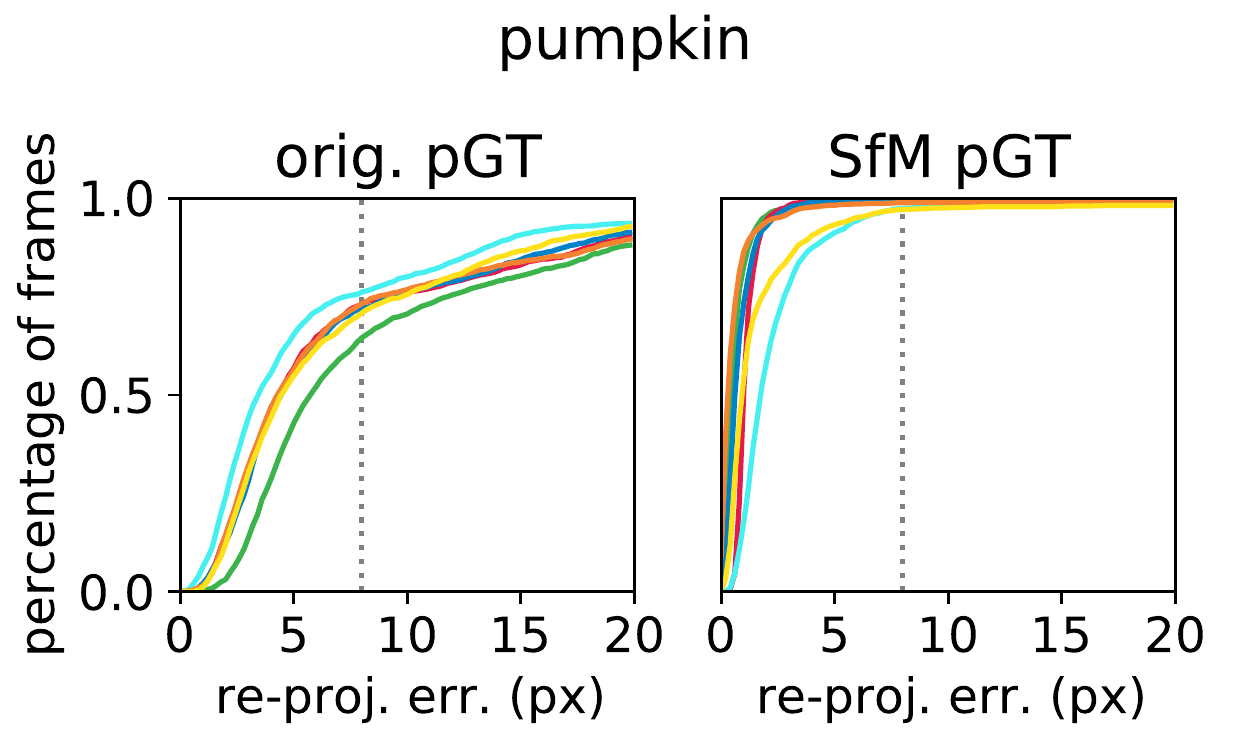}%
    \includegraphics[width=0.33\linewidth]{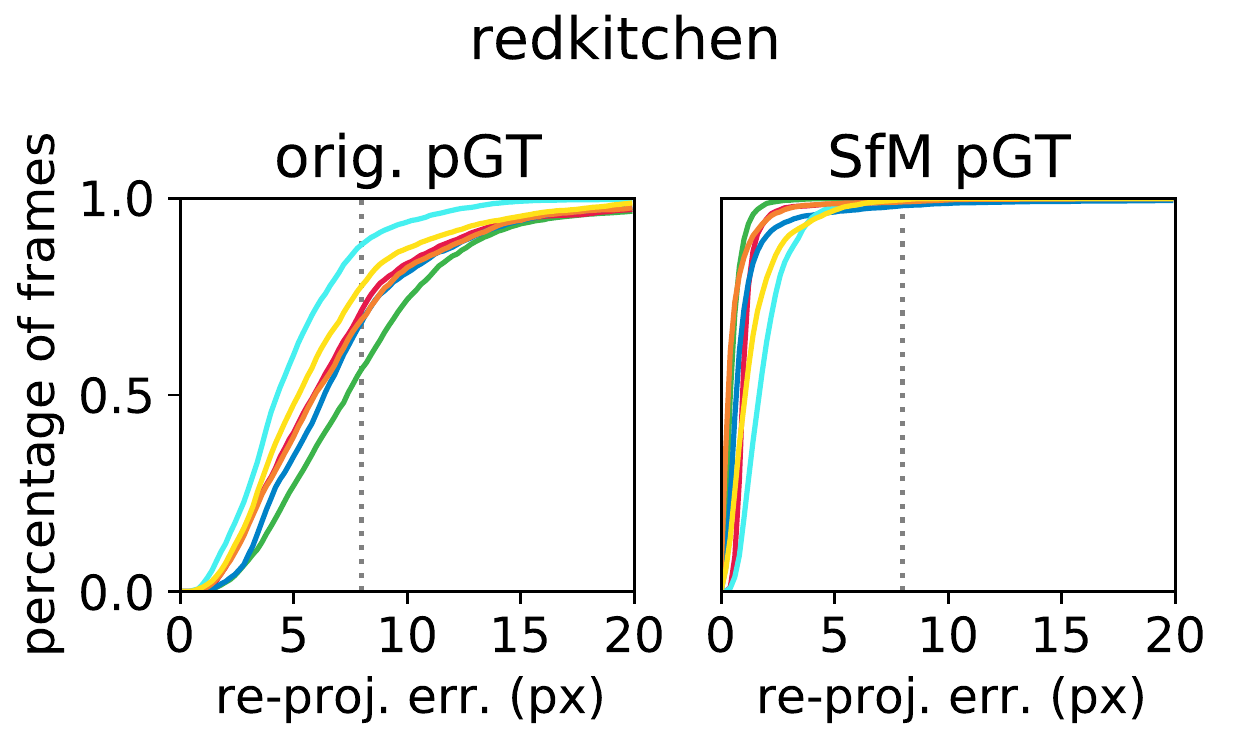}\\%
    \includegraphics[width=0.33\linewidth]{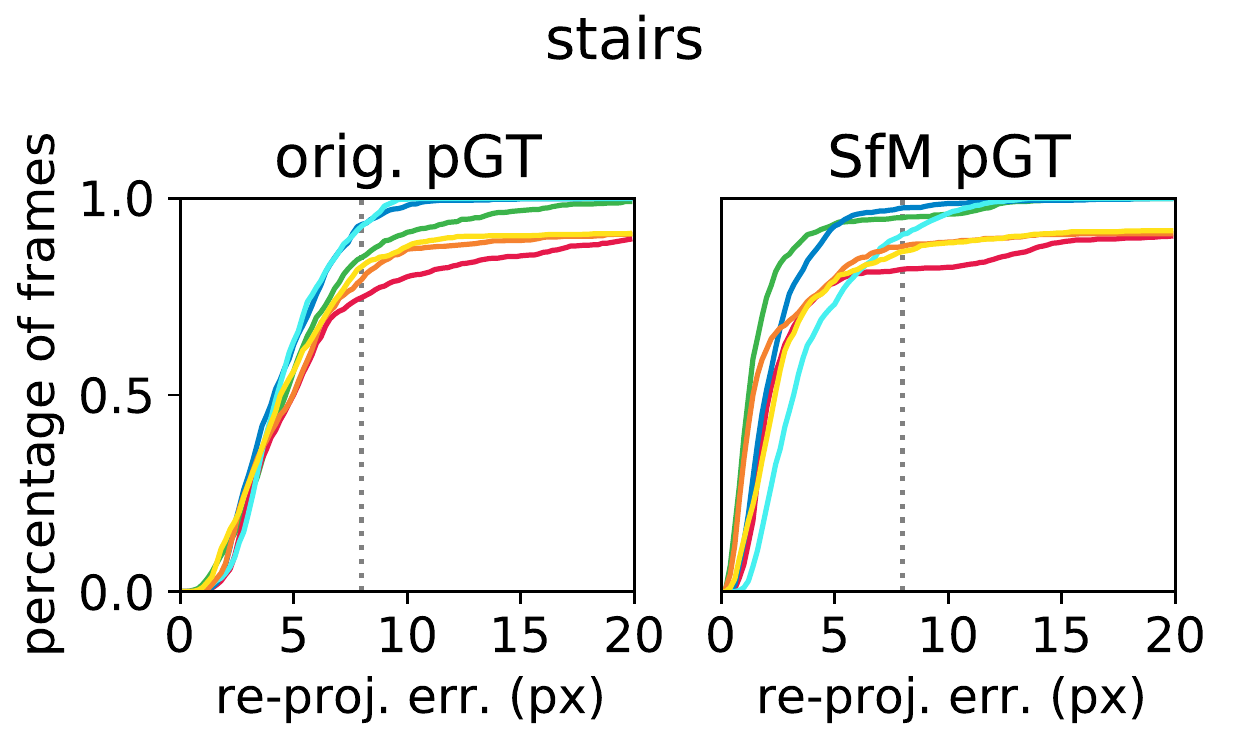}%
    \caption{\textbf{Mean DCRE for 7Scenes.} We show cum.~distributions of the DCRE (Dense Correspondence Re-Projection Error \cite{Wald2020ECCV}) for all scenes of 7Scenes \cite{Shotton2013CVPR}, taking the mean re-projection error per test image. The dotted line corresponds to 1\% of the image diagonal.}
    \label{fig:7s_mean_plots}
\end{figure*}

\begin{figure*}[!t]
    \centering
    \includegraphics[width=0.33\linewidth]{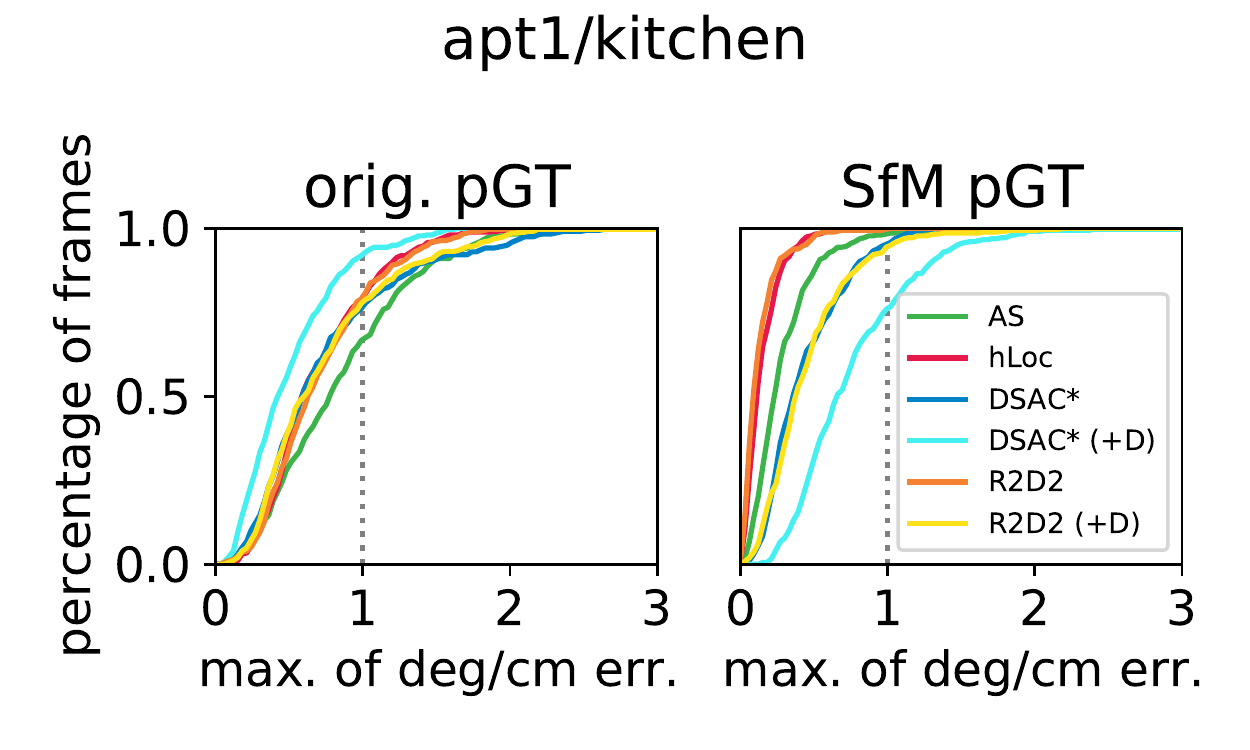}%
    \includegraphics[width=0.33\linewidth]{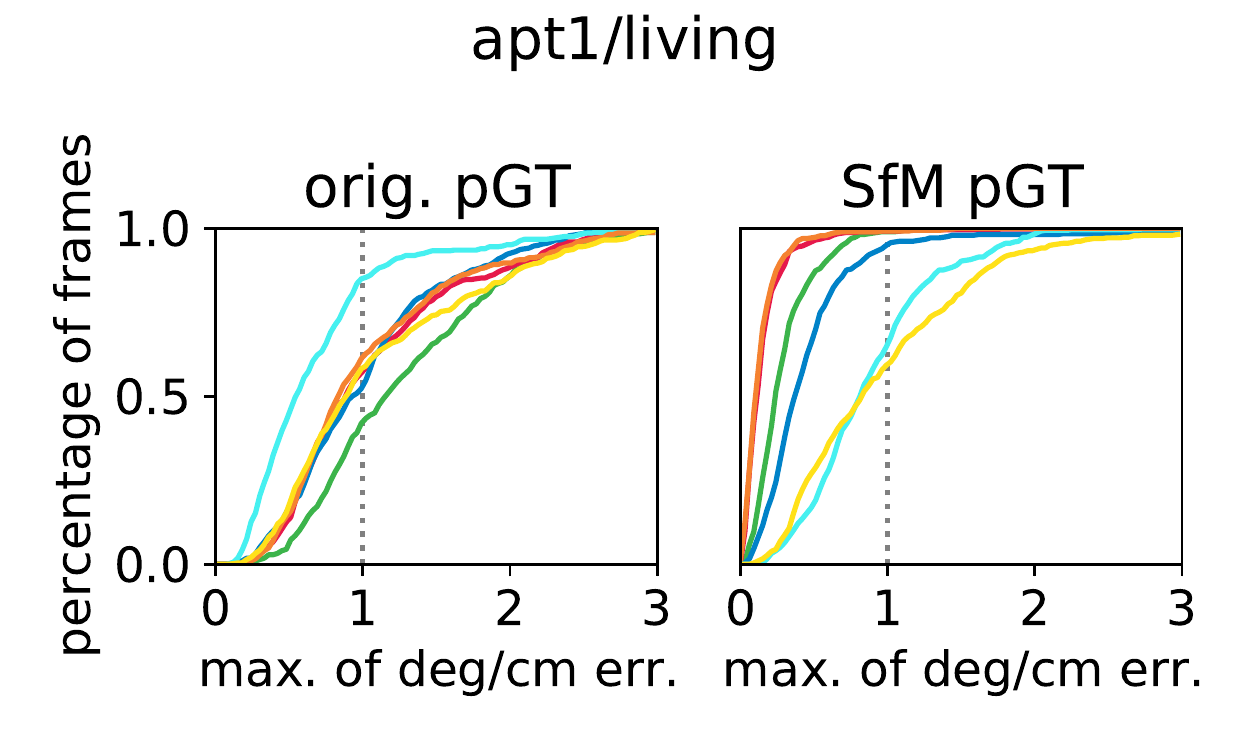}%
    \includegraphics[width=0.33\linewidth]{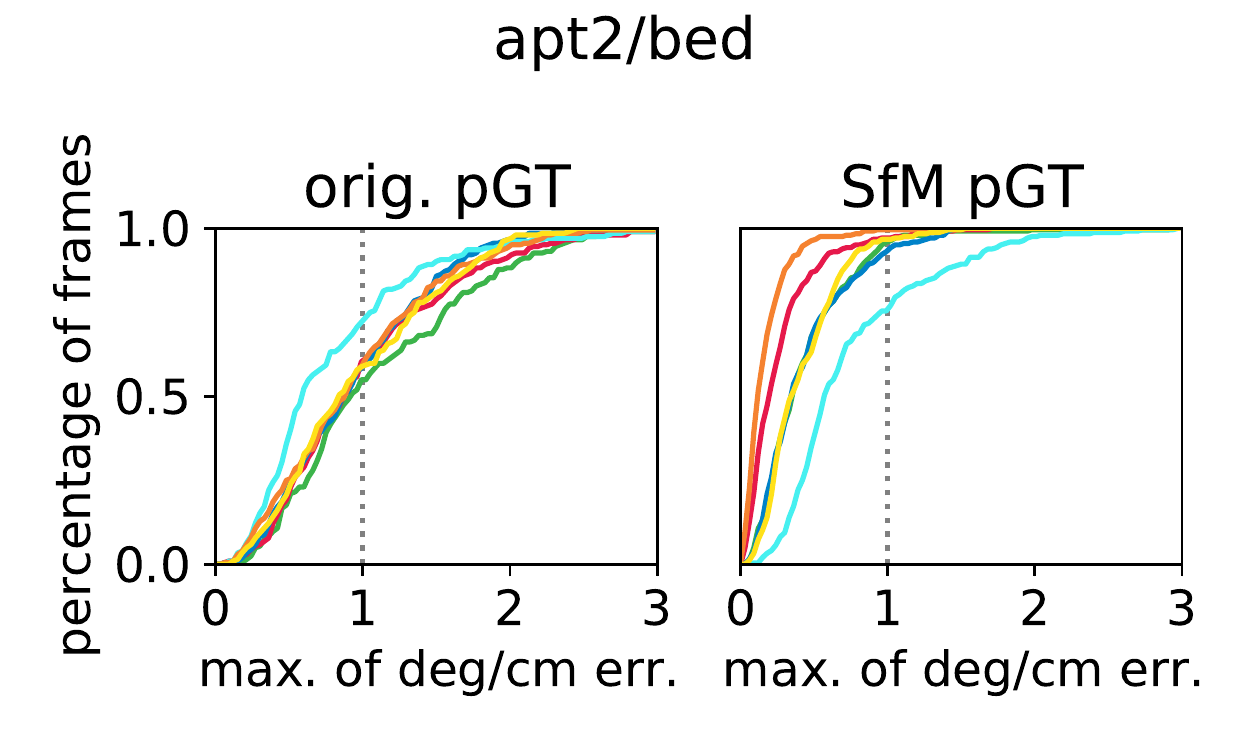}\\%
    \includegraphics[width=0.33\linewidth]{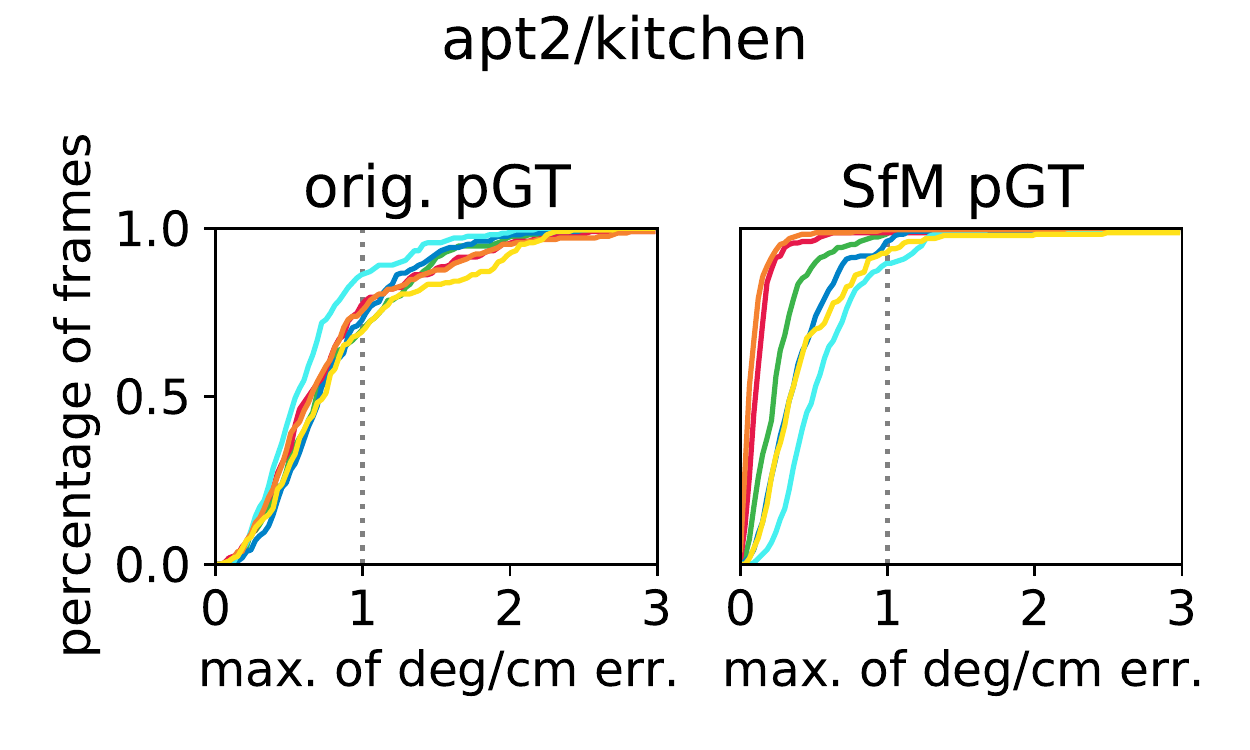}%
    \includegraphics[width=0.33\linewidth]{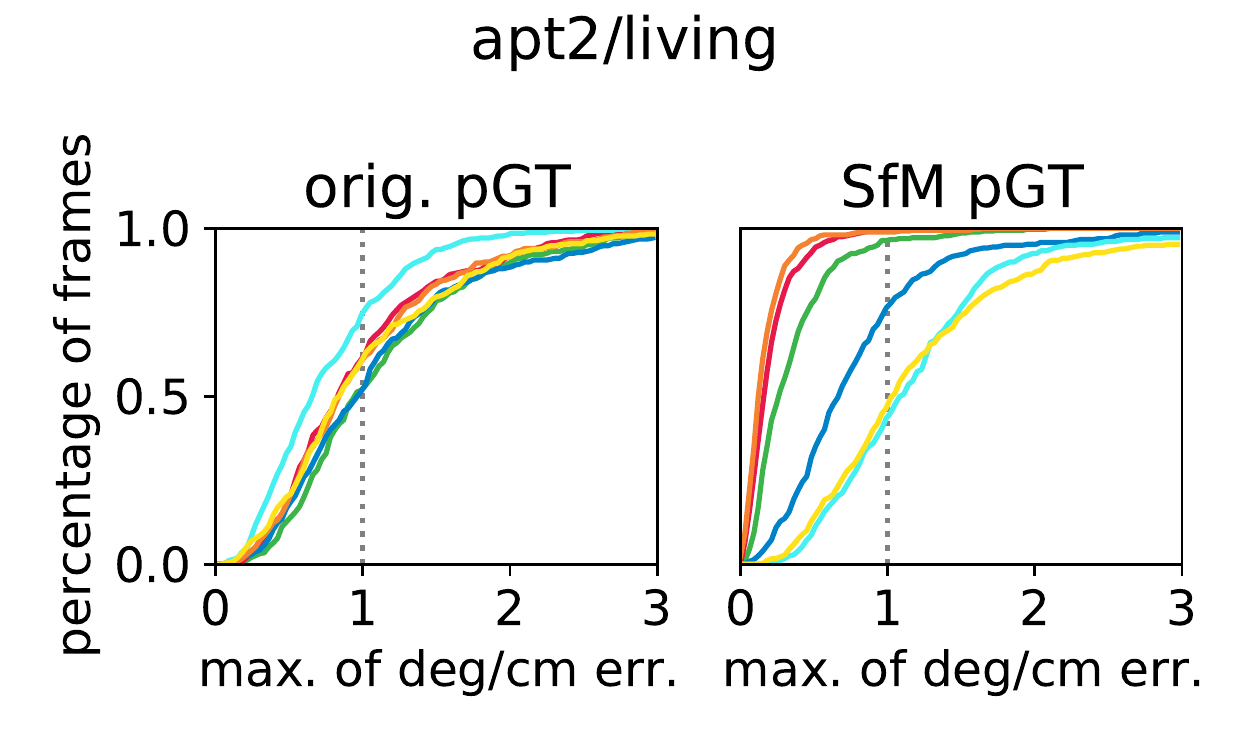}%
    \includegraphics[width=0.33\linewidth]{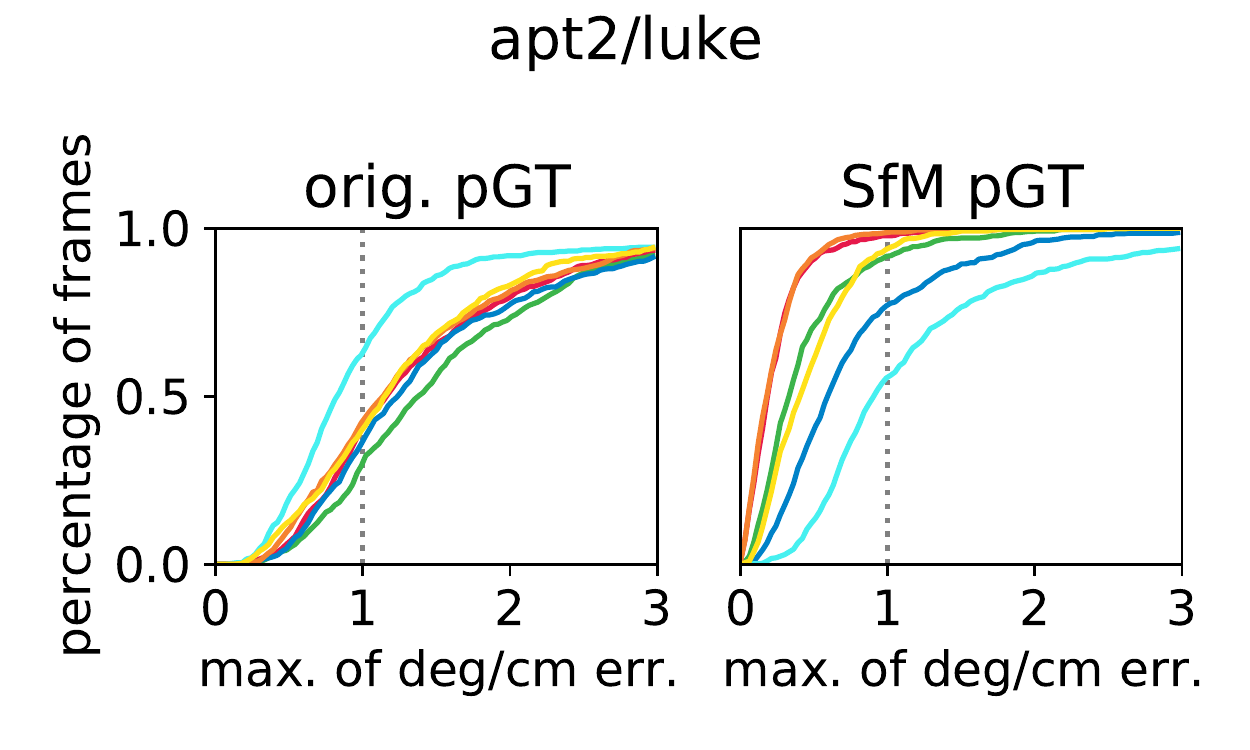}\\%
    \includegraphics[width=0.33\linewidth]{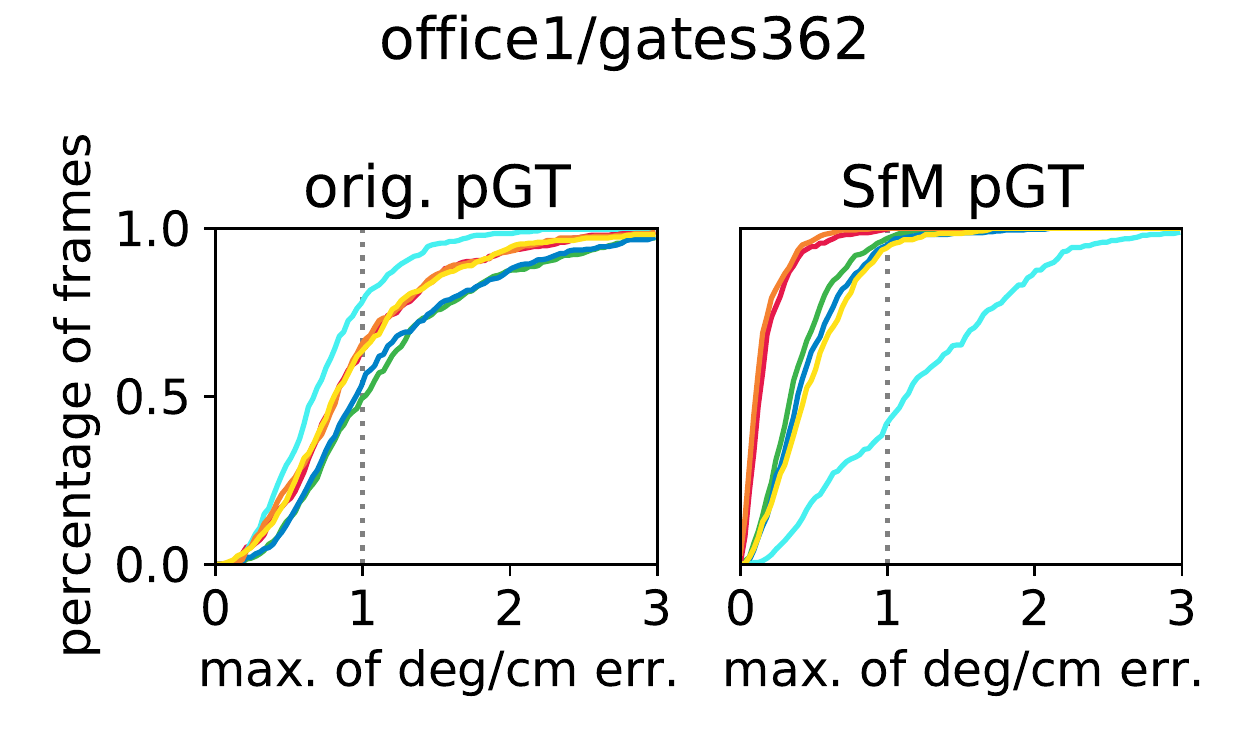}%
    \includegraphics[width=0.33\linewidth]{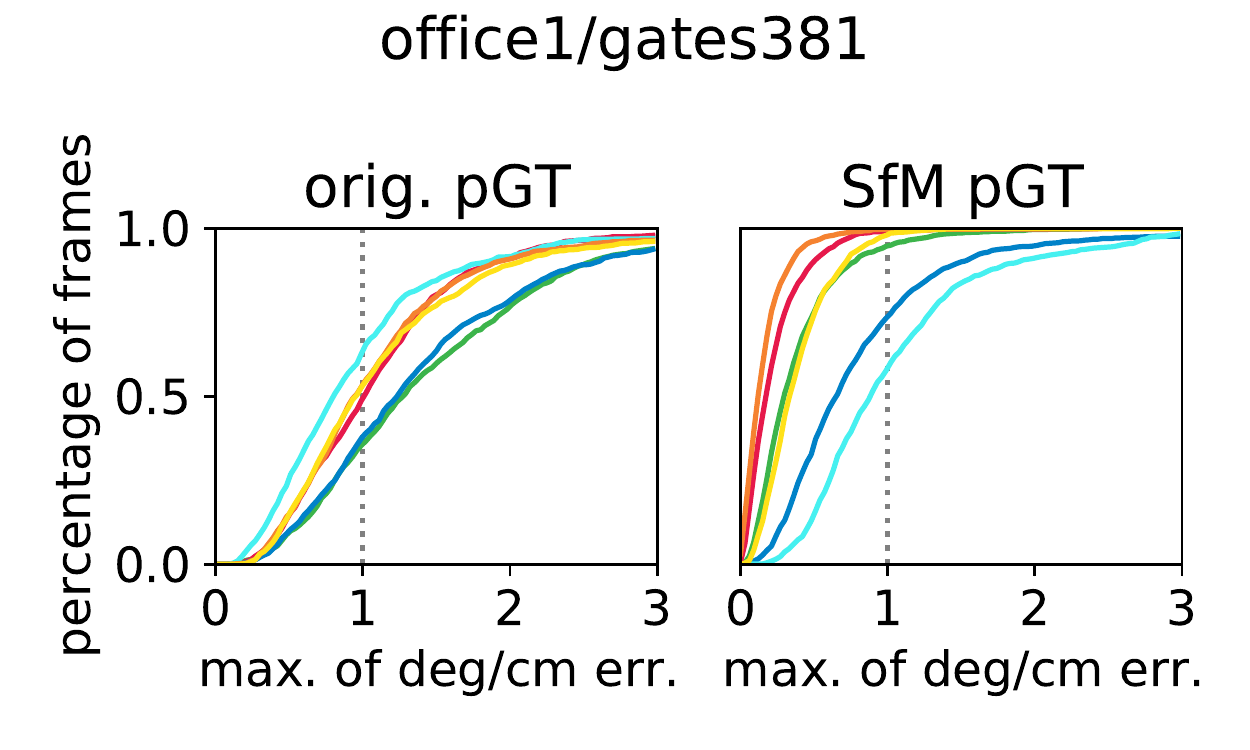}%
    \includegraphics[width=0.33\linewidth]{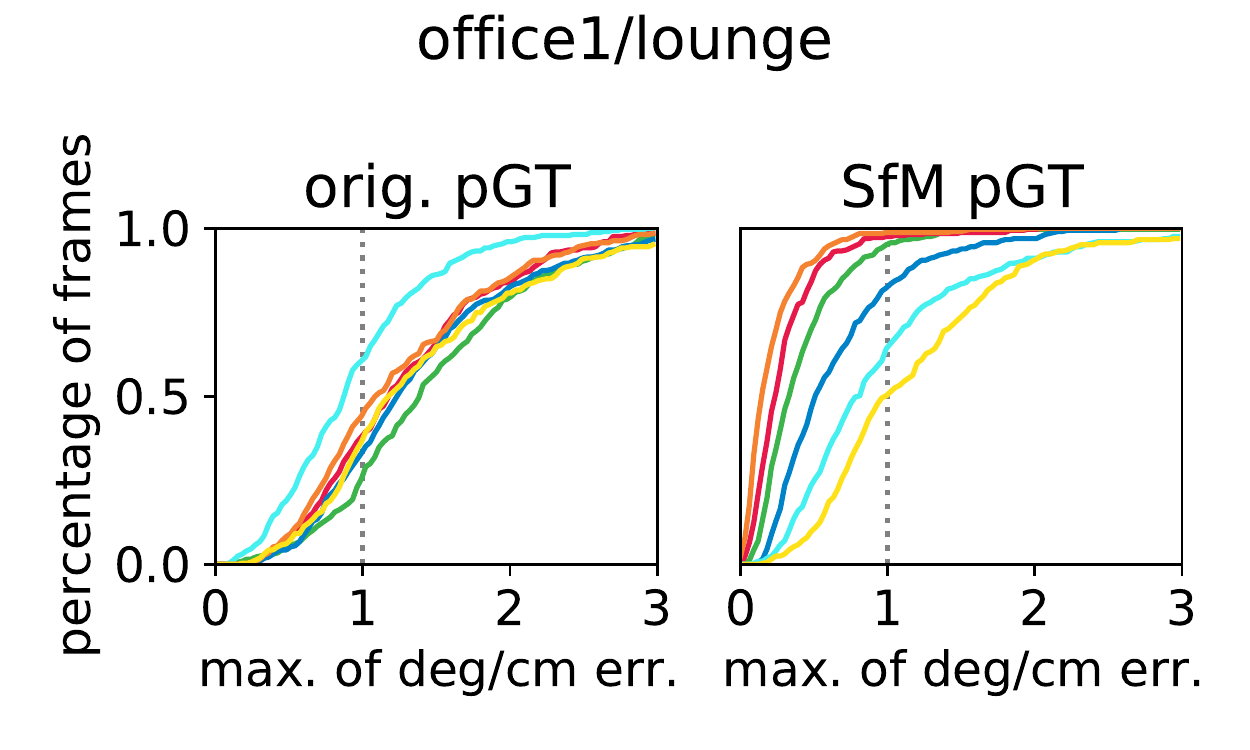}\\%
    \includegraphics[width=0.33\linewidth]{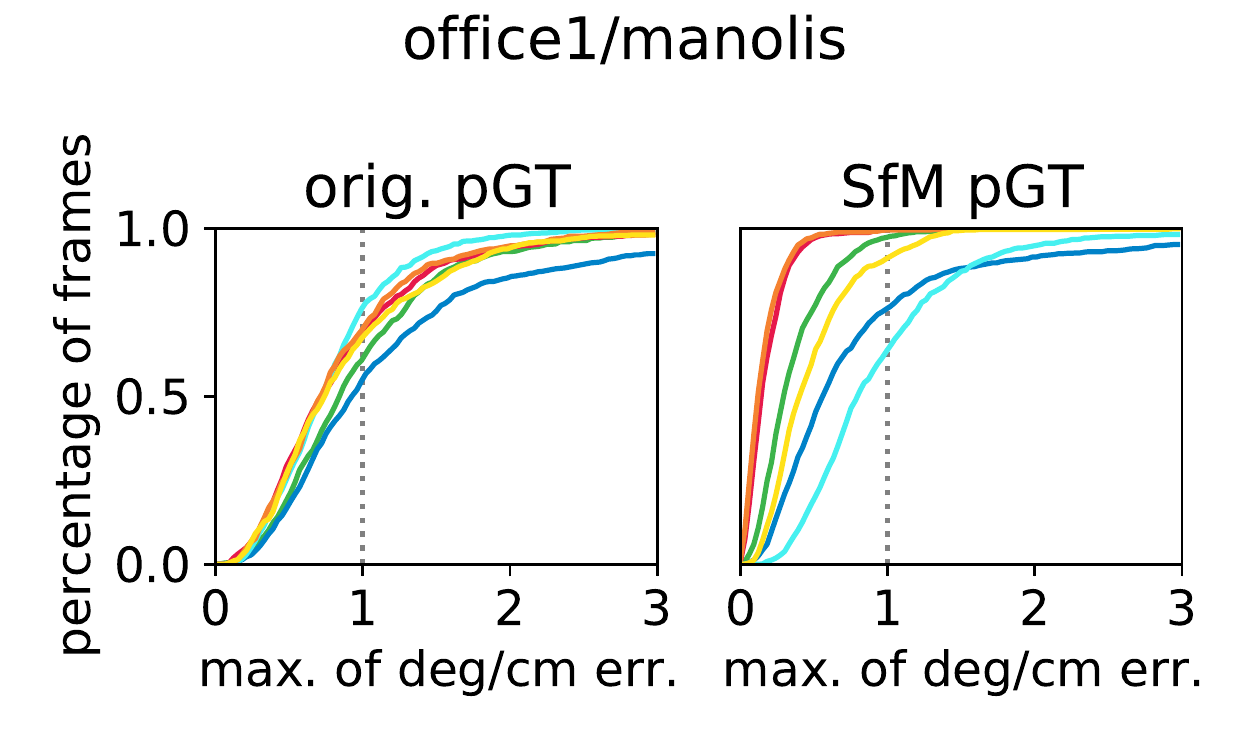}%
    \includegraphics[width=0.33\linewidth]{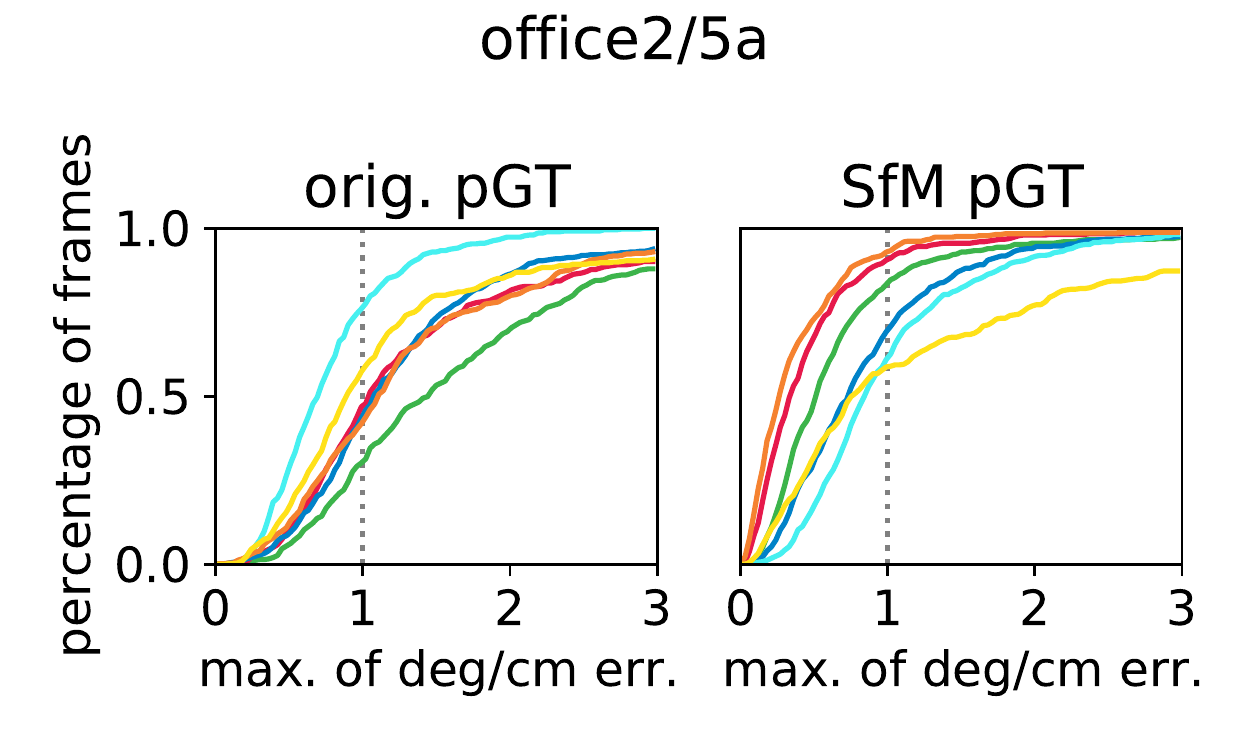}%
    \includegraphics[width=0.33\linewidth]{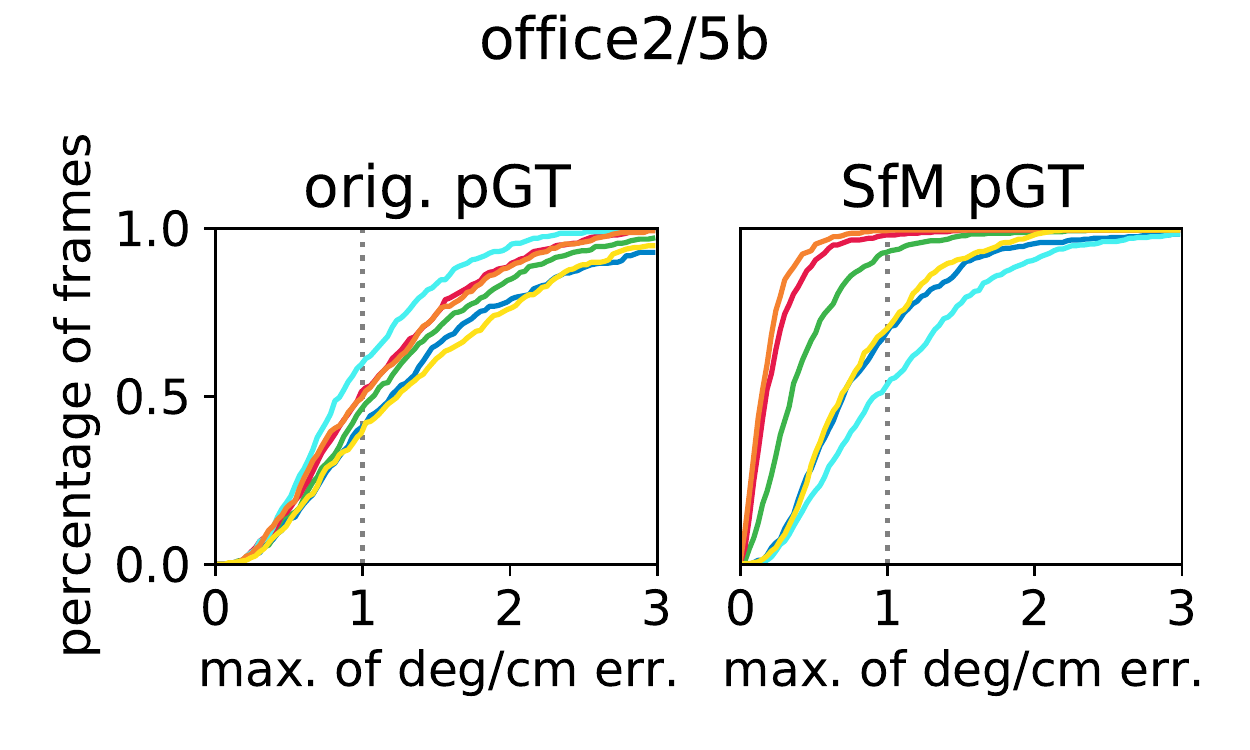}%
    \caption{\textbf{Pose error for 12Scenes.} We show cum.~distributions of the pose error (max.~of rotation and translation error) for all scenes of 12Scenes \cite{Valentin20163DV}. Dotted vertical lines correspond to a 1cm,1$^\circ$ threshold for reference.}
    \label{fig:12s_pose_plots}
\end{figure*}

\begin{figure*}[!t]
    \centering
    \includegraphics[width=0.33\linewidth]{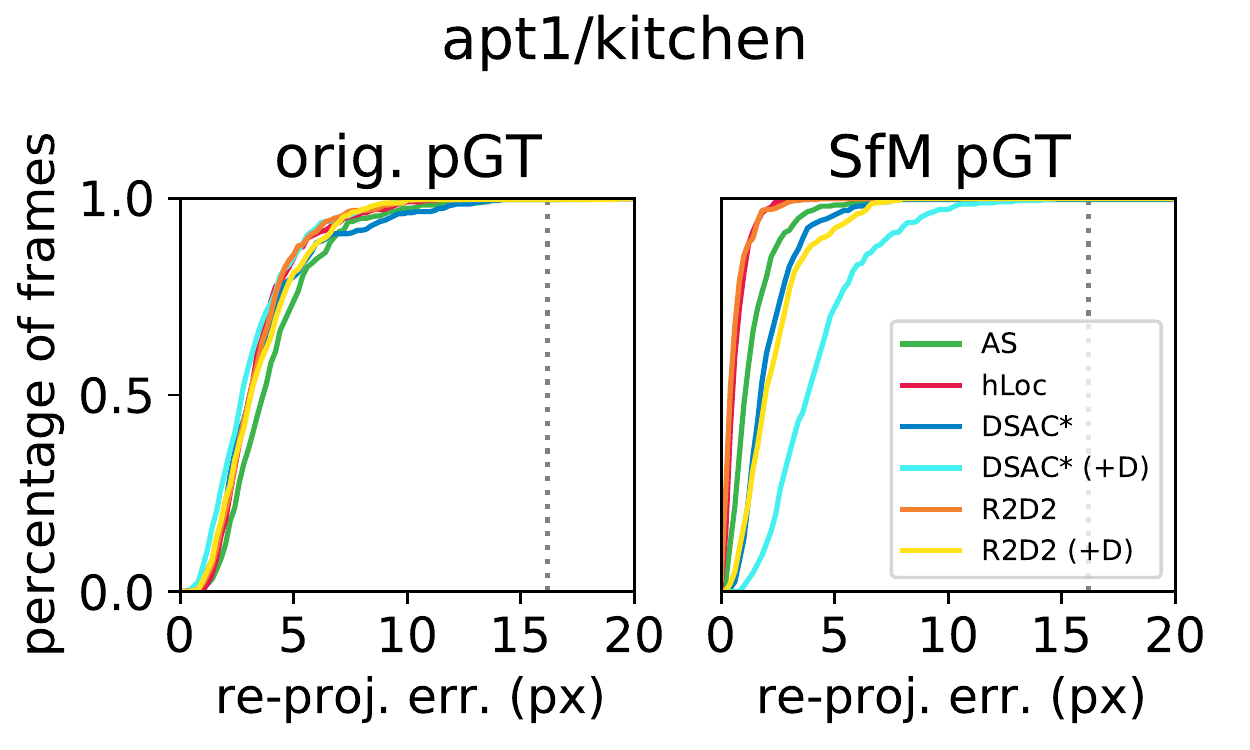}%
    \includegraphics[width=0.33\linewidth]{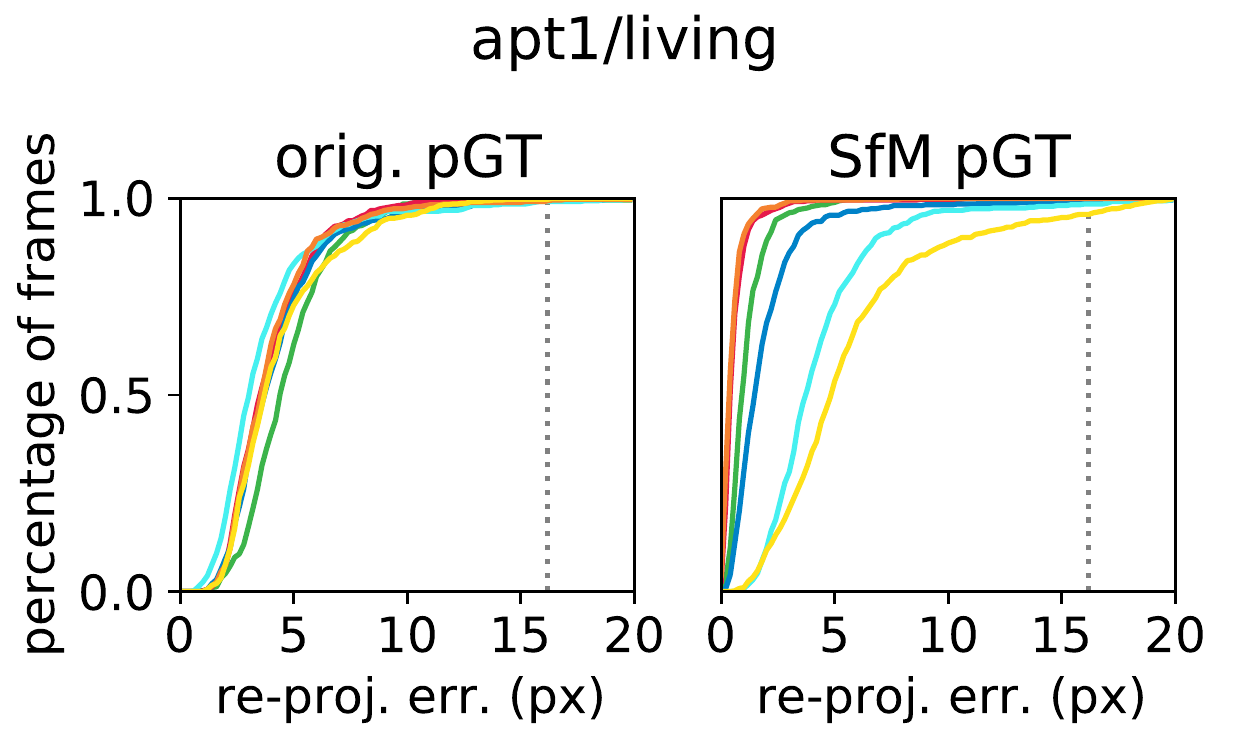}%
    \includegraphics[width=0.33\linewidth]{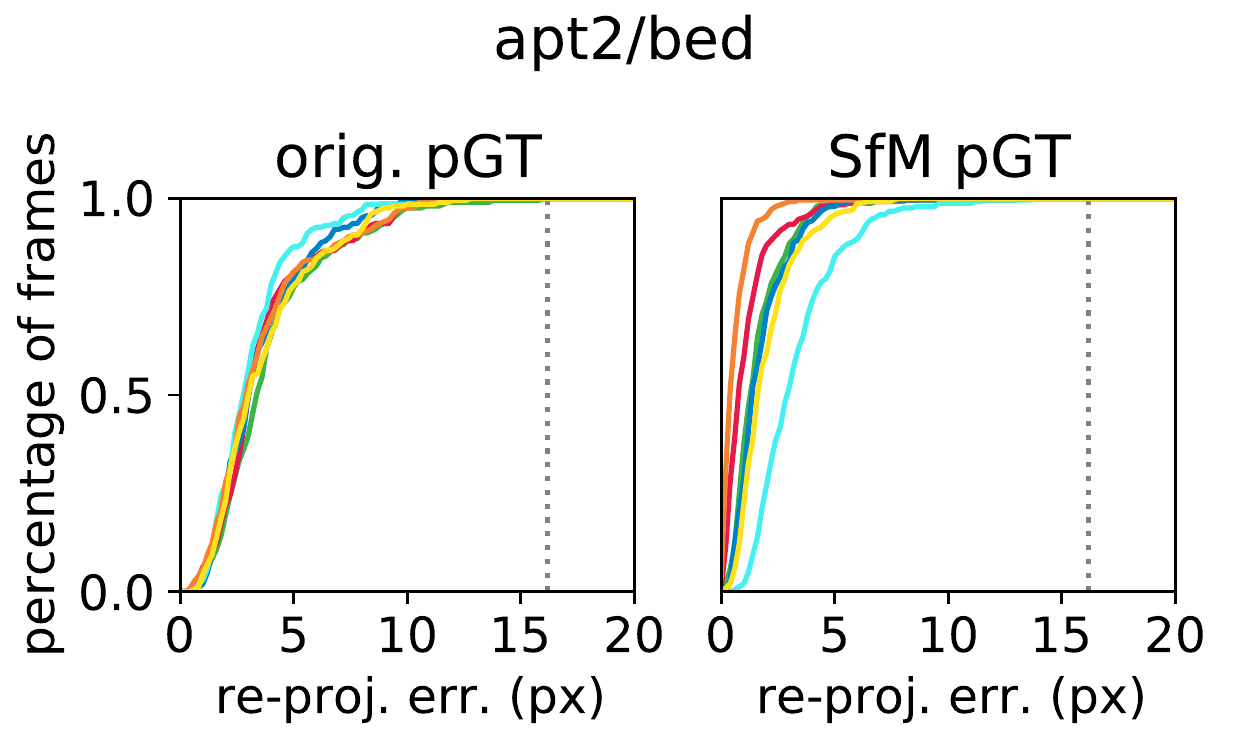}\\%
    \includegraphics[width=0.33\linewidth]{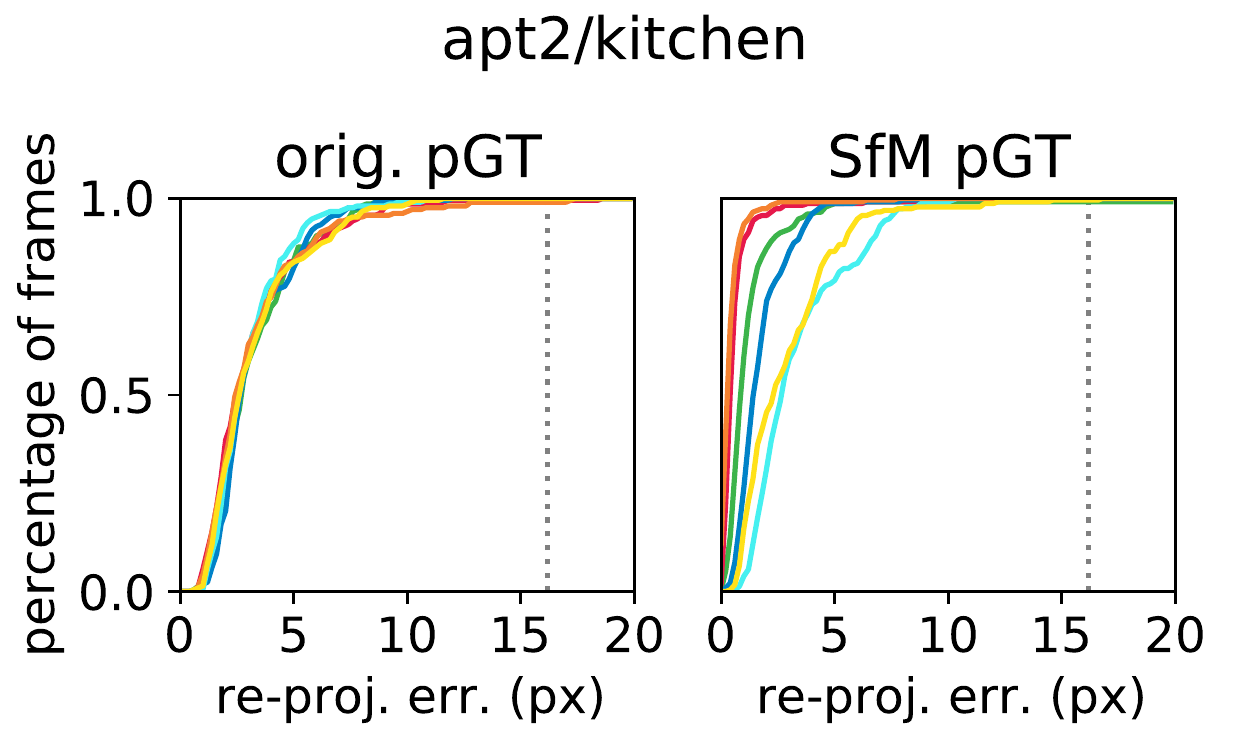}%
    \includegraphics[width=0.33\linewidth]{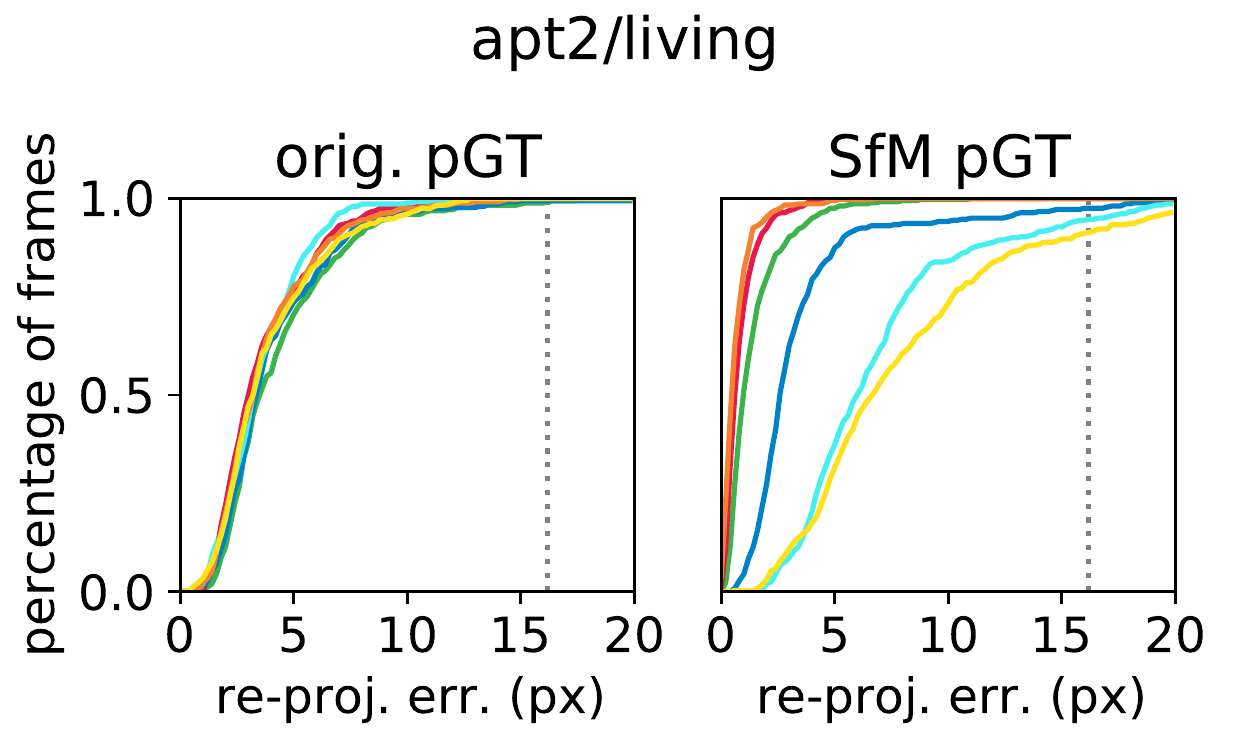}%
    \includegraphics[width=0.33\linewidth]{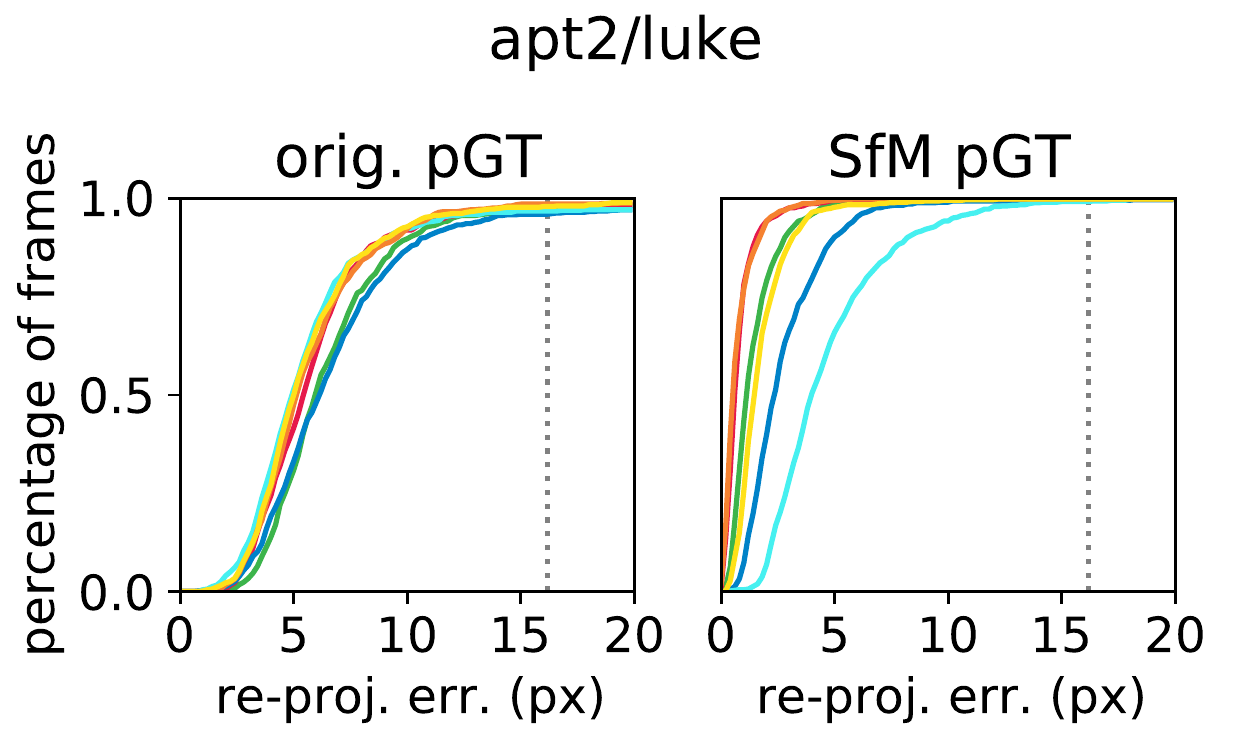}\\%
    \includegraphics[width=0.33\linewidth]{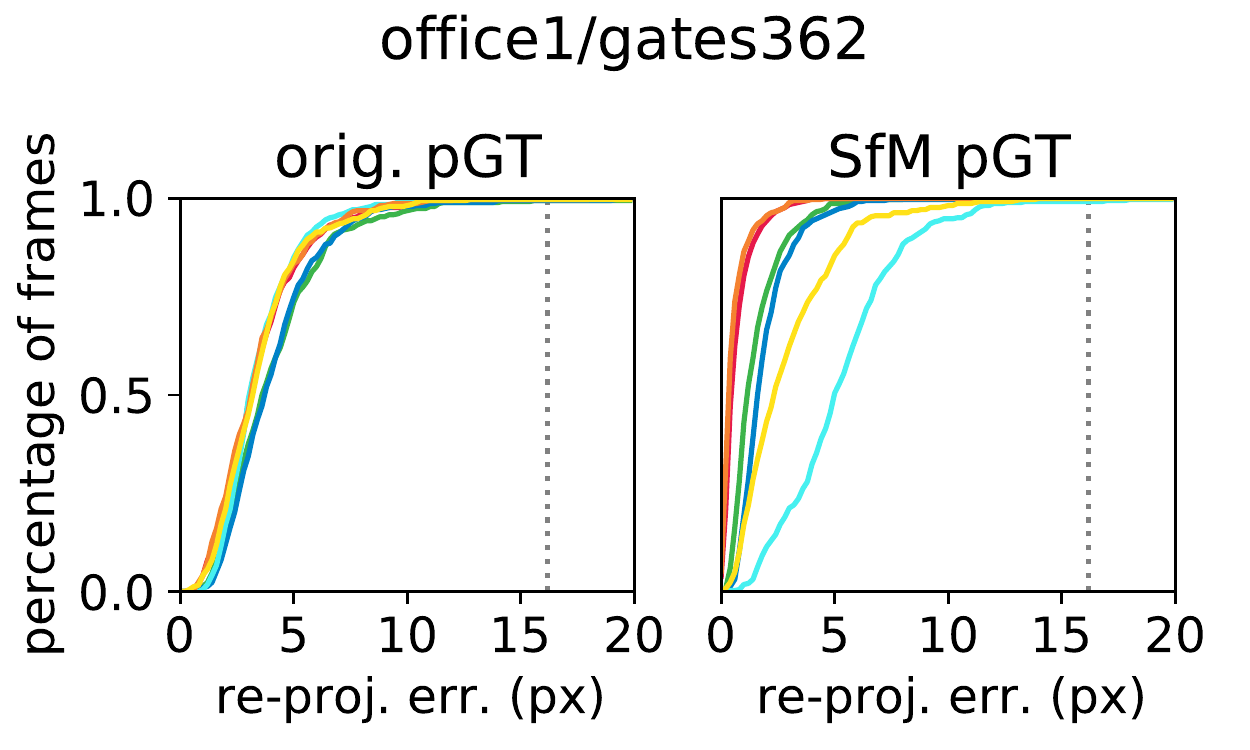}%
    \includegraphics[width=0.33\linewidth]{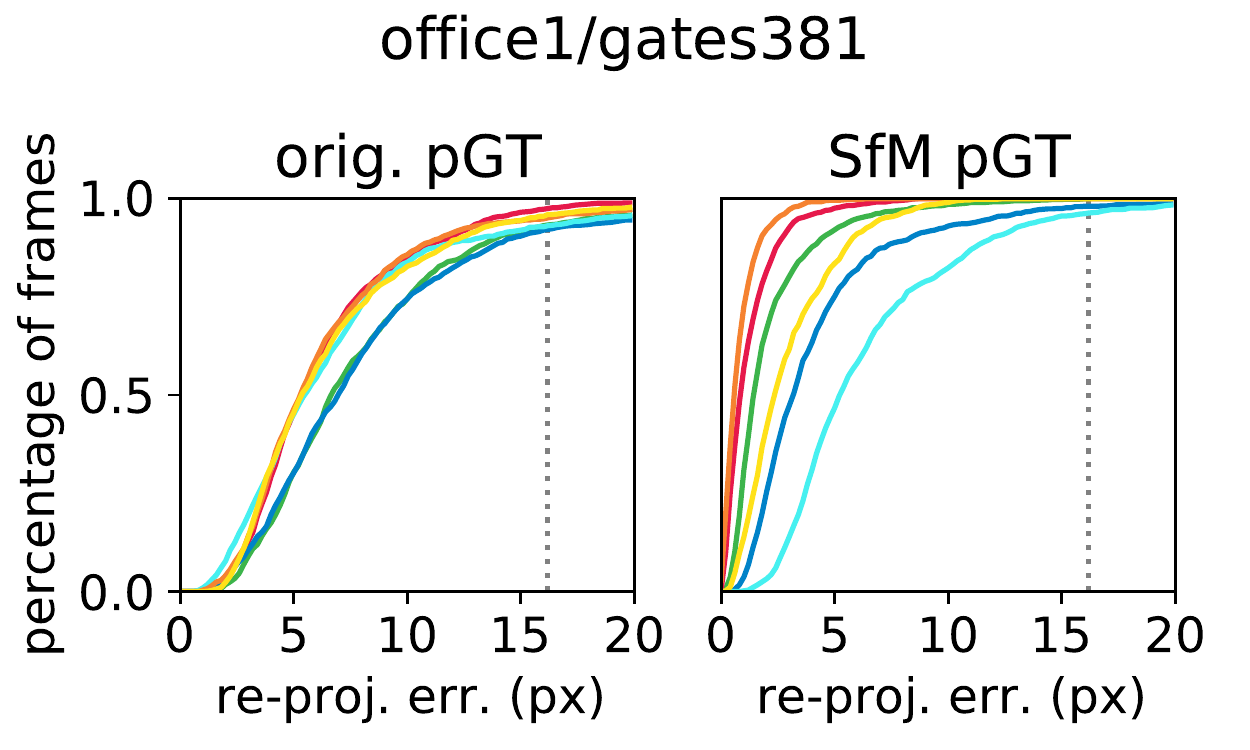}%
    \includegraphics[width=0.33\linewidth]{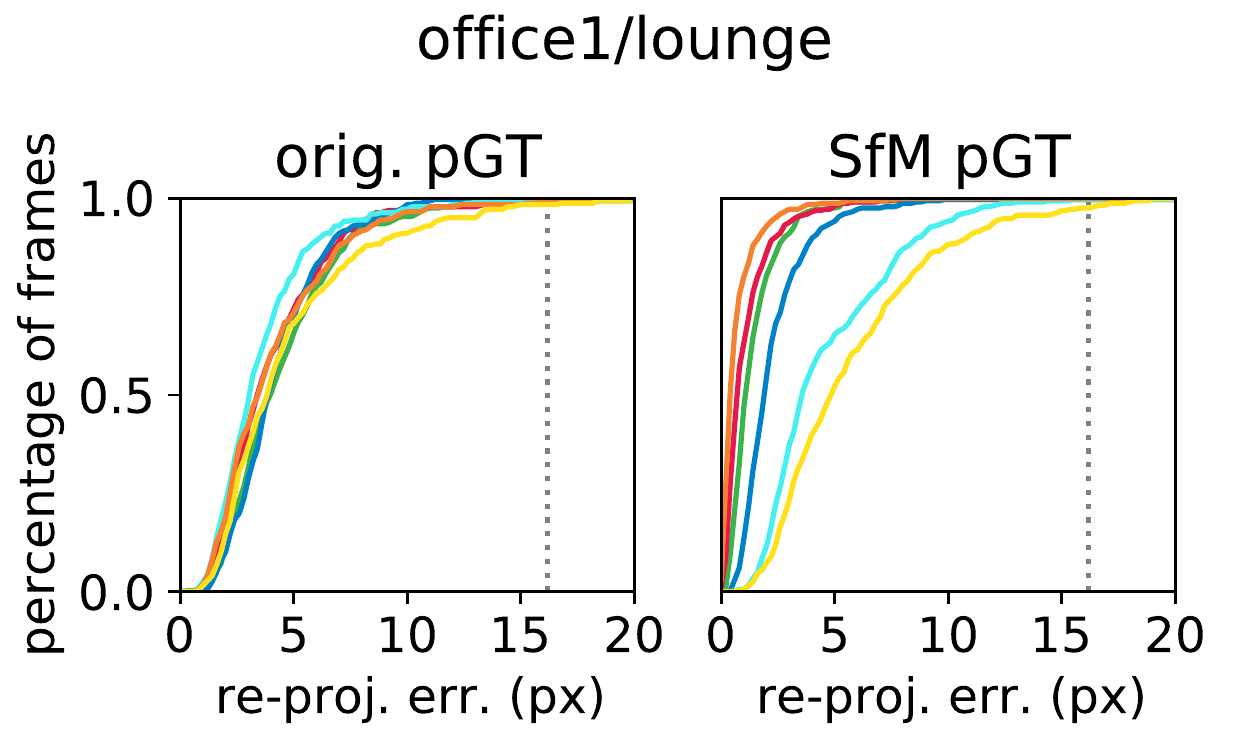}\\%
    \includegraphics[width=0.33\linewidth]{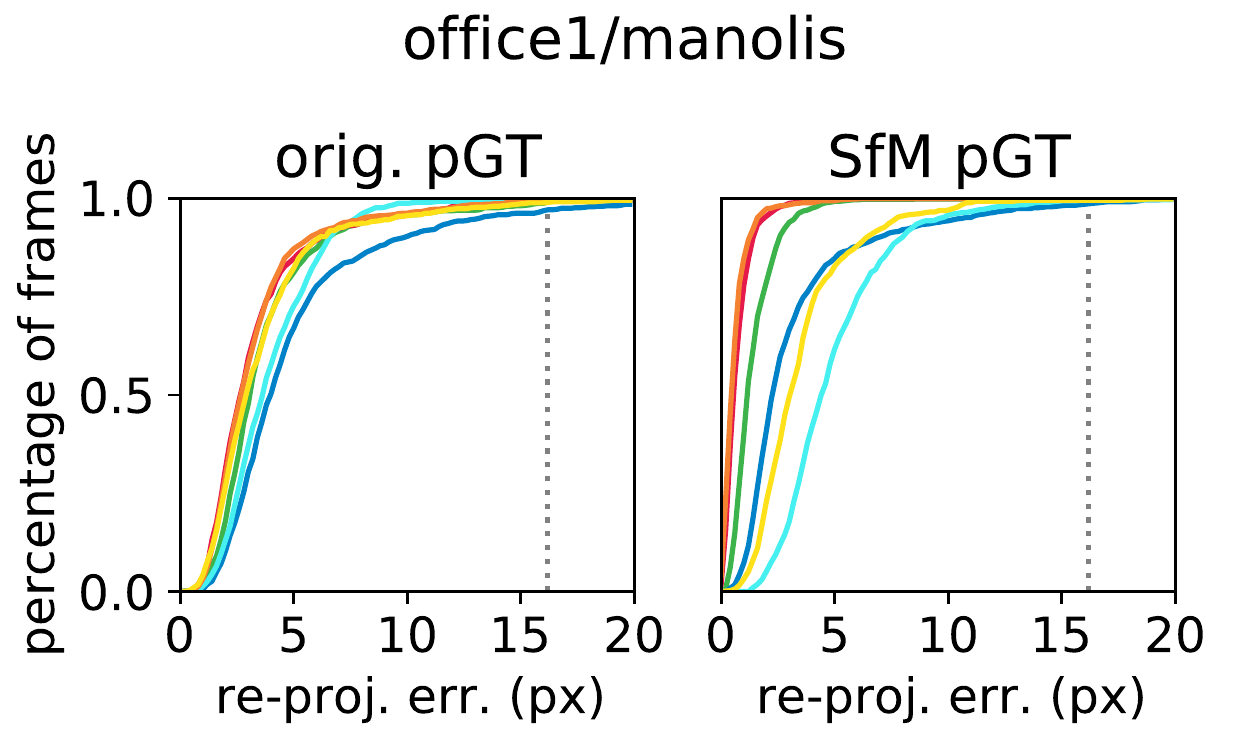}%
    \includegraphics[width=0.33\linewidth]{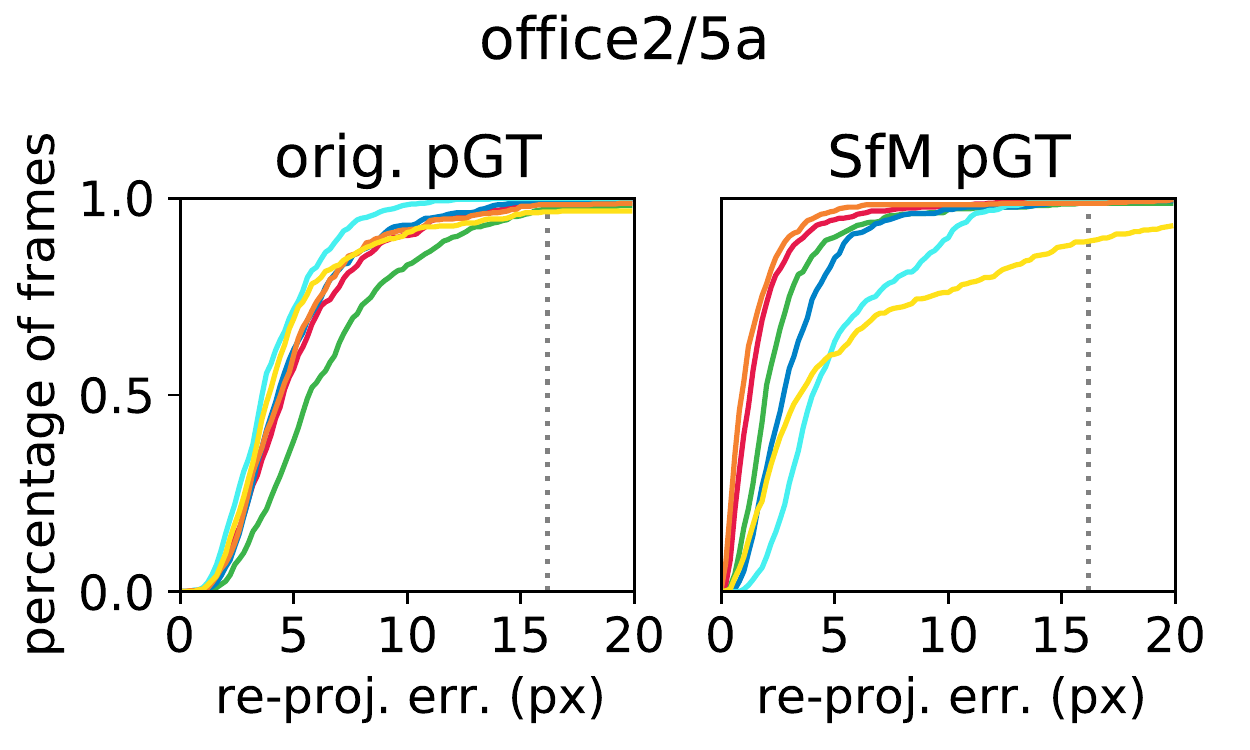}%
    \includegraphics[width=0.33\linewidth]{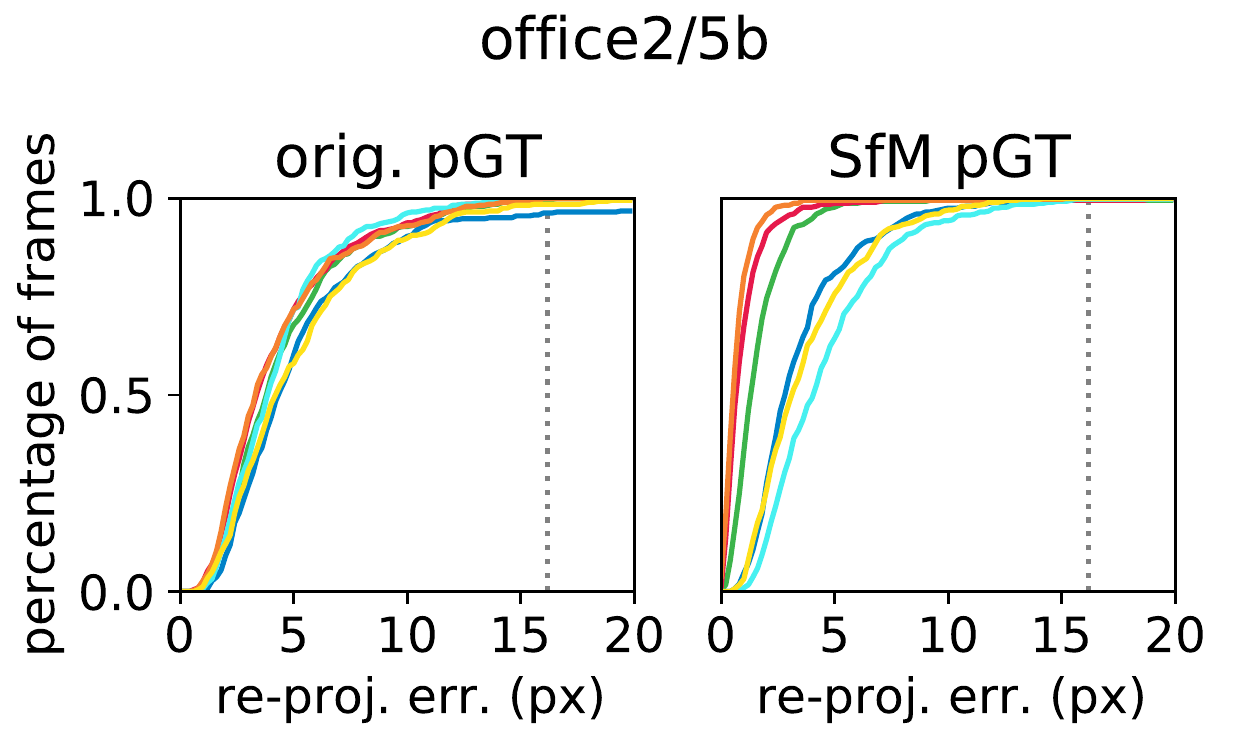}%
    \caption{\textbf{Max.~DCRE for 12Scenes.} We show cum.~distributions of the DCRE (Dense Correspondence Re-Projection Error \cite{Wald2020ECCV}) for all scenes of 12Scenes \cite{Valentin20163DV}, taking the max.~re-projection error per test image. The dotted vertical line corresponds to 1\% of the image diagonal.}
    \label{fig:12s_max_plots}
\end{figure*}

\begin{figure*}[!t]
    \centering
    \includegraphics[width=0.33\linewidth]{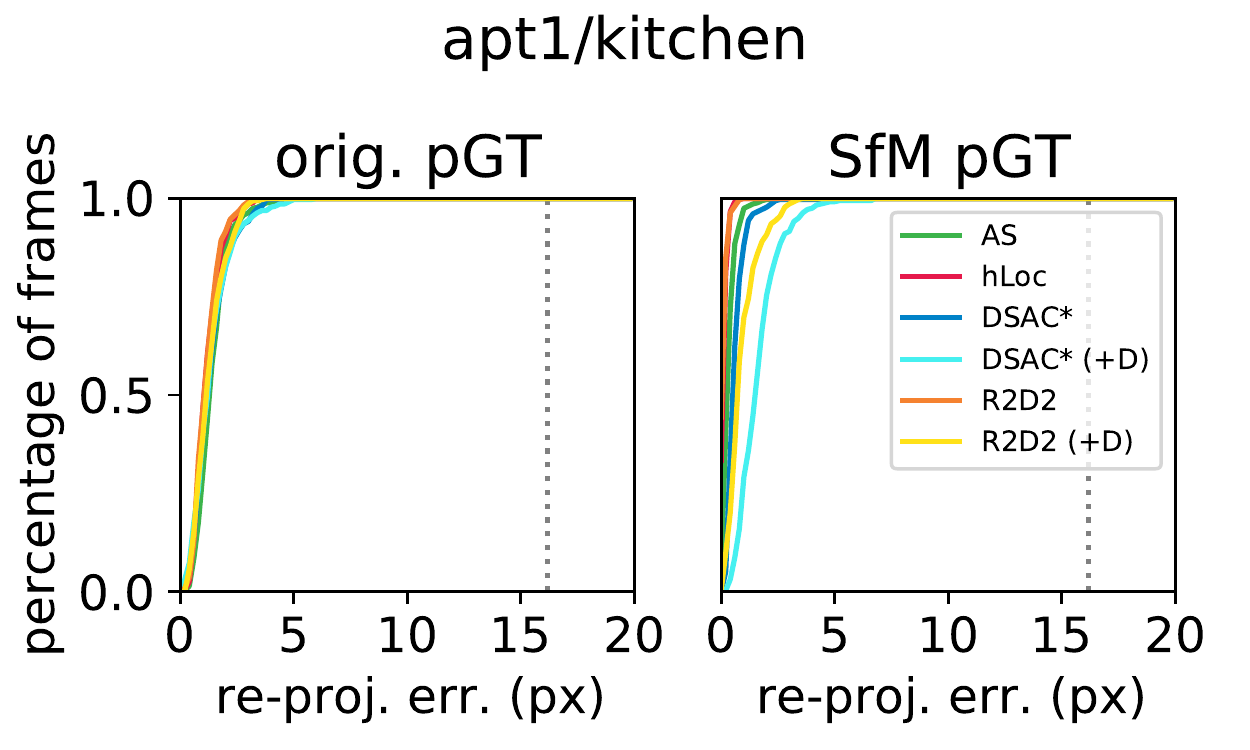}%
    \includegraphics[width=0.33\linewidth]{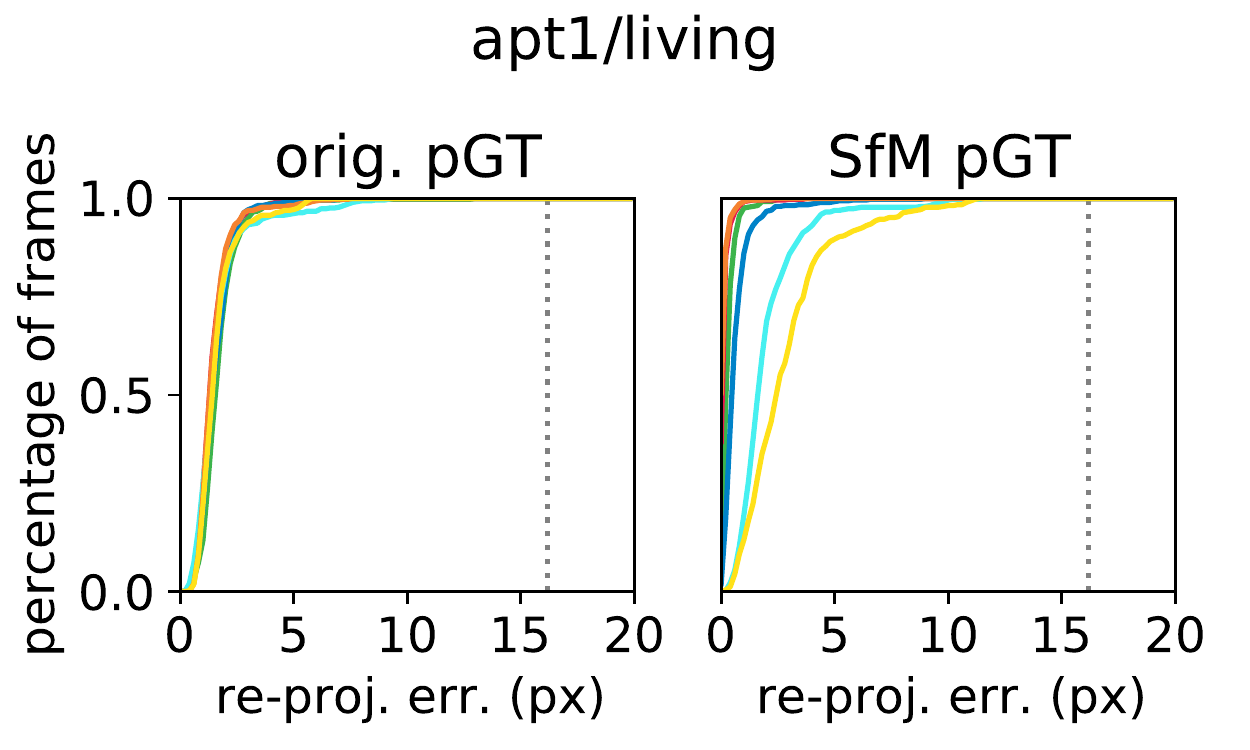}%
    \includegraphics[width=0.33\linewidth]{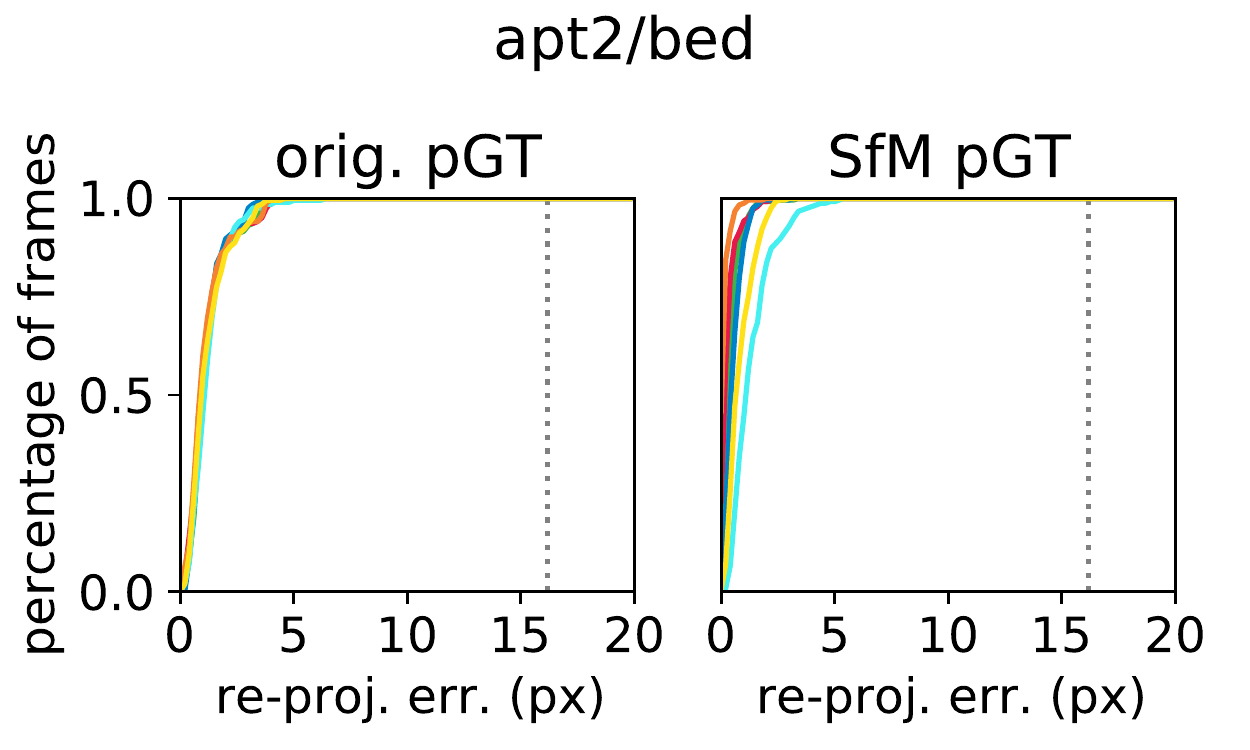}\\%
    \includegraphics[width=0.33\linewidth]{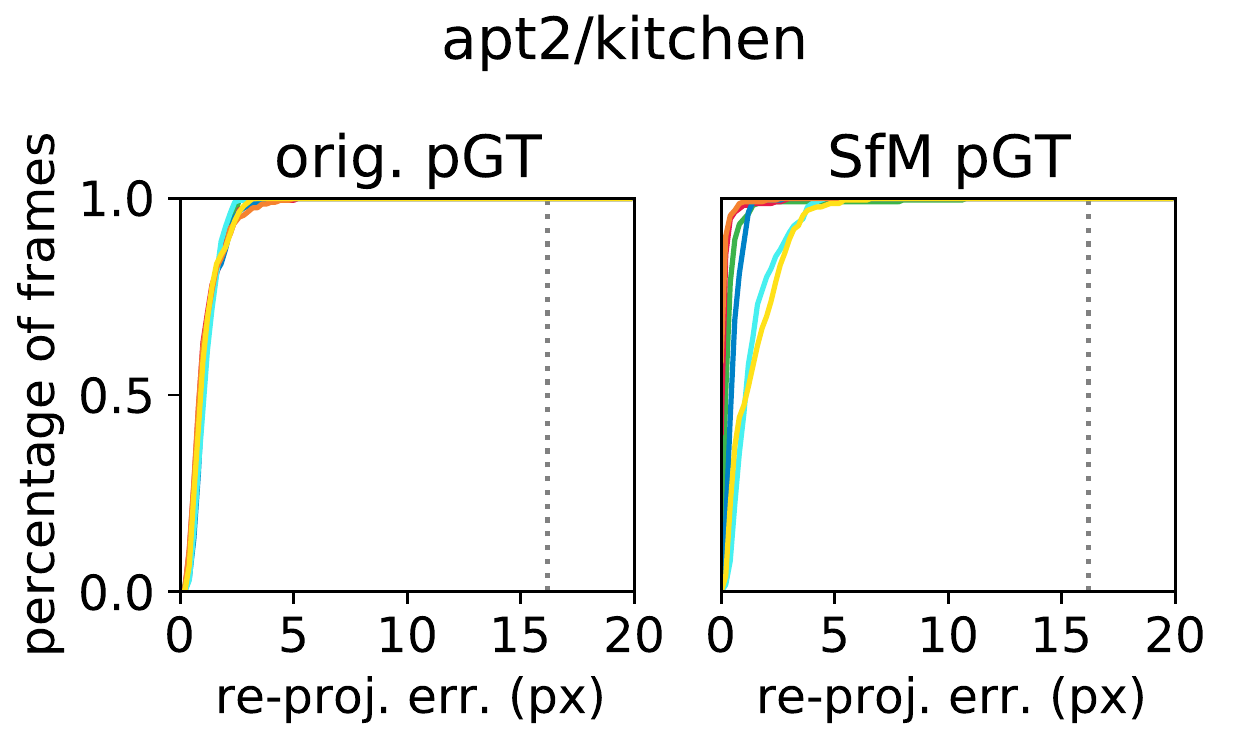}%
    \includegraphics[width=0.33\linewidth]{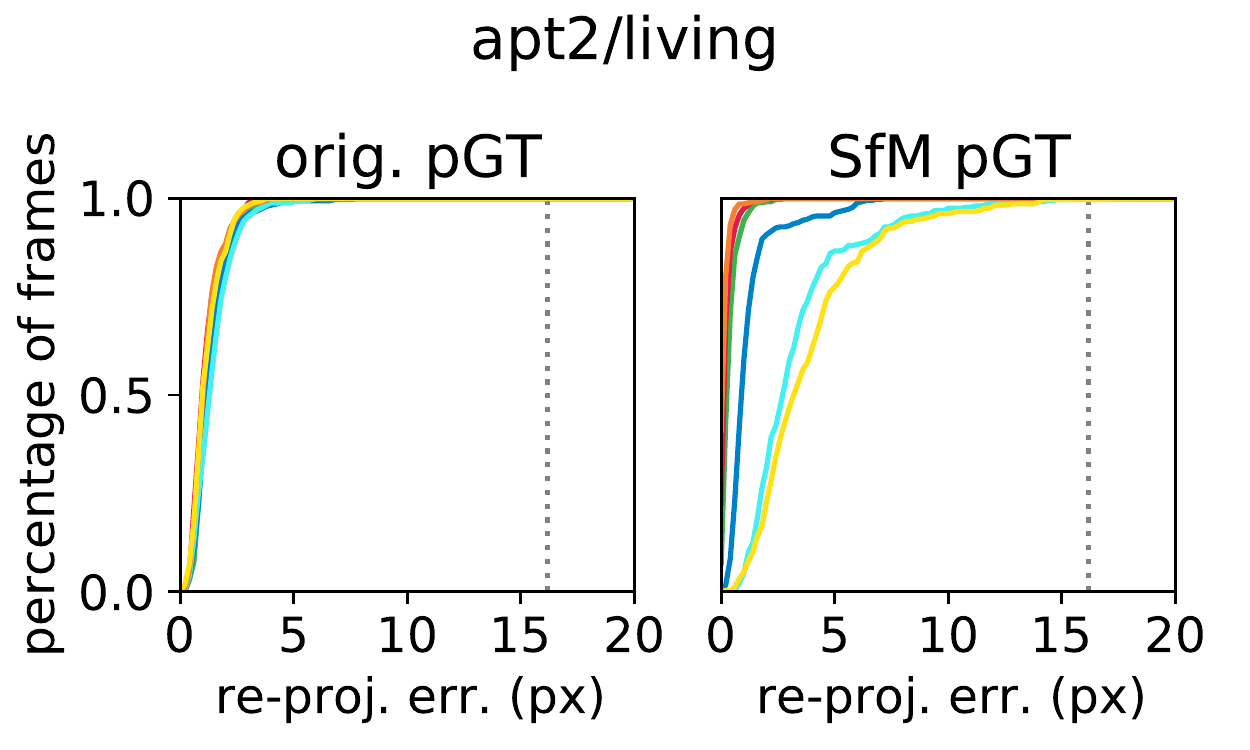}%
    \includegraphics[width=0.33\linewidth]{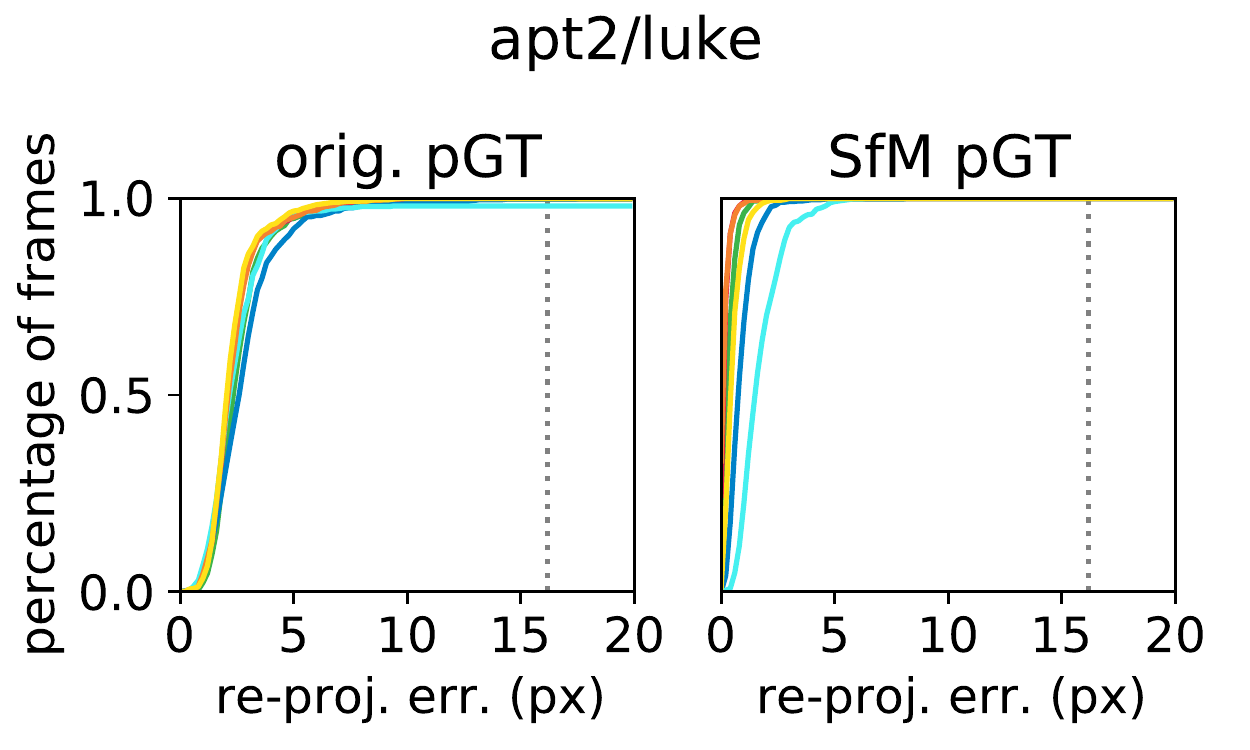}\\%
    \includegraphics[width=0.33\linewidth]{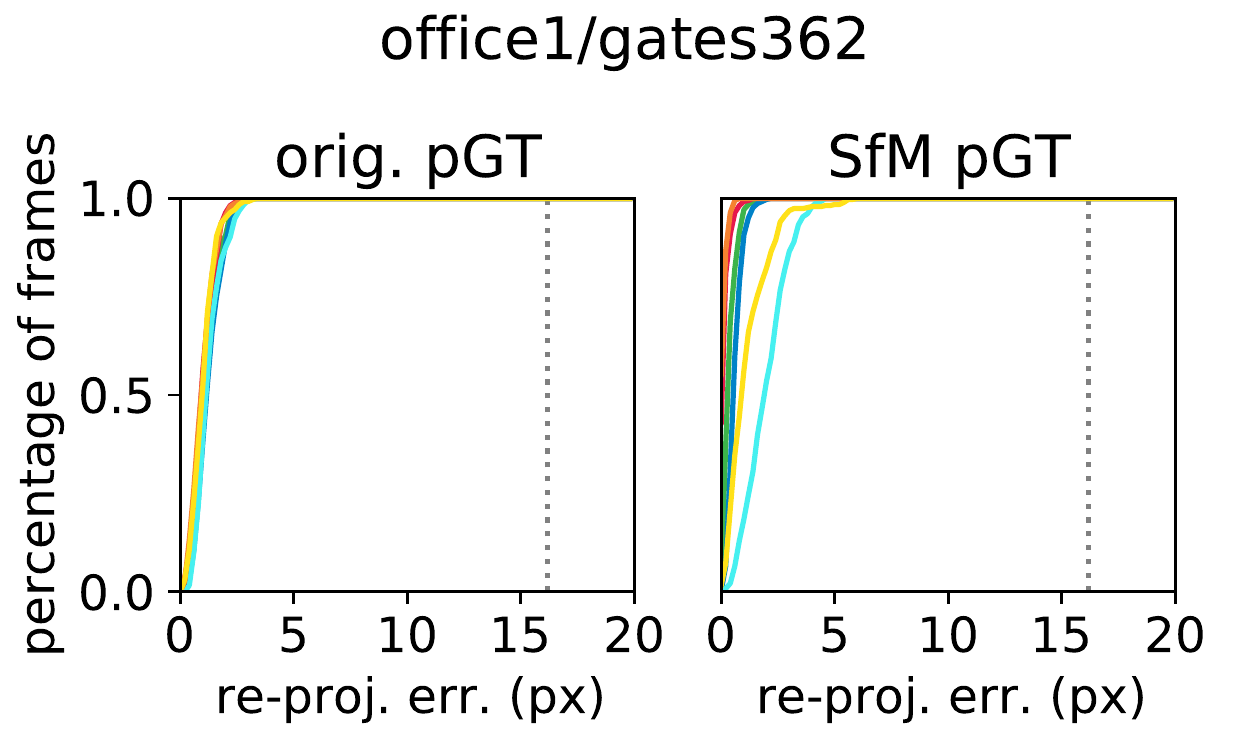}%
    \includegraphics[width=0.33\linewidth]{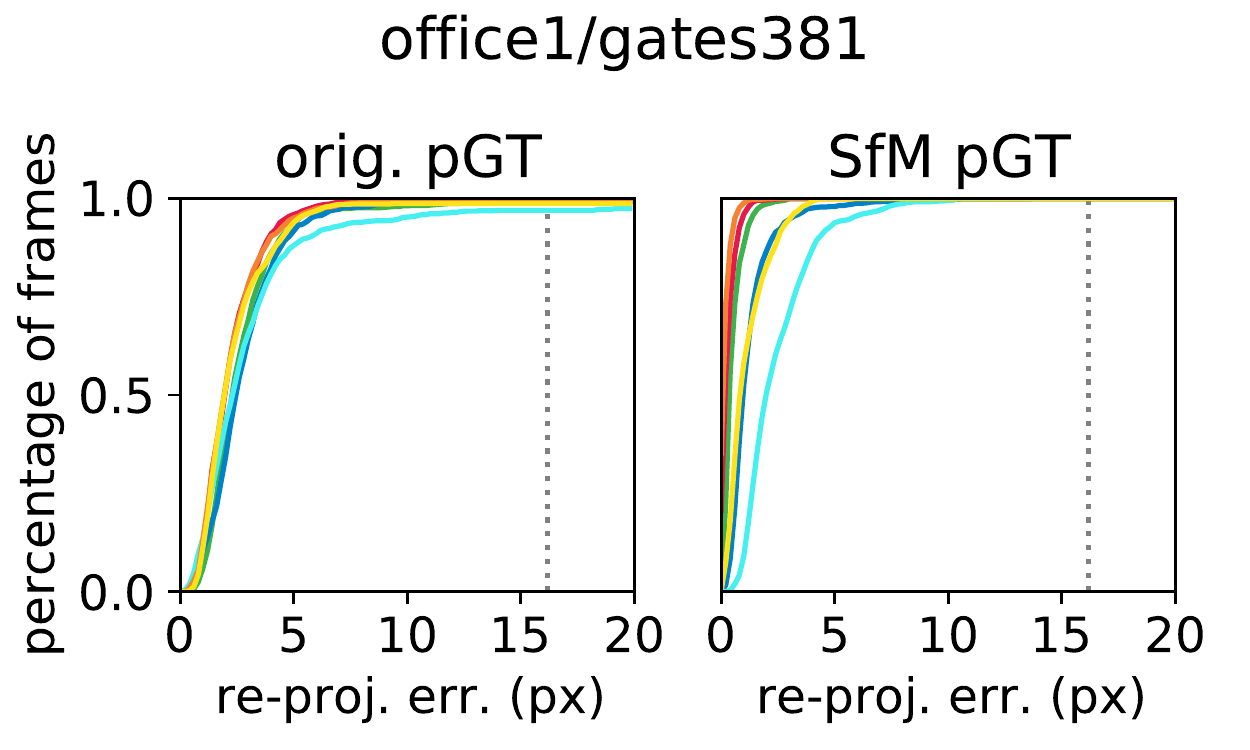}%
    \includegraphics[width=0.33\linewidth]{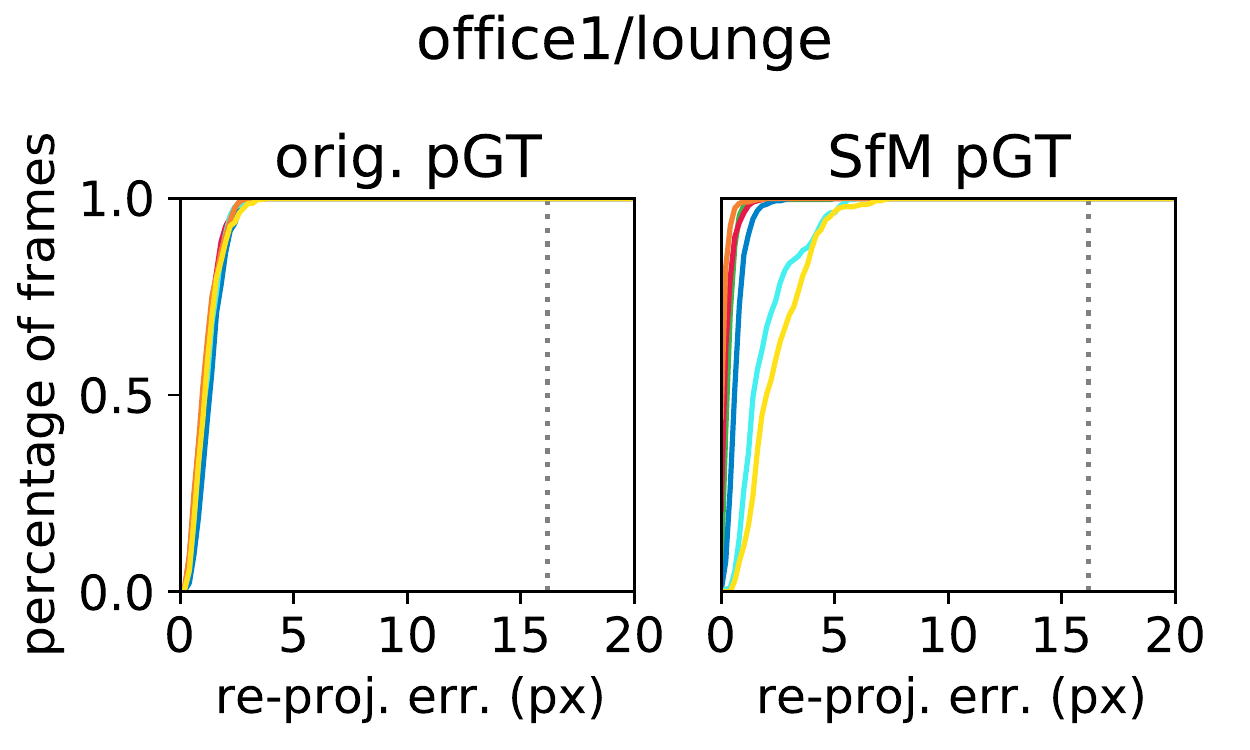}\\%
    \includegraphics[width=0.33\linewidth]{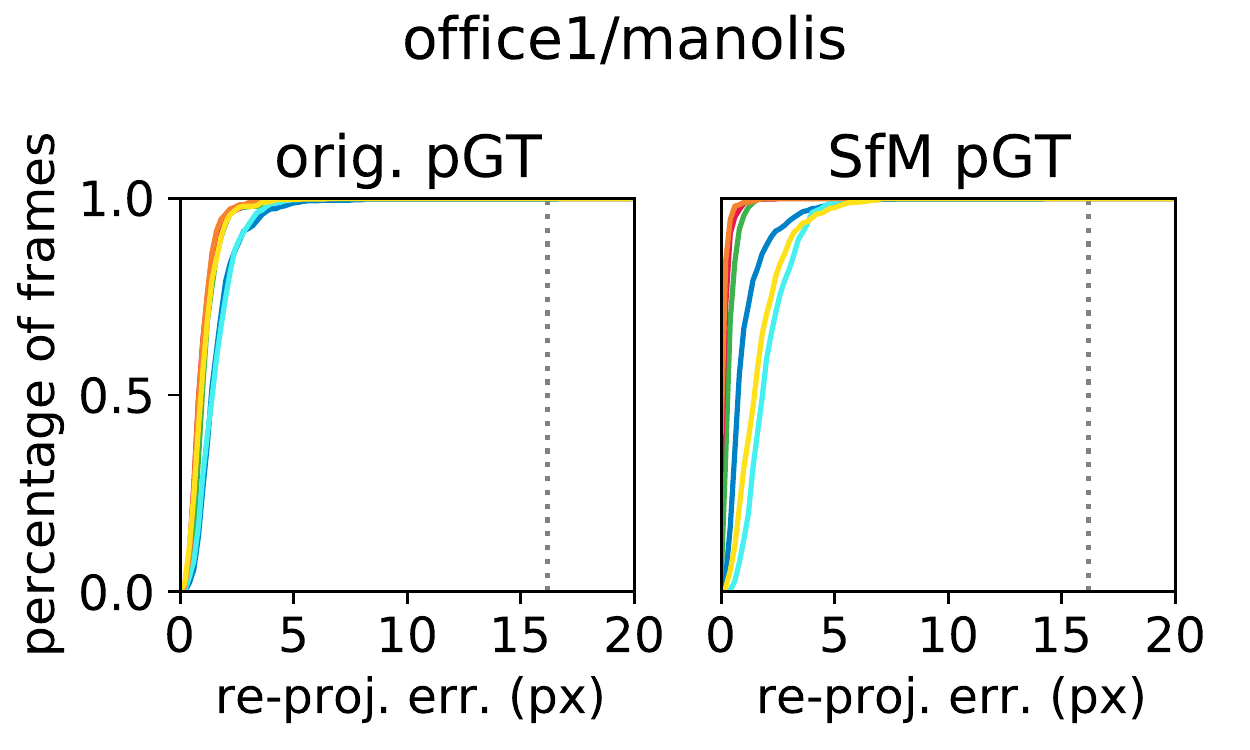}%
    \includegraphics[width=0.33\linewidth]{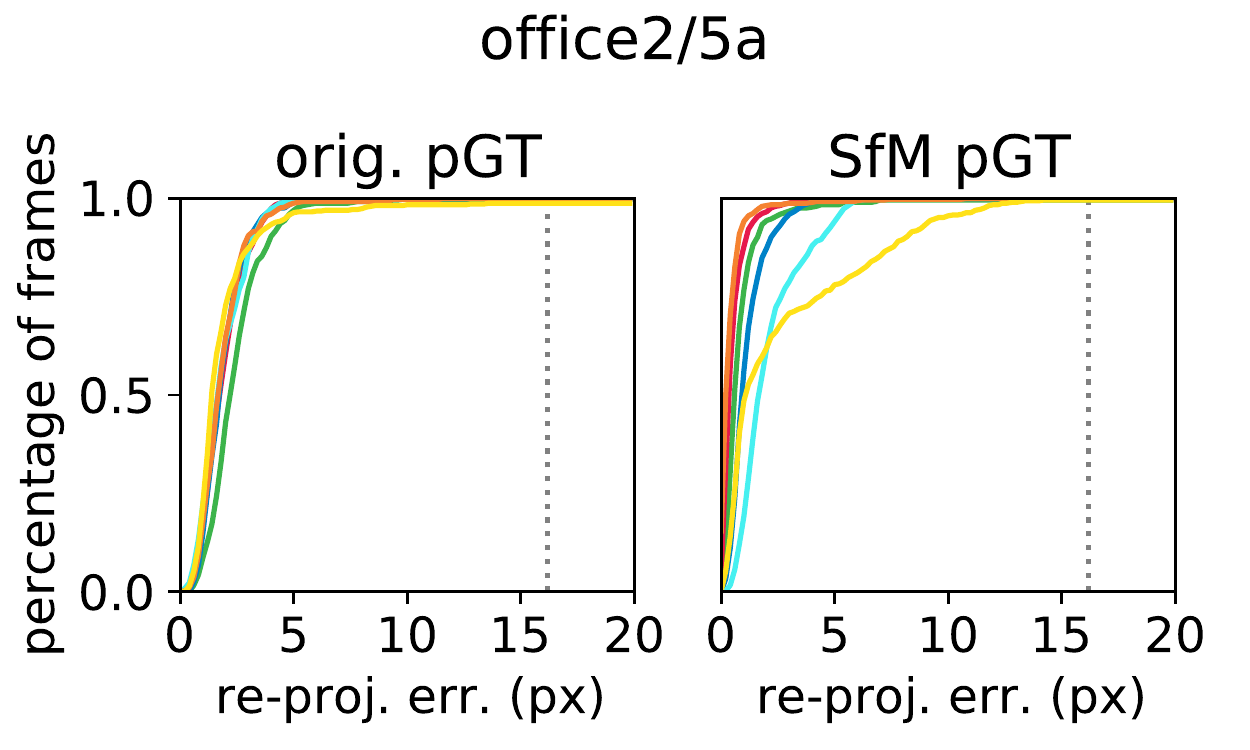}%
    \includegraphics[width=0.33\linewidth]{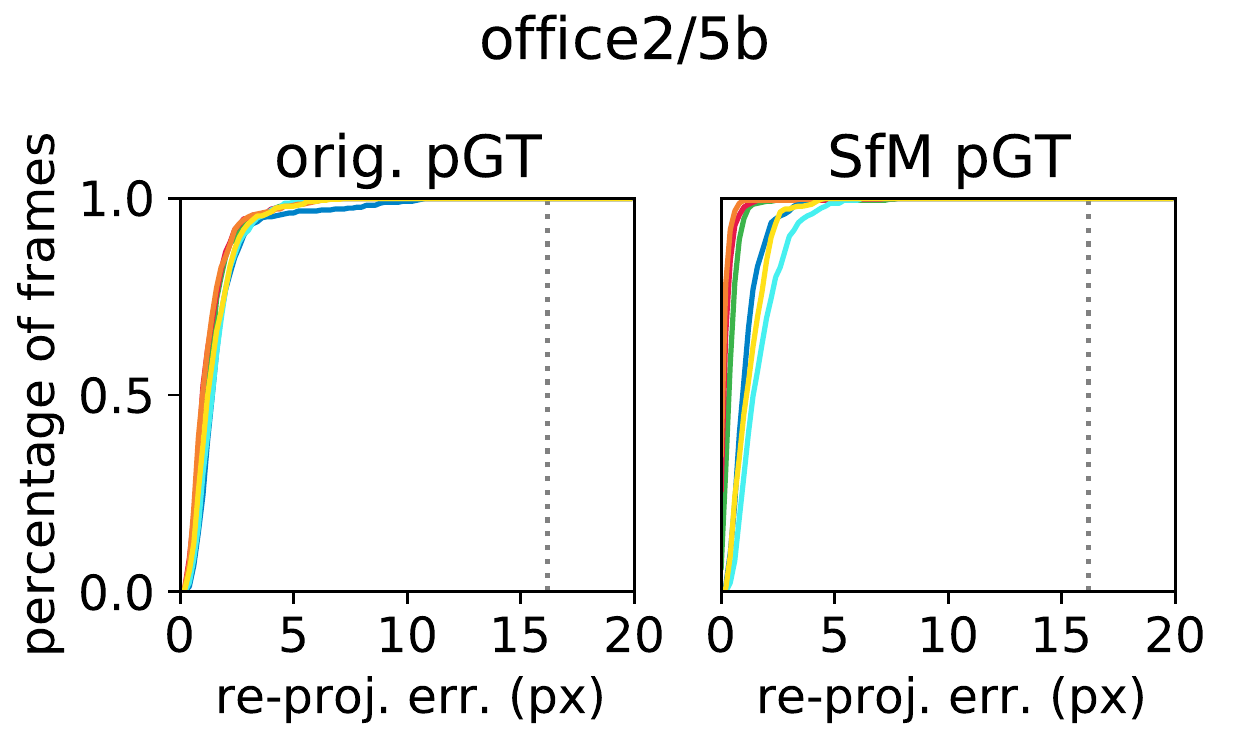}%
    \caption{\textbf{Mean DCRE for 12Scenes.} We show cum.~distributions of the DCRE (Dense Correspondence Re-Projection Error \cite{Wald2020ECCV}) for all scenes of 12Scenes \cite{Valentin20163DV}, taking the mean re-projection error per test image. The dotted vertical line corresponds to 1\% of the image diagonal.}
    \label{fig:12s_mean_plots}
\end{figure*}

As an extension to Fig.~\ref{fig:acc_repro}, we show cumulative pose error plots and cumulative DCRE \cite{Wald2020ECCV} error plots for all scenes of 7Scenes and 12Scenes, separately. See Fig.~\ref{fig:7s_pose_plots}, Fig.~\ref{fig:7s_max_plots} and Fig.~\ref{fig:7s_mean_plots} for pose error plots, max.~DCRE error plots and mean~DCRE error plots, respectively, for 7Scenes.
We show the corresponding plots for 12Scenes in Fig.~\ref{fig:12s_pose_plots}, Fig.~\ref{fig:12s_max_plots} and Fig.~\ref{fig:12s_mean_plots}.

{\small
\bibliographystyle{ieee_fullname}
\bibliography{paper}
}

\end{document}